\documentclass[sigconf, nonacm]{acmart}

\makeatletter
\def\runningfoot{\def@runningfoot{}}
\makeatother

\usepackage{microtype}
\usepackage{multirow}
\usepackage{graphicx}
\usepackage{subfigure}
\usepackage{booktabs} 
\usepackage[normalem]{ulem}

\usepackage{hyperref}

\usepackage{enumitem}

\usepackage{booktabs}       
\usepackage{amsfonts}       
\usepackage{nicefrac}       
\usepackage{microtype}      
\usepackage{xcolor}         

\usepackage{amsmath}
\usepackage{amsfonts}
\usepackage{amssymb}
\usepackage{amsthm}
\usepackage{graphicx}
\usepackage{color}
\usepackage{url}
\usepackage{stackrel}
\usepackage{subfigure}
\usepackage{caption}
\usepackage{xspace}
\usepackage{tabularx}
\usepackage{balance}
\usepackage{wrapfig}
\usepackage{multirow}
\usepackage{booktabs}
\usepackage{tabu}
\usepackage[linesnumbered,ruled,noend]{algorithm2e}
\usepackage{subfigure}
\usepackage{graphicx}
\usepackage{float}

\newcommand{\stitle}[1]{\vspace{0.8ex} \noindent\textsf{\textbf{{#1}}}}

\newenvironment{example}[1]{\vspace{0.0ex}\par{\noindent \bf Example of #1:}\\}{\qed\\\par}

\theoremstyle{remark}

\newcommand{\eg}{{\em e.g.}\xspace}

\newcommand{\squishlist}{
	\begin{list}{$\bullet$}{
		\setlength{\itemsep}{0pt}
		\setlength{\parsep}{3pt}
		\setlength{\topsep}{3pt}
		\setlength{\partopsep}{0pt}
		\setlength{\leftmargin}{1.0em}
		\setlength{\labelwidth}{1em}
		\setlength{\labelsep}{0.5em}
   }
}

\newcommand{\squishenum}{
	
	\begin{list}{\usecounter{scount}}{
		\setlength{\itemsep}{0pt}
		\setlength{\parsep}{3pt}
		\setlength{\topsep}{3pt}
		\setlength{\partopsep}{0pt}
		\setlength{\leftmargin}{1.2em}
		\setlength{\labelwidth}{1em}
		\setlength{\labelsep}{0.5em}
	}
}

\newcommand{\squishend}{
	\end{list}
}

\usepackage{amsmath}
\usepackage{amssymb}
\usepackage{mathtools}
\usepackage{amsthm}

\usepackage[capitalize,noabbrev]{cleveref}

\theoremstyle{definition}
\usepackage[textsize=tiny]{todonotes}

\usepackage{amsmath,amsfonts,bm}

\makeatletter
\def\widebreve{\mathpalette\wide@breve}
\def\wide@breve#1#2{\sbox\z@{$#1#2$}%
     \mathop{\vbox{\m@th\ialign{##\crcr
\kern0.08em\brevefill#1{0.6\wd\z@}\crcr\noalign{\nointerlineskip}%
                    $\hss#1#2\hss$\crcr}}}\limits}
\def\brevefill#1#2{$\m@th\sbox\tw@{$#1($}%
  \hss\resizebox{#2}{\wd\tw@}{\rotatebox[origin=c]{90}{\upshape(}}\hss$}
\makeatletter

\def\1{\bm{1}}

\DeclareMathAlphabet{\mathsfit}{\encodingdefault}{\sfdefault}{m}{sl}
\SetMathAlphabet{\mathsfit}{bold}{\encodingdefault}{\sfdefault}{bx}{n}

\usepackage{tabularray}
\UseTblrLibrary{siunitx,counter}
\usepackage{placeins}
\usepackage{balance}
\usepackage{flushend}

\usepackage[frozencache,cachedir=.]{minted}
\newcommand{\oursys}{\texttt{Data-Juicer}\xspace}
\newcommand{\para}[1]{\textit{\textbf{#1}}}
\newcommand{\opterm}[1]{\textit{#1}}
\newcommand{\pacterm}[1]{\textsf{#1}}
\newcommand{\twosubsection}[1]{\subsubsection{\textbf{#1}}}

\begin{document}

\title{\oursys: A One-Stop Data Processing System for Large Language Models}

\author{
	Daoyuan Chen$^*$,
	Yilun Huang$^*$, 
	Zhijian Ma$^*$,
	Hesen Chen$^*$,
	Xuchen Pan$^\dagger$,
	Ce Ge$^\dagger$, 
	Dawei Gao$^\dagger$, \\
	Yuexiang Xie,
	Zhaoyang Liu,
	Jinyang Gao,
	Yaliang Li$^\ddagger$,
	Bolin Ding$^\ddagger$,
	Jingren Zhou \\ 
	Alibaba Group
}

\renewcommand{\shortauthors}{}
\renewcommand{\shorttitle}{}

\begin{abstract}
The immense evolution in Large Language Models (LLMs) has underscored the importance of massive, heterogeneous, and high-quality data.
A data recipe is a mixture of data of different types and from different sources for training an LLM, which has been known as one of the most important factors that decide the LLM's performance.
Existing open-source tools for LLM data processing are mostly tailored for preparing specific data recipes. 
To continuously uncover the potential of LLMs, incorporate (after cleaning) data from new sources, and improve LLMs' general-purpose or domain-specific performance, we build a data processing system, named \oursys, with which we can efficiently generate diverse data recipes, explore different possibilities in forming the data mixtures, and evaluate their effects on the model performance.
Different from traditional data-analytics pipelines, \oursys faces some unique challenges. 
Firstly, the possible data sources for forming data recipes are truly heterogeneous and massive with various qualities (\eg, considering all web-pages on the Internet). 
Secondly, it is extremely expensive to precisely evaluate data recipes' impact on the LLMs' performance. 
Thirdly, sufficient flexibility needs to be provided to the end users of \oursys, model developers, to configure and evaluate different data recipes.

\oursys features a fine-grained abstraction of the pipeline for constructing data recipes, with over 50 built-in operators that can be freely composed and extended. By incorporating visualization and auto-evaluation capabilities, \oursys enables a timely feedback loop after data processing for both LLM pre-training and fine-tuning.
Further, \oursys is optimized and integrated with ecosystems for LLM training, evaluation, and distributed computing.
With the help of \oursys, we derive data recipes that achieve remarkable performance boosts on state-of-the-art LLMs, demonstrating up to $7.45\%$ increase in averaged score across $16$ LLM benchmarks and $17.5$\% higher win rate in pair-wise GPT-4 evaluations. More importantly, we hope that \oursys promotes broader data-centric research on training and understanding LLMs. \oursys and our data recipes are released and actively maintained at https://github.com/alibaba/data-juicer.

\end{abstract}

\maketitle

\renewcommand*{\thefootnote}{\fnsymbol{footnote}}
\footnotetext[1]{Co-first authors.} 
\footnotetext[2]{Equal contribution.} 
\footnotetext[3]{Corresponding authors, email addresses: \{yaliang.li, bolin.ding\}@alibaba-inc.com}
\renewcommand*{\thefootnote}{\arabic{footnote}}

\begin{figure*}[!ht]
	\centering
	\subfigure{
		\begin{minipage}[]{0.95\textwidth}
			\centering
			\includegraphics[width=\textwidth]{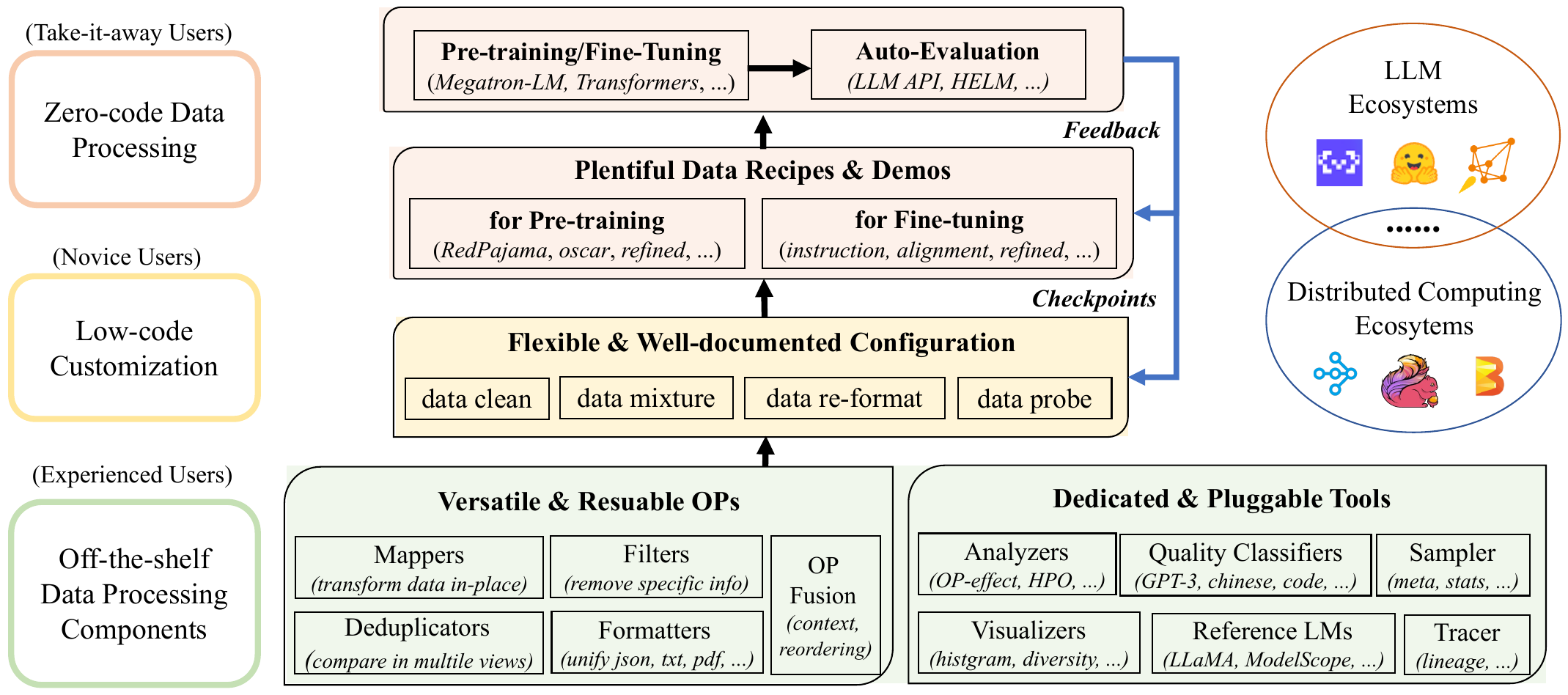}
		\end{minipage}%
	}

	\caption{Overview of \oursys.}

	\label{fig:overview}
\end{figure*}

\section{Introduction}
\label{sec:intro}

Large Language Models (LLMs) \cite{brown2020gpt3,ouyang2022training-instructgpt,Taori-github-2023-Stanford-alpaca,Sun-arXiv-2021-ERNIE3.0,Chen-arxiv-2021-codex,openai2023gpt4} have achieved unprecedented intelligence, enabling applications that would otherwise be infeasible due to unsatisfied performance. 
As the ``food'' for LLMs, \textit{data} plays a pivotal role in these exciting advancements \cite{Longpre-arxiv-2023-pretrainer,Gao-arxiv-2021-Pile,Penedo-2023-arxiv-Refinedweb,Xie-arxiv-2023-doremi}.
LLMs are built by pre-training on large-scale general-purpose corpus and are fine-tuned with specific-purpose data for alignment or downstream tasks.
For pre-training data, a collection of diverse data, including web texts, dialogues, academic papers, code bases, and others, help to develop the vast repository of knowledge and great applicability \cite{radford2018gpt1,brown2020gpt3,Li-2023-arxiv-Starcoder}. Fine-tuning data, which 
further refines LLMs and aligns model behavior with human values \cite{OpenAI-blog-2022-alignment,Askell-arxiv-2021-Anthropic,kopf-arxiv-2023-openassistant}. 
As ``garbage in, garbage out'' suggests, the input data for training or tuning an LLM has a direct impact on the quality of the derived model \cite{jain2020-data-quality-overview,gunasekar2023textbooks}. Building effective data processing solutions for LLMs remains a sophisticated yet fully under-explored task, given the common challenges in processing both pre-training and fine-tuning data, which pursue good data quality, proper data diversity, and large data volume.

Unfortunately, there exist only a few open-source projects contributing their LLM training data and the corresponding processing codes \cite{together2023redpajama,laurençon2023bigscience-roots}, particularly in comparison to numerous open-source projects on models and training infrastructures \cite{Scao-arxiv-2022-BLOOM,Touvron-arxiv-2023-LLaMA,Zeng-arxiv-2022-GLM,Black-CoRR-2022-GPT,nijkamp-arxiv-2022-Codegen,Biderman-arxiv-2023-Pythia,vicuna2023}. 
Such limited development of data processing will obstruct the progress of quantitatively understanding and enhancing LLMs from the perspective of data, especially accompanied by the following noteworthy {\underline{\bf C}}hallenges for LLM data processing.

{\bf ({C1}) High Heterogeneity in LLM's Data Recipe.}
LLMs involve several developmental stages and enable diverse usages including coding and dialog assistance, and even aiming at Artificial General Intelligence.
As a result, they demand an extensive variety of data types, formats, and quality in their training data, leading to highly complex data-processing pipelines.
A \textbf{\textit{data recipe}} for training or tuning an LLM is such a mixture of processed data from different types of sources, with their ratios and processing pipelines properly set \cite{together2023redpajama,nature-data-recipe-term}.
Existing systems, \eg, \cite{together2023redpajama,Scao-arxiv-2022-BLOOM}, release certain processing scripts to generate data recipes for the pre-training purpose, whereas \cite{Taori-github-2023-Stanford-alpaca, chen2023alpagasus} focus on data recipes for improving data diversity and quality in LLaMA's \cite{Touvron-arxiv-2023-LLaMA} fine-tuning stage.
However, due to the lack of abstraction of processing pipelines and composability of operators (OPs), such as those for data editing, cleaning, and filtering, it is difficult to incorporate new data sources in data recipes provided by these systems, or to extend their pipelines for exploring other possibilities of data recipes.

{\bf ({C2}) Timely Feedback for Data Recipe.}
The search space of LLM's data recipes is huge due to the high degree of heterogeneity in data sources and numerous ways to mix them (with proper processing OPs, combinations, and ratios).
We want to explore as many data recipes in the search space as possible with timely feedback to uncover the potential of LLMs and improve their performance.
However, as the size of an LLM (number of model parameters) is usually billions or even larger, it is super expensive, in terms of both the time and computational resources, to evaluate the impact of a data recipe on the LLM's performance by training or tuning it with the recipe \cite{shoeybi2020megatronlm} and running evaluation benchmarks \cite{liang2022holistic-helm}.

{\bf ({C3}) Usability and Customizability.}
The workflow of training or tuning LLMs starts from processing raw data.
Exacerbated by the above two challenges, there is an urgent need for a data-centric infrastructure, so that the model developers can easily re-use or implement their own OPs and tools for data processing, configure their processing pipeline, explore various data recipes, and evaluate the resulting LLMs' performance. We need such a system to accelerate the exploration and understanding of LLMs' potentials.

{\bf ({C4}) Massive Data Volume.}
Last but not least, LLMs are trained on vast corpora, with data volumes stretching to an unprecedented magnitude of billions or even trillions of tokens (a modeling unit of text dependent on the used tokenizer \cite{Kudo2018SentencePieceAS}). 
Efficient LLM data processing of such volume is critical but arduous.
However, considerations on system performance optimization are often bypassed by existing studies, leaving significant room for enhancement in ensuring the stability of data processing and facilitating the deliveries of processed data and trained weights for LLMs.

\stitle{Overview of \oursys.}
In this paper, we advocate for a one-stop data processing system that addresses these challenges, enabling comprehensive, user-friendly, and efficient data processing abilities to facilitate data-centric LLM research and development. 
The proposed system, named \oursys and illustrated in a bottom-up view in Figure \ref{fig:overview}, is strategically designed to generate data recipes making data more ``juicy'' and digestible for LLMs. 
We decouple the mixture elements of existing solutions for LLM data processing, such as specific data types, auxiliary models, and downstream tasks.
As highlighted by the green boxes, \oursys fosters a fine-grained abstraction and implementation of composable modules with over 50 versatile OPs and dedicated tools.
We make \oursys end-to-end configurable to help prepare traceable, comparable, and refinable data recipes at various scenarios of LLM pre-training and fine-tuning, as shown in the yellow and pink boxes. 
Coupled with established auto-evaluation capabilities, \oursys supports a timely feedback loop at multiple development stages of data recipes and LLMs, thereby promoting the production of valuable LLM data.

To meet diverse user backgrounds and needs (marked by the left three rectangle boxes), we design \oursys as an easy-to-use, flexible and extensible system. 
Beginners are shielded from underlying complexities and benefit from numerous ready-to-use datasets, data recipes, and pluggable tools, supporting zero-code LLM data processing.
With the help of the flexible configuration module, experienced users can simply modify built-in data recipes, reorganize the order of OPs and tools, and tune the value of their hyper-parameters, to meet their lightweight customization needs. 
Thanks to the standardization and modularization, advanced users are empowered to conveniently extend and register their new OPs and tools into \oursys, facilitating quick engagement in secondary development. 
Furthermore, we offer more than a dozen interactive tutorials implemented by streamlit \cite{streamlit} to help users with their LLM data processing journey.

\oursys hinges itself on the \pacterm{Huggingface-datasets} library \cite{lhoest-etal-2021-datasets}, providing a unified intermediate representation of data and achieving optimized space-time efficiency and robustness through various techniques such as context management, OP fusion, caching, and checkpoint mechanisms. 
Furthermore, as the right two circles show, \oursys seamlessly integrates with ecosystems for LLM training and evaluation such as Megatron-LM \cite{shoeybi2020megatronlm} and HELM \cite{liang2022holistic-helm}, and distributed computing ecosystems such as Ray \cite{ray} and Beam \cite{apache-beam}, thus facilitating comprehensive LLM data processing and enhancing large-scale data processing capabilities.

Leveraging the proposed system, we refine several open-sourced datasets and derive numerous data recipes for both LLM pre-trained and fine-tuning. 
These refined datasets are not only higher in quality but also more digestible by LLMs, leading to effective performance improvements of LLMs. 
Empirical analysis showcases an improvement of up to $7.45\%$ averaged score across $16$ LLM benchmarks using our refined pre-training data. 
Even pre-trained on only $43\%$ quantity of compared data, we observe superior performance over state-of-the-art (SOTA) LLMs such as Falcon \cite{Ebtesam-arxiv-2023-Falcon40b}.
Moreover, compared with SOTA LLMs fine-tuned on competitive open English and Chinese data, LLMs fine-tuned on \oursys's data gain an average of $10.3\%$ higher win rate of pair-wise GPT-4 evaluation, while with an average $56.8\%$ fewer data quantity.
Finally, we introduce its utility in real-world deployment, and validate its superior system efficiency and scalability of \oursys, by up to $88.7\%$ reduction in single-machine processing time and $77.1\%$ savings in memory usage, and $7.91$x distributed processing acceleration.

\stitle{Contributions.}
Our contributions are summarized as follows:
\begin{itemize}[leftmargin=*]
	\item 
	We propose and build a novel system for LLM data processing, \oursys, which is featured by decoupled modules and over 50 versatile OPs and tools. To easily dive into data quality and insights, \oursys fosters a timely feedback loop with interactive visualizations and auto-evaluation capabilities. 
	
	\item
	Demonstrated by extensive empirical evidence, \oursys produces numerous high-quality data recipes to enhance LLMs and exhibits superior system performance, powered by dedicated optimization and integrated distributed computing ecosystems.
	
	\item 
	We integrate data-centric methodologies for LLM data processing and LLM development with user-centric interface designs, with the aim that \oursys can ease access for diverse users and democratize LLM data processing.
	
	\item 
	To promote further research and development, our system, data recipes, and tutorials are maintained and released at \url{https://github.com/alibaba/data-juicer}, which we hope can help pave the way for next-generation production paradigms of LLM data.
\end{itemize}

\stitle{Organization.}
The subsequent sections describe \oursys in detail. 
Sec. \ref{sec:background} elaborates on the background and related studies.
Sec. \ref{sec:op_pool} outlines our OP pool, as a response to high heterogeneity of LLM data recipes (\textbf{C1}).
Sec. \ref{sec:feedback} delves into our formulation of timely feedback loops for data processing and development of LLMs (\textbf{C2}). 
Sec. \ref{sec:recipes_plugins} details our repository of data recipes and tools that counteract usability and customization issues (\textbf{C3}).
Sec. \ref{sec:sys_optimization} expounds on the employed system optimization to tackle massive data volume (\textbf{C4}).
Sec. \ref{sec:eval} focuses on an extensive empirical evaluation for the quality of data recipes, performance and usability of \oursys.
Lastly, we draw a summary in Sec. \ref{sec:conclusion}.
\section{Background and Related Works}
\label{sec:background}
\subsection{Large Language Model (LLM) Data}
\label{background:llm}

\para{Large Language Models (LLMs).} 
Language modeling is a crucial component for achieving machine intelligence \cite{Zhong-2023-arxiv-AGIEval,Malinka-arxiv-2023-Education}. 
In the last few years, this field has witnessed remarkable advancements, particularly with the emergence of the pre-training and fine-tuning paradigms, where language models undergo an initial phase of training with a general-purpose corpus before being fine-tuned with specific-purpose tasks \cite{Peters-NAACL-2018, devlin-etal-2019-bert}.
This procedure has yielded exceptional performance across a spectrum of natural language processing (NLP) tasks \cite{Lewis-ACL-2020-BART, radford2019gpt2}.

Recently, taking advantage of the highly parallelizable nature of the self-supervised Transformer architecture, the scales of model parameters and training corpus for LLMs have significantly been increased \cite{openai2023gpt4,Du-ICML-2022-GLaM}. 
Meanwhile, LLMs have aroused considerable interest in the potential of artificial general intelligence \cite{wei-arxiv-2023-zero,Izacard-arxiv-2022-Few,Feng-arXiv-2023-Towards,Cai-arxiv-2023-Tool,Hao-2023-arxiv-ToolkenGPT,rucLLMSurvey,Bubeck-arxiv-2023-Sparks-AGI}.
While model-centric studies proliferate, how to better process LLM data remains an intricate domain yet to be completely unfurled, whether for pre-training or fine-tuning data.

\para{Pre-training Data.}
Pre-training serves as the foundation for LLM intelligence.
By being trained on large amounts of high-quality data, LLMs can acquire elementary language comprehension and generation capabilities \cite{Han-AIopen-2021-PTM}.
Aiming to elucidate the link between data and LLMs intuitively, let us consider a typical pre-training objective prevalent among mainstream LLMs. 
Given a token sequence $[t_1, ..., t_i, ..., t_n]$, an LLM $\theta$ is trained to maximize the joint probability of the text as follows: 
\begin{equation}
\small
\theta_0 = \arg\max_\theta \sum_{i=1}^{n} \log p(t_i | t_{1:i-1}; \theta).
\end{equation} 
This objective is for auto-regressive language modeling and allows the pre-trained $\theta_0$ to predict the probability of the next token by adhering to the inherent sequential ordering of the language \cite{Wang-ICML-2022-What}.

Exploiting this unified yet simple modeling goal, researchers collect a large volume and diverse range of corpus data, which usually contains hundreds of billion tokens or even trillion tokens.
After tokenization and pre-training, LLMs have succeeded in stimulating a wide range of advanced capabilities. 
The LLM pre-training data generally includes various types derived from the web crawlers \cite{commoncrawl,Penedo-2023-arxiv-Refinedweb}, dialogues or social media \cite{Zhang-arxiv-2022-OPT}, book-length formal texts \cite{Gutenberg,Zhu-ICCV-2015-Aligning}, rigorous encyclopedias and academic texts \cite{Wikipedia,Gao-arxiv-2021-Pile}, structured coding texts \cite{Chen-arxiv-2021-codex,Li-2023-arxiv-Starcoder}, and more texts from financial, medical and legal domains \cite{Yang-2023-arxiv-FinGPT,Li-arxiv-2023-ChatDoctor,Sun-arxiv-2023-A}. 
A challenge is nonetheless posed in the careful processing and formulation of pre-training data to filter noise, redundancy, irrelevance, and potentially toxic \cite{Longpre-arxiv-2023-pretrainer,Gehman-2023-arxiv-RealToxicityPrompts}.

\para{Fine-tuning Data.}
Numerous studies have underscored that fine-tuning – the process of refining pre-trained LLMs using a smaller, task-specific dataset – can further enhance or unlock additional capabilities of LLMs \cite{Honovich-arxiv-2022-Unnatural,Lester-ACL-2021-The,Wei-2023-arXiv-symbol,Wei-arxiv-2022-Emergent}. Crucially, this process also paves the way for better aligning the behavior of these advanced models with human values and preferences \cite{Liu-2023-arxiv-G-Eval,OpenAI-blog-2022-alignment}.

In this phase, though the data volume decreases exponentially compared to the pre-training phase, the format of fine-tuning data is quite different \cite{qiao2023reasoning}.
Typically, given a textual dataset \{$(x_1, s_1, y_1), ...,$ $(x_j, s_j, y_j), ..., (x_m, s_m, y_m)$\}, the goal of fine-tuning is to adjust the pre-trained LLM $\theta_0$ to find $\theta^*$ that maximizes the likelihood of the task-oriented response $y_j$ for the user query $x_j$:
\begin{equation}
\small
\theta^{*} = \arg\max_{\theta} \sum_{j=1}^{m} \log p(y_j | x_j, s_j; \theta); \quad  \theta \leftarrow \theta_0.
\end{equation}
Here $s_j$ stands for task-specific instructions, such as ``summarize the following text: '', optionally accompanied by a few demonstrative samples for in-context learning \cite{brown2020gpt3}.

The fine-tuning data can be broadly categorized into two types: 
\textit{Instruct Fine-Tuning (IFT)} datasets to enhance the instruction-following abilities of LLMs and are usually adapted from existing NLP benchmarks \cite{bach-etal-2022-promptsource,Longpre-arxiv-2023-The}; and \textit{Chat Fine-Tuning (CFT)} datasets for enhanced dialog ability and human value alignment \cite{Taori-github-2023-Stanford-alpaca,ouyang2022training-instructgpt}.
There are preliminary explorations emphasizing the importance of data diversity over volume for fine-tuning data \cite{Chung-arxiv-2022-Scaling,Wang-EMNLP-2022-Super}.
Several studies also indicate that data types representing human values can potentially lead to degraded general performance, a phenomenon known as the ``alignment tax'' \cite{ouyang2022training-instructgpt}. 
However, how to more effectively process the fine-tuning data to maximize its usefulness and minimize potential risks remains an open area for further investigation.

\begin{table*}[!ht]
	\centering
	\caption{Overview of the operator (OP) pool in \oursys, with a detailed list continuously maintained at the official documentation: https://github.com/alibaba/data-juicer/blob/main/docs/Operators.md. }
	\label{tab:all_ops}
	\begin{tblr}{
			width = \linewidth,
			colspec = {Q[85]Q[123]Q[119]Q[119]Q[92]Q[402]},
			cells = {c},
			hline{1,6} = {-}{0.08em},
		}
		\textbf{Category} & \textbf{Function} & \textbf{Input} & \textbf{Process Level} & \textbf{Output} & \textbf{OP Usage Examples}\\
		\hline
		Formatters & Data format unifying & Dataset & Dataset & Dataset & Load and unify dataset-hub, txt, json, md, codes, html, pdf, docx, ...\\
		\hline
		Mappers & In-place text editing & Sample & { Single-sample;\\Multi-samples} & Sample; Samples & Transform specified headers, textual elements; Fix messy codes; Enable text enhancement\\
		\hline
		Filters & Conditional text removing & Sample & Single-sample; Dataset & Boolean & Filter by meta-info, stats (\eg, lines count); model scores; external resources (\eg, flagged words)\\
		\hline
		Dedup-licators & Duplication removing & Single or Paired Dataset & Dataset & Dataset & Compare with hash-based and vector-based deduplication methods
	\end{tblr}
\end{table*}

\para{The Symbiotic Nature of Pre-training and Fine-tuning Data.}
It is worth pointing out the analogous properties shared between these two types of data, which motivate our synergetic approach when bearing quality, diversity, and volume considerations in mind. 

Specifically, the quality aspect of the text has been studied extensively in existing literature \cite{Longpre-arxiv-2023-pretrainer}.  
Efforts have been made to enhance aspects such as text structure, the soundness of arguments, contextual richness, writing correctness, comprehensiveness, levels of anonymization, and harmlessness. 
The widespread implementation of cleaning, deduplication, and anonymization processes in pre-training data typifies the aforementioned pursuit. For example, researchers may opt to iterate over additional epochs with Wikipedia-style data in LLM training \cite{Touvron-arxiv-2023-LLaMA}.
Similarly, fine-tuning data processing also employs filtering, deduplication, and detoxification strategies, aiming to enhance the user experience and the degree of aid offered by LLMs \cite{Gehman-2023-arxiv-RealToxicityPrompts,chen2023alpagasus}.

Diversity is another shared property studied at length in both types of data. 
Mixing various types of data and finding suitable mixture weights to achieve appropriate diversity has been a primary concern in works for pre-training data processing \cite{Xie-arxiv-2023-doremi}. Analogously, efforts for fine-tuning data aim to increase multi-view diversity such as tuning tasks and expression styles, which further underscores this shared property \cite{Raffel-JMLR-2020-Exploring,ouyang2022training-instructgpt,Taori-github-2023-Stanford-alpaca}.

In addition, the pursuit of quality and diversity tends to trade off with data volume, which is also reflected in these two types of data. 
Researchers have incessantly strived to empower LLMs with massive amounts of data, hoping to encapsulate as much human knowledge as possible. 
For instance, there has been an influx in pre-training data volumes to terabyte levels \cite{openai2023gpt4,Penedo-2023-arxiv-Refinedweb}, and fine-tuning data volumes have grown from mere thousands to millions \cite{Wei-ICLR-2022-Finetuned,bach-etal-2022-promptsource}. 
However, the counter effects of these initiatives are also brought into these large volumes of data, including heightened noise, potential inferior quality, and increased bias, which necessitate additional data processing efforts and surging LLM training overheads.

\subsection{Existing LLM Data Processing Solutions}
LLM data processing is an early area that is still working towards common standards, and we aim to embody a pioneering system for the community.
With a commitment to open-source ethos, \oursys caters to the increasing demand for versatile, flexible, user-friendly and efficient \textit{data processing} solutions, details of which will be described later. 
This contrasts the well-known LLMs that were largely closed-source in \textit{data} or \textit{data processing}, such as the GPT derivatives \cite{openai2023gpt4,brown2020gpt3,Shen-2023-arXiv-Hugginggpt,Chen-arxiv-2021-codex}, LLaMA derivatives \cite{Touvron-arxiv-2023-LLaMA,ChatLLaMA,Taori-github-2023-Stanford-alpaca,vicuna2023,su-2023-arxiv-pandagpt}, and others \cite{Zhipu-GitHub-2023-ChatGLM2-6B,Ebtesam-arxiv-2023-Falcon40b,Ren-arXiv-2023-PanGusigma,Zhang-arxiv-2022-OPT,wu-arxiv-2023-bloomberggpt}.
While some progress has been made in the open-source LLM data processing landscape  \cite{laurençon2023bigscience-roots,bach-etal-2022-promptsource,together2023redpajama,DolmaToolkit}, they have not fully delivered the abstraction and breadth of functionalities that \oursys aims to bring to the forefront of the field. 

Examining this from the perspective of the target datasets, existing works typically \textit{fixate on specific data sources and use cases for LLMs}, spanning alignment of specialized English sub-datasets for LLaMA pre-training \cite{Touvron-arxiv-2023-LLaMA}, assembly of multi-lingual corpora for pre-training \cite{laurençon2023bigscience-roots}, or crowdsourcing for fine-tuning prompt data \cite{bach-etal-2022-promptsource}. 
However, they lack the systematic and modular processing abilities required to proficiently manage heterogeneous data, which is an area \oursys strives to push its boundaries. 
These limitations become especially apparent when handling new data types, engaging in language transfer, or implementing particular data cleaning and transformations for LLM applications.

Moreover, existing works suffer from sub-optimal \textit{usability and ability to explore data insight}. 
Most of these works only offer the processed data along with purpose-built processing codes specific to those data, lacking in ease-of-use considerations and support of assistive tool-kits. 
This hinders their adaptability to diverse users and alternative usages. 
Users might face a daunting task when substituting data processing goals or conducting analyses due to a dearth of complementary data-analytical capabilities. 
The re-development of data processing tools and analytical methodologies, specifically tailored for LLMs, remains largely uncharted territory.

Furthermore, the focus of current works gravitates towards functionality rather than \textit{system performance}, leaving large room for enhancement in efficiency, space management and scalability. 
Noteworthy shortcomings include reliance on single-machine Python scripts, inappropriate handling of large-scale data, and poor processing speeds due to the utilization of Python’s plain \verb|dict| object. 

We will provide further empirical comparisons in terms of both the quality of the generated data recipes (Sec. \ref{exp:recipe_quality}) and the performance of the data processing system (Sec. \ref{exp:system_perf}).

\section{Standardized Operator Pool}
\label{sec:op_pool}
In addressing the heterogeneity of data recipes for LLMs (Challenge 1 in Sec. \ref{sec:intro}), we devise a set of standardized operator (OP) pool. 
As outlined in Table \ref{tab:all_ops}, the OPs are organized into four primary categories: \textit{Formatters, Mappers, Filters,} and \textit{Deduplicators,} which incorporate diverse categories, functions, inputs, processing levels, outputs, and application scenarios. 
Core principles of decoupling and composability guide their structuring, resulting in a varied yet standard set of procedures that contribute to flexibility and user interaction at multiple processing levels.
This strategic implementation enhances reusability and reduces complexity, aiding streamlined and decoupled data recipe construction.

\subsection{Unified Data Representation}
\label{sec:unify_op}
We first introduce \opterm{Formatter} OPs designed to unify diverse data sources into an intermediate data representation. 
Specifically, we choose to build \oursys upon \pacterm{Huggingface-datasets} \cite{lhoest-etal-2021-datasets} due to its compatibility with mainstream LLM datasets and its column-oriented storage ability backed by \pacterm{Apache Arrow} \cite{apache-arrow}. 
Our Formatters maintain data objects that are instantiated from several unified base classes that simplify the process design for follow-up OPs and facilitate data accessing efficiency.
We support numerous text input formats - txt, JSON, parquet, html, md, pdf, code files such as .py and .cpp, amongst others - and homogenize them into a structured format composed of certain columns with \textit{nested access support}, which are conceptually organized by three primary parts ``text'', ``meta'', and ``stats''. 
These parts respectively hold the raw textual data, metadata information (\eg, date and version), and statistical data that can be generated and consumed by \oursys's other OPs and tools.
This interface works at either the text \textit{sample} or \textit{dataset} level, and is independent of underlying in-memory or disk data layout, alleviating the potential worry over heterogeneous data formats by OP developers.

\subsection{Versatile Data Processing}
Next, we elaborate on the functionality of the OP pool in \oursys, which is pivotal to the comprehensive data processing tailored for LLMs. 
Besides the Formatters, which play an essential role in unifying data formats and ensuring a consistent and efficient data flow throughout the processing pipeline, we now give more details about the other three types of data-transformation OPs in Table \ref{tab:all_ops}.

Mappers facilitate crucial functionalities of in-place text editing, necessary for single-sample or multi-sample processing across various needs of LLM data processing, such as modifying texts for pre-training and enhancing text diversity for fine-tuning. 
They effectively handle processing tasks like the removal of specific file headers, messy code rectification, and text enhancements.

Filters come into play by conditionally filtering texts via individual-sample metrics, dataset-level statistics, or external resources like stop-word lists. 
In doing so, they can eliminate unnecessary text samples, contributing to data focus, cleanliness, and the cost reduction of follow-up LLM training processes significantly.

Deduplicators reduce potential storage waste and improve efficiency. 
As indicated by several studies \cite{Lee-ACL-2022-Deduplicating,Kandpal-ICML-2022-Deduplicating,Carlini-arxiv-2022-Quantifying}, duplicate samples adversely affect both the pre-training stability and the performance of LLMs. 
Besides, Deduplicators help prevent unintentional data leakage during training into evaluation benchmarks, particularly for zero-shot or few-shot tasks \cite{Hendrycks-ICLR-2021-Measuring}. 
To ensure accurate detection and removal of duplication, we provide efficient and robust methods including hash-based and vector-based comparisons \cite{BRODER2000630-minhash,simhash,shahmirzadi2019text-simtext}.

It is noteworthy that the outputs of Filter OPs are Booleans, which helps to decouple the implementations of actual data processing and computation for various statistics. 
This dedicated segregation results in two key advantages. 
Firstly, it enables our dedicated analyzer-related tools (detailed in Sec. \ref{sec:dedicated_tools}) to utilize these computed statistics for the entire dataset, rather than a filtered subset. 
Users are also allowed to generate fingerprints for specific partial samples. 
Secondly, this decoupling enhances compatibility between \pacterm{Huggingface-datasets} and \oursys, thereby enabling the efficient reuse of the \textsc{Dataset.map} and \textsc{Dataset.filter} interfaces to perform these sub-processes in a streamlined manner.
As a result, users can effortlessly extend their own custom OPs that only vary from existing OPs in specific partial processing behaviors.
In Appendix \ref{sec:appendix_base_class_ops}, we offer an illustrative code example of this decoupling in Listing \ref{listing:base_op_classes}. 

\subsection{Composability}
\oursys's OPs serve as a testament to our system's versatility. 
They enable users to effortlessly process a variety of data types in a composable and modular manner, showcasing \oursys's dedication to user adaptability and high-quality data production for LLMs. 
Besides the \textit{functions, inputs, outputs} and \textit{processing levels} summarized in Table \ref{tab:all_ops}, this composability is embedded in more facets, including the \textit{fields to be processed, OP hyper-parameters,} and \textit{recommended use cases of each OP.}

Each OP in \oursys is designed to serve a distinct function and can be commanded by users to process different text fields. 
For example, OP A could process the sample field ``text.abstract'', while OP B could focus on ``text.main\_body''. 
By default, each OP process on ``text'' field, which can be freely specified to other ``meta'' or ``stats'' related data fields according to users' needs. 
This adaptability allows for immense flexibility by simultaneously using OPs with different fields, enabling users to easily manipulate specific text snippets such as removing GitHub codes based on their star counts.

Moreover, these OPs establish a one-size-fits-all solution that encompasses a multitude of configurable parameters such as the number of tokens, filtering thresholds, auxiliary models, and much more. 
This adjustability of a single OP, amalgamated with the composability of OP pipelines, empowers \oursys to manage a spectrum of input, output, and processing granularity, contributing to its powerful processing abilities. 

For usage combinations, OPs are labeled with typical usage scenarios. We maintain OP tags as general usage, LaTeX source files, programming codes, financial data processing, or language-specific processing such as English and Chinese, and so on. 
These labels facilitate easy navigation and operation, underscoring our aim to blend simplicity with power in \oursys's architecture.

\section{Feedback-Driven Data Processing}
\label{sec:feedback}

\begin{figure}[!ht]
	\centering
	\subfigure{
		\begin{minipage}[]{0.465\textwidth}
			\centering
			\includegraphics[width=\textwidth]{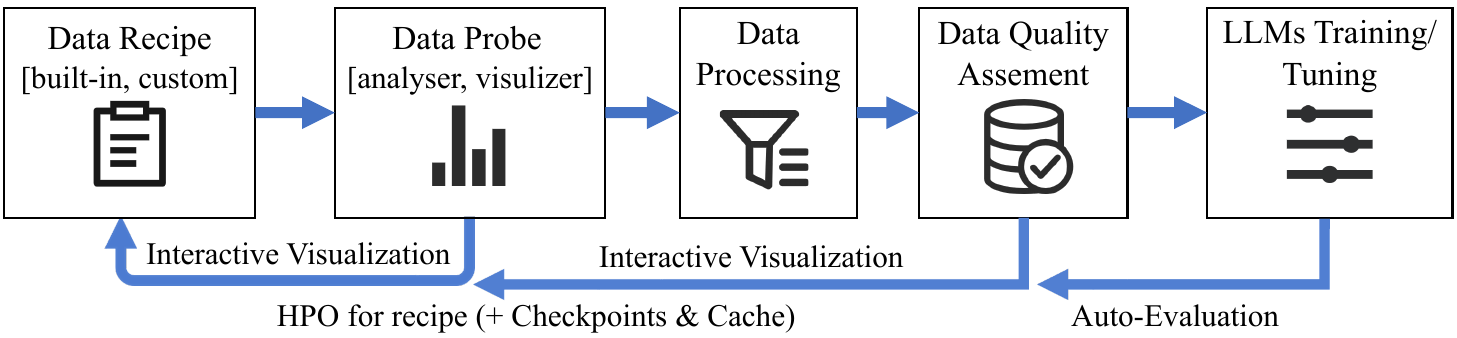}
		\end{minipage}%
	}

	\caption{The feedback loop of \oursys.}

	\label{fig:feedback_loop}
\end{figure}

Addressing Challenge 2 outlined in Sec. \ref{sec:intro}, we incorporate a dynamic feedback loop into the data processing pipeline, which allows users to process and understand data effectively via built-in visualization and automated tracking abilities. 
As demonstrated in Figure \ref{fig:feedback_loop}, our system (\oursys) enables timely perception and swift iterative refinement of data recipes (indicated by the left and upward arrows) within a holistic feedback loop of LLM data processing and LLM training (indicated by the right arrows).

We will discuss the modeling of the data processing feedback in a hyper-parameter optimization (HPO) perspective (Sec. \ref{sec:hpo}), and go through the utility of the interactive visualization (Sec. \ref{sec:interactive-visual}), and the integration of ecosystems for LLM training and evaluations (Sec. \ref{sec:llm-eco}).
The synergy of these techniques offers an efficient and effective solution to debug and dive into LLM data processing.

\subsection{HPO for Data Processing}
\label{sec:hpo}
In \oursys, we incorporate the concept of hyper-parameter optimization (HPO) into the data processing procedure. This is done by tying data-processing-specific hyper-parameters to a variety of feedback signals, including custom target metrics and visualization results. We enhance our system's functionality by innovatively speeding up the data processing iteration through Checkpoint and Caching mechanisms, and by integrating an automated HPO tool.

\twosubsection{Acceleration with Checkpoint and Caching}
LLM data processing often necessitates \textit{frequent re-conduction} due to the  \textit{alterations in OP hyper-parameters} and potential conduction failures, such as exceeding available memory, disk or pre-defined time limits, especially for massive datasets.
Accordingly, we provide built-in checkpoint and caching management to foster resilient and reliable data processing.
Based on a carefully organized directory structure, \oursys automatically monitors every running process for configuration changes, and creates new files to safely store data and processing states only when any error or exception occurs.
While the checkpoint preserves the whole dataset and processing state enabling complete recovery of the processing site, the cache solely saves the dataset object for each OP and is more suited for smaller-scale adjustments as it reduces the overhead of pre-order caches.
These techniques allow for a swift recovery during system restarts or failures, reverting to the most optimal recent processing state stored in the checkpoints, thus mitigating processing redundancy and increasing the feedback frequencies.

Additionally, the proposed state-saving mechanism enables a flexible space-time trade-off at different feedback stages. 
Users have the option to save states after each OP in the data processing flow, ensuring minimal re-execution time at the cost of maximum storage overhead. 
Conversely, they could choose to only save the last OP's checkpoint and cache, incurring minimal storage overhead but increased re-execution time, especially when needing to revert to early steps in the process.

To facilitate a good space-time trade-off, we further perform space complexity analysis for individual OPs, which aids in predicting peak space occupancy and guides us in determining how many checkpoints and caches to store based on available space. 
By default, \oursys actively monitors disk space usage at the start of data processing, and automatically determines if, and when, checkpoints and cache should be deployed. 
User-specified saving frequencies and rules are also supported. 
Consequently, strategic checkpoint and cache management reinforce both the resilience and efficiency of the feedback loop for LLM data processing. 
The detailed space usage analysis can be found in Appendix \ref{appendx:details:space_usage_cache_ckpt}. 

\twosubsection{Auto-HPO} 
We incorporate an automated HPO tool\footnote{\href{https://docs.wandb.ai/guides/sweeps}{W\&B Sweeps, https://docs.wandb.ai/guides/sweeps}} into \oursys to streamline the finding of good data processing hyper-parameters. 
To reduce search costs of different data recipes, we support leveraging advanced HPO algorithms such as Bayesian optimization \cite{shahriari2015taking-bo}, progressive early-stop strategies, such as the Hyperband algorithm \cite{li2017hyperband}, and built-in LLM-oriented sampling strategies (detailed later in Sec. \ref{sec:dedicated_tools}).
Specifically, given a pre-defined target metric and search space of data recipes, users can conveniently explore the impact of specific data processing hyper-parameters.
Here, we give an illustrative example as follows:

\begin{example} {Data Mixing with HPO}
	Suppose we aim to find a good set of sampling weights for $M$ datasets to be mixed, where our search space is defined as $w_i \in [0, 1], i \in [1, M]$.
	The pipeline can be structured as follows:
	\begin{enumerate}[leftmargin=*]
		\item We specify the target text fields across all $M$ datasets, and unify their meta-tags and name of text fields via Formatter OPs.
		\item We leverage meta-tag Filters to cater to specific usage scenarios. Here we only include samples with the language tag ``EN''.
		\item A datasets $\mathcal{D}_{mix}$ is generated from the $M$ datasets, with mixture weights $[w_i]$ drawn by the HPO scheduler to maximize the target metric in step (5).
		\item A pre-configured data processing including de-duplication OPs is executed on the mixed dataset, ensuring dataset cleanness.
		\item The target metric is calculated on $\mathcal{D}_{mix}$ as $(n/N + s)$, where $N$ is the total number of tokens of all $M$ datasets, $n$ and $s$ is the number of tokens and average quality score of $\mathcal{D}_{mix}$ using built-in GPT-3 quality classifier (detailed in Sec. \ref{sec:dedicated_tools}) respectively.
	\end{enumerate}
	The mixture dataset $\mathcal{D}_{mix}$ is iteratively refined by carrying out iterations steps (3)$\sim$(5) to get a larger quantity and better quality. 
\end{example}

\begin{figure}[!ht]
	\centering
	\subfigure{
		\begin{minipage}[]{0.47\textwidth}
			\centering
			\includegraphics[width=\textwidth]{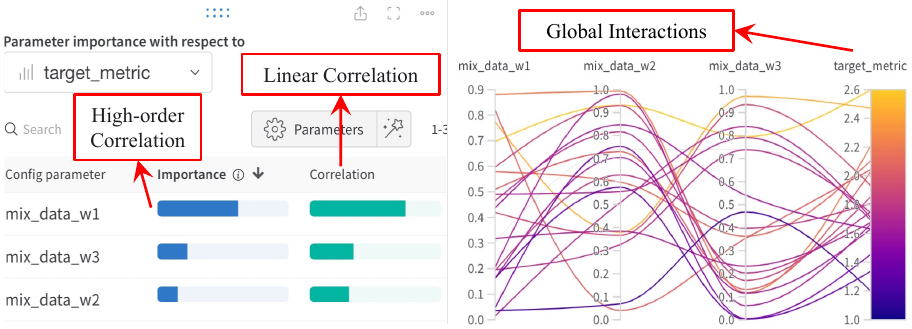}
		\end{minipage}%
	}

	\caption{Demonstration of HPO for data recipe.}

	\label{fig:hpo_demo}
\end{figure}

The HPO results offer a powerful means of visualizing and understanding data insights as shown in Figure \ref{fig:hpo_demo}, where the importance, correlation and interaction of $w_i$ for the quality score are estimated and plotted.
Besides the quality score demonstrated in this example, the target metric can be customized to include other trade-off terms composed of intrinsic data measures – such as toxicity, helpfulness, or other scores predicted by auxiliary models – or even performance measures of LLMs, such as training loss or benchmark scores (which we will discuss later in Sec. \ref{sec:llm-eco}).

\subsection{Interactive Visualization}
\label{sec:interactive-visual}
\begin{figure}[!t]

	\centering
	\begin{minipage}[b]{.56\linewidth}
		\subfigure[Tracking Specific Data Samples]{\includegraphics[width=1\linewidth]{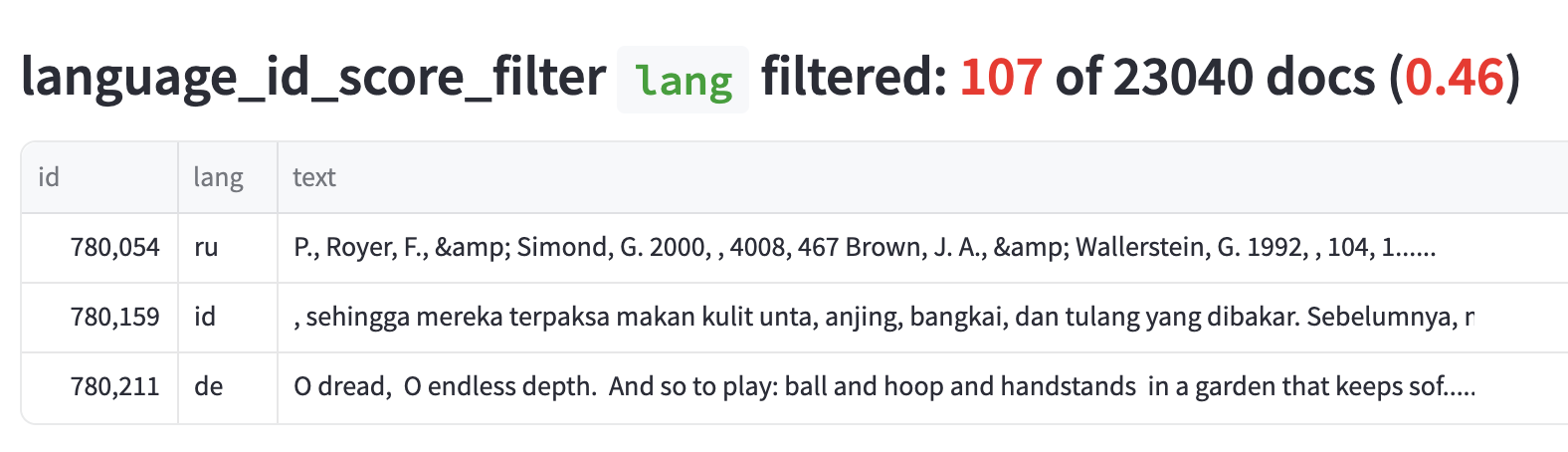}}%
		\\
		\subfigure[Effect of OP Pipeline (Number of Samples)]{\includegraphics[width=1\linewidth]{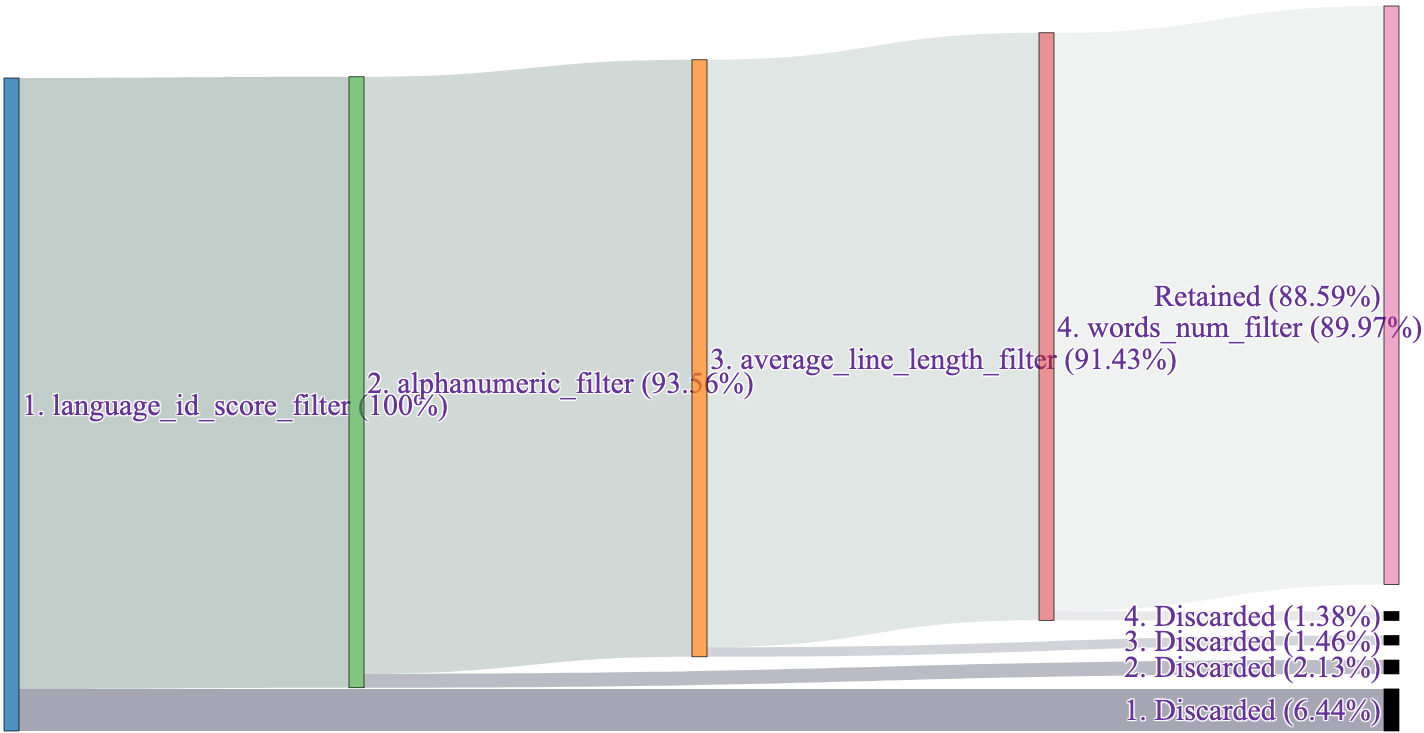}}
	\end{minipage}%
	\hfil%
	\begin{minipage}[b]{.38\linewidth}
		\subfigure[Data Distribution Diff.]{\includegraphics[width=1\linewidth]{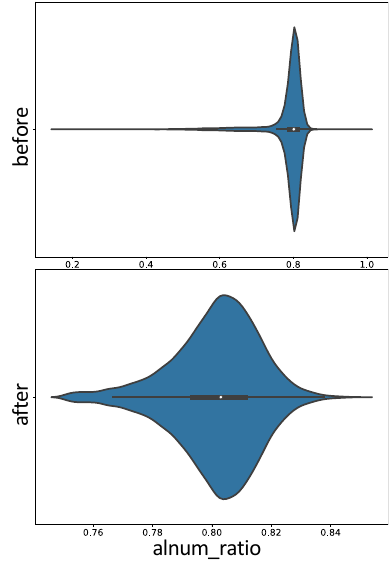}}%
	\end{minipage}%

	\caption{The illustration of interactive visualization of \oursys. More demos are publicly available.}

	\label{fig:interactive_visual}
\end{figure}

The ability of interactive visualization is integral to multiple feedback stages of \oursys. 
Specifically, as Figure \ref{fig:interactive_visual}.(a) demonstrates, users can visually track the effects of individual OPs in terms of the processed data samples.
This is facilitated by an innovative built-in tool, \textbf{tracer}, which records sample changes after applying each operation for \oursys. 
For example, tracer presents discarded samples for Filters, pre- and post-editing differences for Mappers, and (near-) duplicate sample pairs for Deduplicators. 
Coupling this tracking ability with fruitful built-in sampling and visualization tools, \oursys enhances users' control over the data processing and boosts their confidence and rationals of the process.

Transitioning to the mid-term stage of LLM data processing, \oursys offers a comparative visualization of the data before and after the entire processing from the view of OP pipeline and statistical analysis, as Figures \ref{fig:interactive_visual}.(b) and \ref{fig:interactive_visual}.(c) show. 
Aided by a built-in tool, \textbf{analyzer}, \oursys provides statistical analysis (counts, means, standard deviations, min/max, quantiles, entropy, etc.) to allow a deep understanding of the data. 
By default, the summary of per-sample statistics covers 13 dimensions and automatically displays histograms and box plots for each statistical variable, including diverse criteria like sample perplexity, word count, flagged word percentage, and paragraph length, among others. 
Users also have the flexibility to adjust the dimensions to observe, with a bespoke visualization and data processing experience.

\subsection{Feedback with Integrated LLM Libraries}
\label{sec:llm-eco}
In the later stages of our pipeline, we utilize robust ecosystems designed for LLM training and evaluation, ensuring seamless integration with widely-used libraries such as Megatron-LM \cite{shoeybi2020megatronlm}, DeepSpeed \cite{Rasley-KDD-2020-DeepSpeed}, and HuggingFace's Transformers \cite{Wolf-EMNLP-2020-Transformers}.
With this integration, users can easily train LLMs on datasets produced by \oursys and evaluate their performance to obtain feedback using our pre-built tools and scripts, without getting bogged down in complicated LLM training and evaluation details.

Notably, our system facilitates the timely assessment of model abilities by incorporating multiple dimensions. 
The system's capability to swiftly identify potentially ineffective data and training allows us to terminate unwanted LLM data processing promptly.
Instead of solely relying on model loss as the basis for evaluating model performance, we support the LLM assessment across various metrics or benchmarks, and track shifts in target scores.
Consequently, we can determine whether continued training of an LLM on the produced dataset is justified, thereby helping us minimize data processing and LLM training costs.

Specifically, \oursys's evaluator supports SOTA LLM benchmarks such as HELM \cite{liang2022holistic-helm}, LM-harness \cite{eval-harness} and GPT-API-based evaluation \cite{vicuna2023}, as well as the extension of customized evaluation benchmarks and tasks. 
For a balanced and straightforward evaluation, \oursys supports a leaderboard-style comparison by consolidating results from different target evaluation scenarios, such as ranking averaging, score-normalized averaging, or other customized strategies.
The leaderboard-style scoring utility enhances the visualization of strengths and weaknesses of models, guiding subsequent iterations of data recipes and LLMs' refinements.
We also make available \textit{Reference Models} - these are model checkpoints binding with traceable training data in \oursys, popular LLM architectures, training parameters, computation costs, and corresponding evaluation results.
They facilitate effortless comparison among different training configurations, particularly for further research on diverse, iteratively developed data recipes.

\subsection{Feedback Loop Showcase}
\label{sec:feedback_case_study}

\begin{figure*}[!ht]
	\centering
	\includegraphics[width=0.98\textwidth]{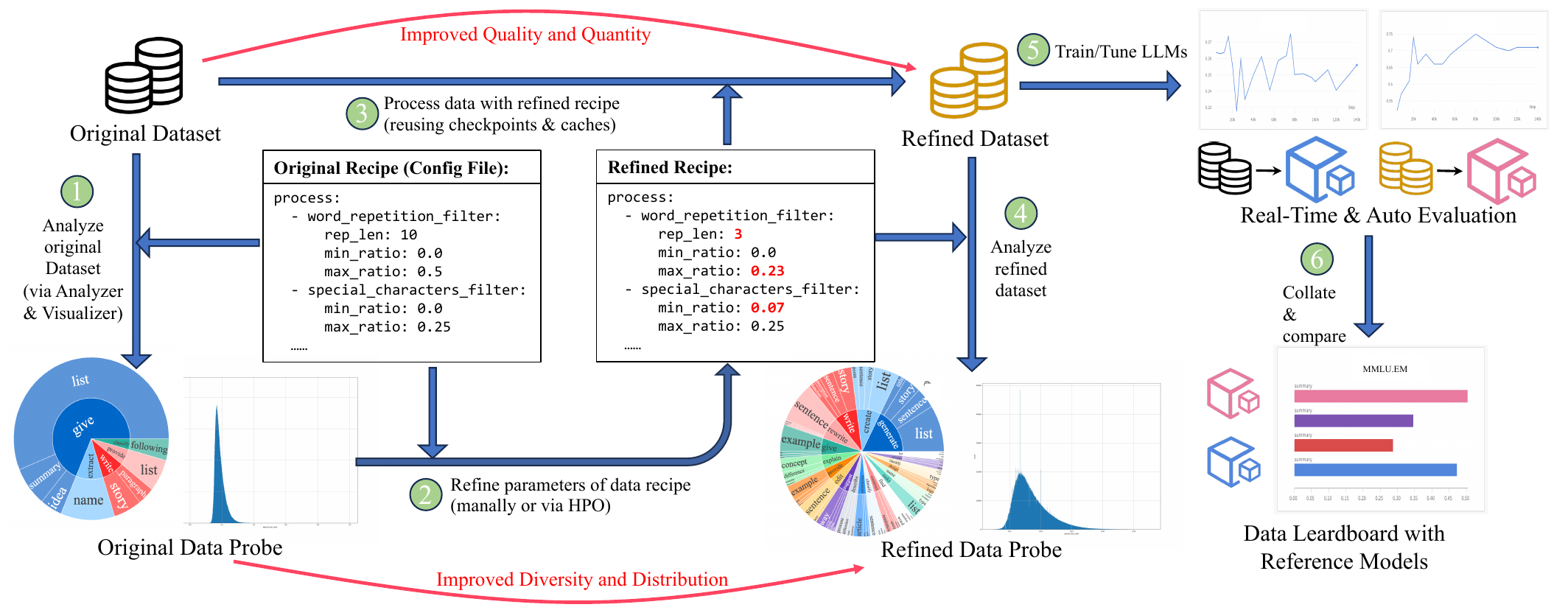}
	\caption{The demonstration of data processing feedback of \oursys, which helps to generate better data recipes for LLMs.}
	\label{fig:case_feedback_loop}
\end{figure*}

The general feedback loop has been discussed before in Figure \ref{fig:feedback_loop}. 
We now further expound on this by presenting a concrete development example. 
Here, we intertwine several previously mentioned tools to demonstrate the Data-in-the-LLMdev-Loop process, which results in improved LLM data.
As illustrated in Figure \ref{fig:case_feedback_loop}, we begin with a raw dataset and aim to refine it for better pre-training or fine-tuning of an LLM. The entire process flows as per the following steps:

(1) \textbf{Analyze the original dataset.} 
We can opt to utilize an existing data recipe (a specific configuration file) or craft a new one based on prior understandings of data processing needs. 
Our built-in Analyzer and Visualizer facilitate this process by computing more than a dozen measures such as linguistic diversity, textual statistics, and others to generate a data probe. The two pie plots within Figure \ref{fig:case_feedback_loop} indicate the top 20 most common root verbs (inner circle) and their top 4 direct noun objects (outer circle) for the data in field \textit{``text.instructions''}.

(2) \textbf{Refine parameters of the original recipe.} 
Based on the data probe, users figure out the weaknesses of the original dataset, such as low diversity in expression manners, and long-tail statistics of word counts. Then we refine the parameters in the recipe by adding/removing some OPs or tightening/relaxing filter ranges. 
During refining, we could find out the effect of each OP easily based on the interactive visualization tool mentioned in Sec. \ref{sec:interactive-visual}.

(3) \textbf{Process the original dataset with the refined recipe.} Then we process the original dataset with the refined recipe using \oursys and get a refined dataset and several saved checkpoints for further adjustments. This step can be facilitated with the help of our cache and checkpoint mechanisms.

(4) \textbf{Analyze the refined dataset.}
Like step (1), we analyze the refined dataset again to obtain a new data probe. 
Based on the statistics and visualization results, we assess the degree of improvement in the data quality. 
If the refined data fails to meet our expectations, we revert to step 2 to manually adjust the data recipe or employ our HPO tool for automatic refinement (refer Sec. \ref{sec:hpo}).

(5)  \textbf{Get LLMs with the refined dataset.} Then we can train or fine-tune LLMs with the refined dataset and training frameworks integrated into \oursys (Sec. \ref{sec:llm-eco}). 
During the training or fine-tuning process, our auto-evaluation tools offer timely, multi-view assessments of LLMs. 
These tools inspect numerous metrics across multiple evaluation datasets. 
This feature provides us the advantage of halting the process prematurely if the refined data weakens LLM performance, thereby preventing unnecessary costs.

(6)  \textbf{Collate results and compare with reference models.} 
Finally, \oursys automatically collates the evaluation results and compares them with reference models in the data leaderboard, providing a clear representation of the effects of data processing alone. Consequently, we derive either a superior LLM, which can be auto-registered as a reference model, or additional refining guidance from the LLM perspective to further enhance data recipes.

\section{Boosting Usability with Built-ins}
\label{sec:recipes_plugins}
In response to the challenge of varied user customized preferences and technical expertise (Challenge 3 in Sec. \ref{sec:intro}), we offer an easy-to-use configuration paradigm for data recipes, ready-to-use data recipe templates, and extensive tools, as detailed below.

\subsection{Configuring Your Data Recipe}
\label{sec:recipes}
Notably, we make the end-to-end pipeline of data processing configurable in \oursys, including specified processing environment parameters, OP lists, tools used, and so on. 
This principle of all-in-one configuration ensures reproducibility and traceability, and simplifies changing specifications in data processing, thereby facilitating the formation of data recipes for further refinement and reuse, and enabling the quantitative exploration and automatic optimization of data processing (Sec. \ref{sec:hpo}).

Specifically, built upon Jsonargparse \cite{jsonargparse}, we provide unified, flexible, easy-to-use and powerful configuration capabilities.
It is engineered to automatically register configuration items for OPs and tools, and accept varying sources of configurations such as command line entries, yaml and jsonnet files, environment variables, default hard-coded values, and a mixture of those for convenient incremental modifications.

For example, users can easily build up their own config files by two recommended methodologies—``subtraction'' or ``addition''. 
The ``subtraction'' approach utilizes a pre-set configuration file containing \textbf{all available OPs, tools, and their default parameters}. 
Users can simply remove or re-order these OPs and adjust these parameters per their requirements. 
Conversely, the ``addition'' approach lets users build their configuration files from scratch, leveraging our extensive examples of pre-built data processing recipes for totally \textbf{more than 20 high-quality and diverse data recipes for pre-training, fine-tuning, English, Chinese, etc.}
More quantitative analysis on certain recipes are in our experiments (Sec. \ref{exp:recipe_quality}).

\subsection{Dedicated Pluggable Tools}
\label{sec:dedicated_tools}
To further enhance usability, facilitate system customization and augment users' data handling capabilities, \oursys includes an extensible collection of powerful dedicated tools that can be conveniently plugged into different stages of the LLM data processing.

\para{Quality Classifier.}
As an illustrative example, we describe our text quality classifier for culling high-quality text from heterogeneous data sources like CommonCrawl. This tool is a reproduced model based on the closed-source GPT-3 quality scorer \cite{brown2020gpt3}. 
Moreover, we have expanded its applicability to Chinese text and various code types. Encapsulated as a callable pipeline, this tool provides users with the freedom to adapt it to other various scenarios. 

The functionality of the classifier is backed by PySpark's standard Tokenizer or Sentencepiece model \cite{Kudo-EMNLP-2018-SentencePiece}, along with HashingTF as the feature extractor. 
It then applies a binary logistic regression classifier to gauge the quality of a text. 
We provide more empirical verification of them in Sec. \ref{exp:quality-classifier}.

\para{Enhanced Sampler for LLM data.} 
In \oursys, we have designed several advanced data sampling utilities specialized for large-scale data chunk handling in LLMs. 
Our solutions effectively streamline representative extraction, optimize processing time and resources, and meet the distinctive needs of LLM developers.

Our stratified sampling technique is noteworthy in this LLM data context. 
It capitalizes on information within the metadata or statistical fields, thus accommodating varied selection metrics in crafting an effective data sample. 
To ensure a comprehensive yet flexible representation of the data corpus, we consider various heterogeneous criteria such as document length, token count, the frequency of boolean predicates for post-conditional checks, and even linguistic diversity formulated via occurrences of verb-noun pair (as shown in the pie plots in Figure \ref{fig:feedback_loop}) . These dynamic criteria are tailored to distinct analytic needs and promote efficient data processing, seamlessly integrating with downstream OPs and tools.

\para{Full Toolkit.}
As for other tools, readers can refer to Sec. \ref{sec:feedback} for an examination of multiple previously discussed tools, including \textbf{tracer} and \textbf{analyzer} (Sec. \ref{sec:interactive-visual}), and \textbf{evaluator} and \textbf{reference models} (Sec. \ref{sec:llm-eco}). 
We diligently maintain and evolve the toolkit in \oursys, and make the full set publicly accessible.

\subsection{User-Friendly Experiences in \oursys}

\oursys is designed not just for functionality but also for adaptability, catering to an extensive user base with diverse expertise and skill sets. While abstracting the intricate system internals, we provide user-friendly interfaces and extensive customizable components.  Accordingly, users can embark on zero-code data processing, engage in low-code customization, or delve into in-depth extensions for complex requirements. 

\begin{itemize}[leftmargin=*]
	\item \textbf{Zero-Code Processing:} For novice users, \oursys supplies a series of ready-to-use data recipes and plug-in tools for immediate use. This requires no knowledge of advanced system architectures or OPs, as discussed in Sec. \ref{sec:recipes} and Sec. \ref{sec:dedicated_tools}.
	
	\item \textbf{Low-Code Customization:} Intermediate users can enjoy the flexibility to alter configurations, data, and external resources to suit their specific needs. They can readily reuse, combine, and edit built-in data configurations; customize quality classifiers and tokenizers; refine data based on our pre-developed recipes; or provide fresh links to auxiliary models or vocabularies from our unified, routinely updated public cloud drive. 
	
	\item \textbf{Advanced Extension:} Experienced users are allowed to easily introduce new OPs by deriving from base classes and implementing their specific ``process()'' and ``compute\_stats()'' functions, as demonstrated in the code Listing \ref{listing:base_op_classes}. This grants the users an end-to-end view of the process for a single sample, while \oursys handles the nitty-gritty of configuration registration and efficiency optimization.

\end{itemize}

Additionally, \oursys's decoupled design facilitates the smooth incorporation of new tools for users at all stages of LLM data processing, ranging from novel visualization dimensions and evaluation datasets to pre- or post-processing scripts. 

To enhance the ease of adoption and use of \oursys, apart from the numerous pre-built data recipes (refer Sec. \ref{sec:recipes_plugins}), we also provide a series of interactive demos, implemented in Streamlit, for varied profiles and scenarios. 
This hands-on learning approach has been designed to enable users of varying skill levels to quickly familiarize themselves with and effectively use \oursys.

\section{Comprehensive System Optimization}
\label{sec:sys_optimization}
To handle large-scale data (Challenge 4 in Sec. \ref{sec:intro}), we employ a series of optimizations in \oursys from various aspects.

\para{Optimized Computation: Context management, Operator (OP) Fusion and Reordering.} 
To elevate computational efficiency in LLM data processing, we provide advanced context management, operator fusion, and operator reordering techniques for nuanced implementation contributions.
The manager meticulously handles shared intermediate variables, such as segmented words, split lines, and others derived from the original textual corpus, across different operators. It allows seamless reuse of these context variables across multiple operators, thereby mitigating the necessity for computationally expensive re-evaluations.

Based on the context manager, the proposed operator fusion method is another new contribution to the field. We propose to identify fusible operators that either share the same contexts or computation sub-procedures. It detects the OP groups first. Successive OPs in the same group should be commutative with each other. It then amalgamates identified fusible operators in each group into a single fused OP, enabling them to be executed faster with a larger localized perspective. The contexts of each sample will be cleaned up after each fused OP, hence little extra memory is required for context management and operator fusion.

Due to the time-consuming increase of single fused OP, we further design a strategy of operator reordering to optimize the execution sequence of the OP list after fusion. 
For example, based on the commutativity of Filters, we delay the running of time-consuming OPs (such as fused Filters) and prioritize other less time-consuming OPs. 
As a result, these time-consuming OPs only need to handle fewer samples because the preceding operators have filtered out some of them, enhancing overall computational efficiency.

\begin{figure}[!ht]
	\centering
	\subfigure{
		\begin{minipage}[]{0.45\textwidth}
			\centering
			\includegraphics[width=\textwidth]{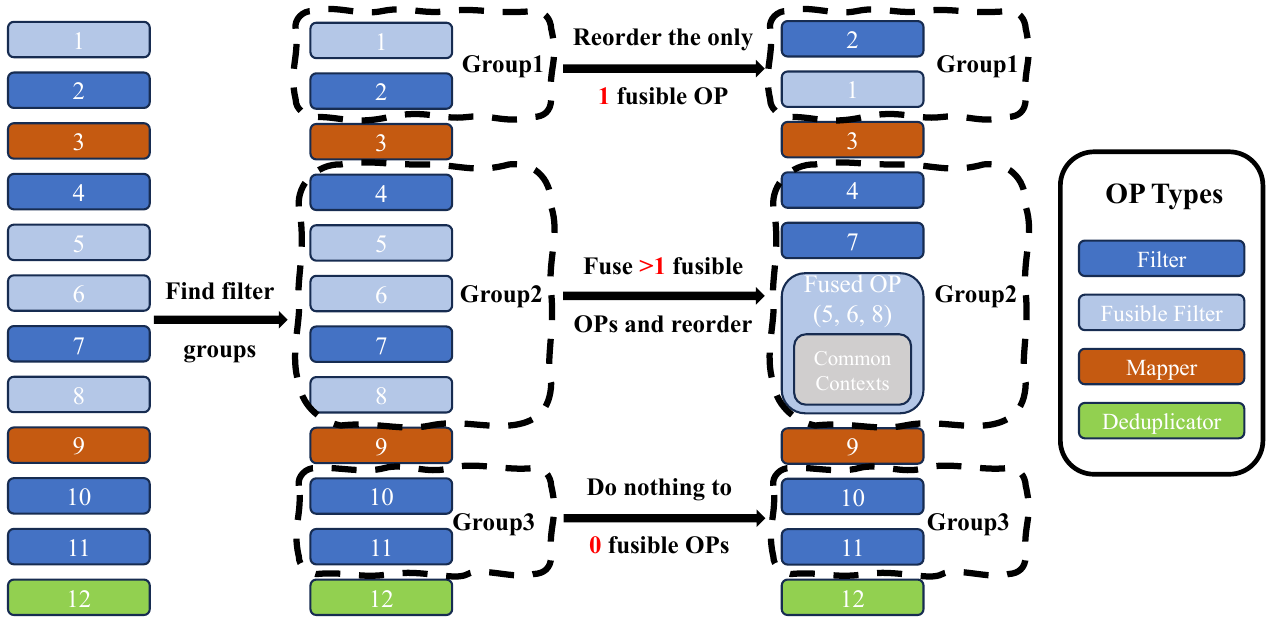}
		\end{minipage}%
	}
	\caption{The OP fusion procedure for an OP list.}
	\label{fig:procedure_of_op_fusion}
\end{figure}

The whole procedure of OP fusion is summarized in Figure \ref{fig:procedure_of_op_fusion}. 
These amalgamation strategies serve dual purposes. 
Firstly, it minimizes redundant computation, eliminating the need for repetitive yet shared computations. 
Secondly, it mitigates the overhead of initializing multiple processes by reducing the total count of processing OPs, thus maintaining expeditious data processing routines.

\para{Optimized Space Utilization: Caching OPs and Compression.} 
Recognizing the inadequacies of the original cache management protocol in the \pacterm{Huggingface-datasets} library, especially pertaining to the handling of non-serializable third-party models and functions in certain OPs, we design a dedicated hashing method to bypass the serialization procedures of those non-serializable objects, which ensures successful caching of each OP and permits \oursys to leverage optimal cache management.

Furthermore, we incorporated the ability for users to activate advanced compression technologies, such as Zstandard (zstd) \cite{zstd-rfc8878} and LZ4 \cite{lz4}, in \oursys. 
It will automatically compress cache files after each OP and decompress these compressed files back to normal cache files when rerunning this OP in the same configuration. Compared with the processing time, compressing/decompressing time is relatively negligible due to the high efficiency of the compression technologies mentioned above. This feature substantially reduces the volume of cache data storage, facilitating the processing of larger datasets without compromising speed or stability.

\para{Optimized Scalability: Distributed Data Processing.} 
The volume of LLM training data can be extremely large, making it difficult to be processed with a single machine.
\oursys meshes with distributed processing frameworks such as \pacterm{Ray} \cite{ray}, Apache \pacterm{Beam} \cite{apache-beam} and Apache \pacterm{Flink} \cite{carbone2015apache-flink}, and offers the ability to seamlessly translate a data processing pipeline running on a single node into a multi-node cluster.
In this way, resources in cluster computing can be utilized to accelerate data processing and generation.

Specifically, we adapt the underlying interfaces of \pacterm{HuggingFace}-datasets for those of \pacterm{Ray}-datasets, such that all OPs of \oursys, even when written as single-machine Python functions, can be executed in a distributed mode with the help of automatic data partitioning by \pacterm{Ray}.
An alternative approach we support is to replace the default \pacterm{Ray} runner of \oursys with other distributed processing back-ends such as \pacterm{Flink}, via pre-translations from \oursys's processing pipelines into the \pacterm{Beam}-compatible ones.
As a result, almost all the OPs within \oursys (Mapper, Filter, and Deduplicator) can be accelerated in a multi-node cluster, and effectively alleviate the bottlenecks on a single node (even with process-based parallelism) caused by memory capacity and IO throughput.
More empirical results can be found in Sec. \ref{exp:scalability-main}.

In a nutshell, all of these optimizations enhance \oursys's scalability from various views, to handle the vast amount of data involved in LLMs, ensuring robust and efficient processing while minimizing resource requirements.

\section{Evaluation of \oursys}
\label{sec:eval}

\subsection{Making Better Data Recipes}
\label{exp:recipe_quality}

The value of an effective LLM data processing system is reflected not only in its comprehensive and flexible operability but also in its capacity to produce high-quality data that LLMs can more readily ``digest''.
\oursys provides specialized features for exploring and making data recipes tailored to LLMs, and we have developed numerous ready-to-use data recipes using \oursys. 
In this section, we evaluate the quality of data recipes generated by \oursys for both LLM pre-training and fine-tuning.

\twosubsection{Refined Pre-training Data Recipes} 
The pre-training data we produced consists solely of publicly available sources, exemplifying the core principles of transparency and reproducibility. 
Specifically, we choose to improve two widely-used, high-quality datasets for LLMs, TogetherAI's RedPajama \cite{together2023redpajama} and EleutherAI's Pile \cite{Gao-arxiv-2021-Pile}, which were curated from 15 highly diverse text sources and subjected to meticulous pre-processing and cleaning to ensure their quality.
With the help of \oursys , we further refine them via data analysis, merging and quality enhancement, employing dozens of OPs with varied configurations.
For detailed statistics, processing steps and refined data recipes, please refer to Appendix \ref{appendx:data-recipe}.

\begin{figure}[!h]
	\centering
	\subfigure{
		\begin{minipage}[]{0.45\textwidth}
			\centering
			\includegraphics[width=\textwidth]{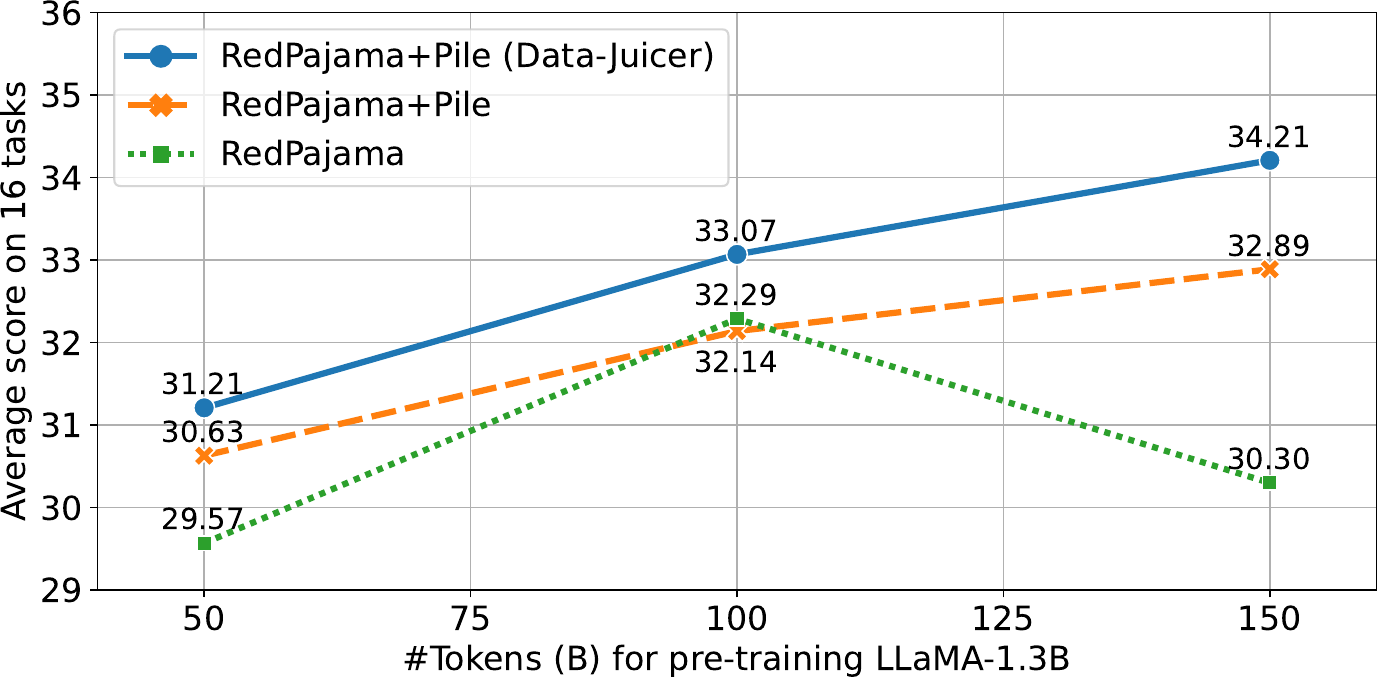}
		\end{minipage}%
	}
	\caption{Evaluation results of reference models trained with different datasets but the same pre-training procedures. 
		\oursys's data recipe gains consistent improvements over baselines.
	}
	\label{fig:pretrain_exps}
\end{figure}

To verify the quality of the data recipes derived by \oursys, we use the original RedPajam and Pile, and our refined datasets to pre-train LLMs with mainstream LLaMA architecture and assess the models' performance across 16 core HELM tasks. 
We keep the training configurations the same while only modifying the training data. Detailed hyper-parameters are in Appendix \ref{appendx:imp-details:pretrain}.
The results of average scores of 16 tasks are visualized in Figure \ref{fig:pretrain_exps}, where we evaluated checkpoints throughout the pre-training process with an increasing number of billion-sized tokens at 50B, 100B, and 150B. 
Notably, through fair comparisons with equivalent training tokens, LLMs pre-trained on \oursys-recipes consistently outperformed those using only RedPajama or its union with the Pile, reinforcing the usefulness and effectiveness of \oursys.

Moreover, we compare \oursys-models with several SOTA baselines and summarize the results in Table \ref{tab:pre-train-res}. 
With only half the data volume (150B tokens), LLaMA-1.3B pre-trained on \oursys-recipe outperformed Pythia-1.4B \cite{Biderman-arxiv-2023-Pythia} (300B tokens), and even beats highly competitive Falcon-1.3B \cite{Penedo-2023-arxiv-Refinedweb} trained on 350B tokens. 
Notably, we further labeled 17 subsets from Alpaca-CoT (a collection of 39 public fine-tuning datasets) with the ``Instruct Fine-Tuning (IFT)'' tag and performed data mixing and processing using \oursys. 
Following the usual practice \cite{Zeng-arxiv-2022-GLM}, we incorporate these large-volume IFT data into the pre-training phase and execute continuous training upon the checkpoint of \oursys (RedPajama+Pile)-150B. 
As reflected in the last two rows of Table \ref{tab:pre-train-res}, \oursys gains a further $4.9$\% relative improvement over the original Alpaca-CoT-IFT while utilizing only $\sim$30\%  data volume.

\begin{table}[!hbt]
	\centering
	\caption{The average score of the pre-trained LLMs on the 16 HELM core tasks. Individual task results and data recipes are detailed in Appendix \ref{appendx:per-task-res}. ``IFT'' denotes the datasets tagged with ``Instruct Fine-Tuning'' in our context.}
	\label{tab:pre-train-res}
	\begin{tblr}{
			row{4} = {m},
			cell{1}{4} = {c},
			cell{2}{4} = {c},
			cell{3}{4} = {c},
			cell{4}{1} = {r=3}{},
			cell{5}{4} = {c},
			cell{6}{4} = {c},
			hline{1,7} = {-}{0.08em},
			hline{2,4} = {-}{},
			hline{5} = {2-4}{},
		}
		\textbf{Model} & \textbf{Training Data} & \textbf{\#Tokens} & \textbf{Score}\\
		Falcon-1.3B \cite{falcon-hf} & RefinedWeb & 350B & 33.97\\
		Pythia-1.4B \cite{pythia-hf} & Pile & 300B & 33.96\\
		LLaMA-1.3B & {\textbf{\oursys}\\\textbf{(RedPajama+Pile)}} & 150B & \textbf{34.21}\\
		& + Alpaca-CoT-IFT & 150B + 15B & 35.04\\
		& \textbf{+ Our Refined IFT} & 150B + 4.7B & \textbf{36.76}
	\end{tblr}
\end{table}

Taken together, these findings underscore the potential of the \oursys system to generate high-quality data and verify the excellence of \oursys-recipes in terms of enhancing LLM performance while reducing LLM training costs.

\twosubsection{Refined Fine-tuning Data Recipes}
For the Alpaca-CoT collection, besides the ``IFT'' tag as validated in Table \ref{tab:pre-train-res}, we also labeled datasets within it with ``Chat Fine-Tuning (CFT)'' for enhanced dialog ability and aligned human value. 
To examine their quality, we first use the CFT and EN tags to filter out several competitive subsets, and then generate two new equal-size datasets by random sampling and our designed recipe respectively. 
Then we conduct fine-tuning on the generated datasets based on the open-source mainstream architecture, English LLaMA-7B \cite{openlm2023openllama}. 
Similarly, we replace the tag ``EN'' with ``ZH'', and use a SOTA LLaMA-2-7B variant \cite{flagalpha-llama} for the Chinese scenario.
Statistics of these datasets and training hyper-parameters are in Appendix \ref{appendx:imp-details:posttune}.

For a thorough and comparative performance evaluation, we used GPT-4 API for pairwise scoring and tallying of wins and ties. 
The results are consolidated in Table \ref{tab:post-tune-res}, from which we can see that LLMs utilizing \oursys-recipes consistently demonstrate high validity. 
Firstly, compared to LLMs trained on the competitive fine-tuning open datasets, Alpaca \cite{Taori-github-2023-Stanford-alpaca} and Belle \cite{BELLE}, LLMs trained on \oursys data gain higher win rates (up to $17.5$\% for English case) while using less data (up to 90.4\% reduction for Chinese case).
Secondly, compared to the LLMs trained on the datasets with trivial processing strategy (mixture by random sampling), LLMs trained on \oursys still gain higher win rates (up to $14.4$\% ), which attests to the effectiveness of our enhanced sampling strategy and quality of \oursys-recipes for LLMs again. 

\begin{table}
	\centering
	\caption{Results of pair-wise model comparisons using GPT4 scoring. ``CFT'', ``EN'' and ``ZH'' indicate meta-tags as Chat Fine-Tuning, English, and Chinese text respectively.}
	\label{tab:post-tune-res}
	\begin{tblr}{
			width = \linewidth,
			colspec = {Q[226]Q[305]Q[132]Q[63]Q[62]},
			column{3} = {c},
			column{4} = {c},
			column{5} = {c},
			cell{2}{1} = {r=4}{},
			cell{2}{5} = {r=2}{},
			cell{4}{5} = {r=2}{},
			cell{6}{1} = {r=4}{},
			cell{6}{5} = {r=2}{},
			cell{8}{5} = {r=2}{},
			hline{1,10} = {-}{0.08em},
			hline{2,6} = {-}{},
			hline{4,8} = {2-5}{},
		}
		\textbf{Model}                      & \textbf{Tuning Data}            & \textbf{\#Samples} & \textbf{Win} & \textbf{Tie} \\
		LLaMA-7B \cite{openlm2023openllama}     & Alpaca                          & 52k                & 16           & 100           \\
		& \textbf{\oursys} & 40k                & \textbf{44}  &              \\
		& Random (\textbf{CFT, EN})       & 40k                & 19           & 105           \\
		& \textbf{\oursys} & 40k                & \textbf{36}  &              \\
		{LLaMA2-7B\\(Chinese, \\ FlagAlpha \cite{flagalpha-llama})} & Belle                           & 543k                & 28           & 99           \\
		& \textbf{\oursys} & 52k                & \textbf{33}  &              \\
		& Random (\textbf{CFT, ZH})       & 52k                & 19           & 96           \\
		& \textbf{\oursys} & 52k                & \textbf{45}  &              
	\end{tblr}
\end{table}

\subsection{Processing Data Efficiently and Effectively}
\label{exp:system_perf}

\twosubsection{End-to-End System Performance}
\label{exp:end2end_perf}

To evaluate the processing performance of \oursys, we compare it with two SOTA baselines: TogetherAI's RedPajama \cite{together2023redpajama} and AllenAI's Dolma \cite{DolmaToolkit}. A more detailed introduction to and comparison with these baselines can be found in Appendix \ref{appendx:sys_baselines}.
For a fair comparison, here we use their official code repositories and run \oursys on the data recipes with the same OPs to process the Books, arXiv, and C4 datasets, which vary in terms of data sizes, distributions and involve diverse processing OPs.

We conduct multiple rounds of experiments on different numbers of processes (np=[32, 64, 128]) and monitor several core metrics, including processing time and average memory usage. The monitored time is the wall-clock time of the whole processing pipeline. 
The average memory usage is monitored every second and aggregated across all relevant processes.
For more experimental details, please refer to Appendix \ref{appendx:imp-details:sys_perf}.

\begin{figure}[!ht]
	\centering
	\subfigure{
		\begin{minipage}[]{0.47\textwidth}
			\centering
			\includegraphics[width=\textwidth]{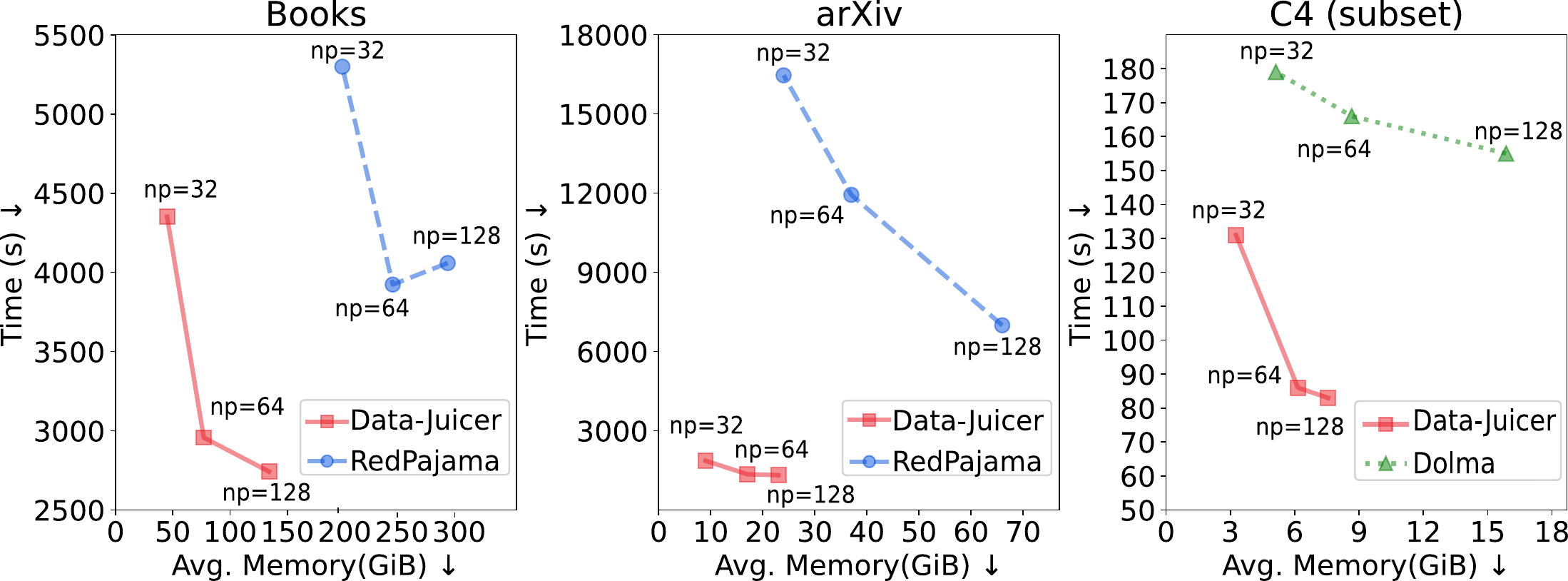}
		\end{minipage}%
	}

	\caption{Comparison of stand-alone performance in various data sizes and processing configurations.}
	\label{fig:stand-alone}
\end{figure}

The experimental results are summarized in Figure \ref{fig:stand-alone}. 
Notably, for all datasets and various numbers of processes, \oursys requires an average of 50.6\% less processing time and 55.1\% less memory. 
In particular, it saves at most 88.7\% processing time for the arXiv dataset compared with the baseline. 
Also, it takes up to only 22.9\% memory of baseline for \oursys to process the Books dataset, which is mainly because the processing procedure of the baseline loads the whole dataset at once. 
Overall, \oursys effectively alleviates the bottleneck caused by IO of cache files, and achieves better end-to-end time-space efficiency than baselines.

\twosubsection{Effect of Context Management, OP Fusion, and Reordering}
\label{appendx:op-fusion}
As introduced in Sec. \ref{sec:sys_optimization}, \oursys employs dedicated optimization to minimize redundant computations and save processing time. 
To examine the optimization effect, we prepared three test datasets of varied sizes and sample counts. Each dataset goes through the same processing recipe which includes 14 OPs (5 Mappers, 8 Filters, and 1 Deduplicator), with 5 of these OPs being fuse-able. 
We conduct comparison experiments with 4 processes, except for the largest dataset, where we utilize 50 processes to assess if these techniques remain effective on larger scales.

\begin{figure}[!ht]
	\centering
	\subfigure{
		\begin{minipage}[]{0.45\textwidth}
			\centering
			\includegraphics[width=\textwidth]{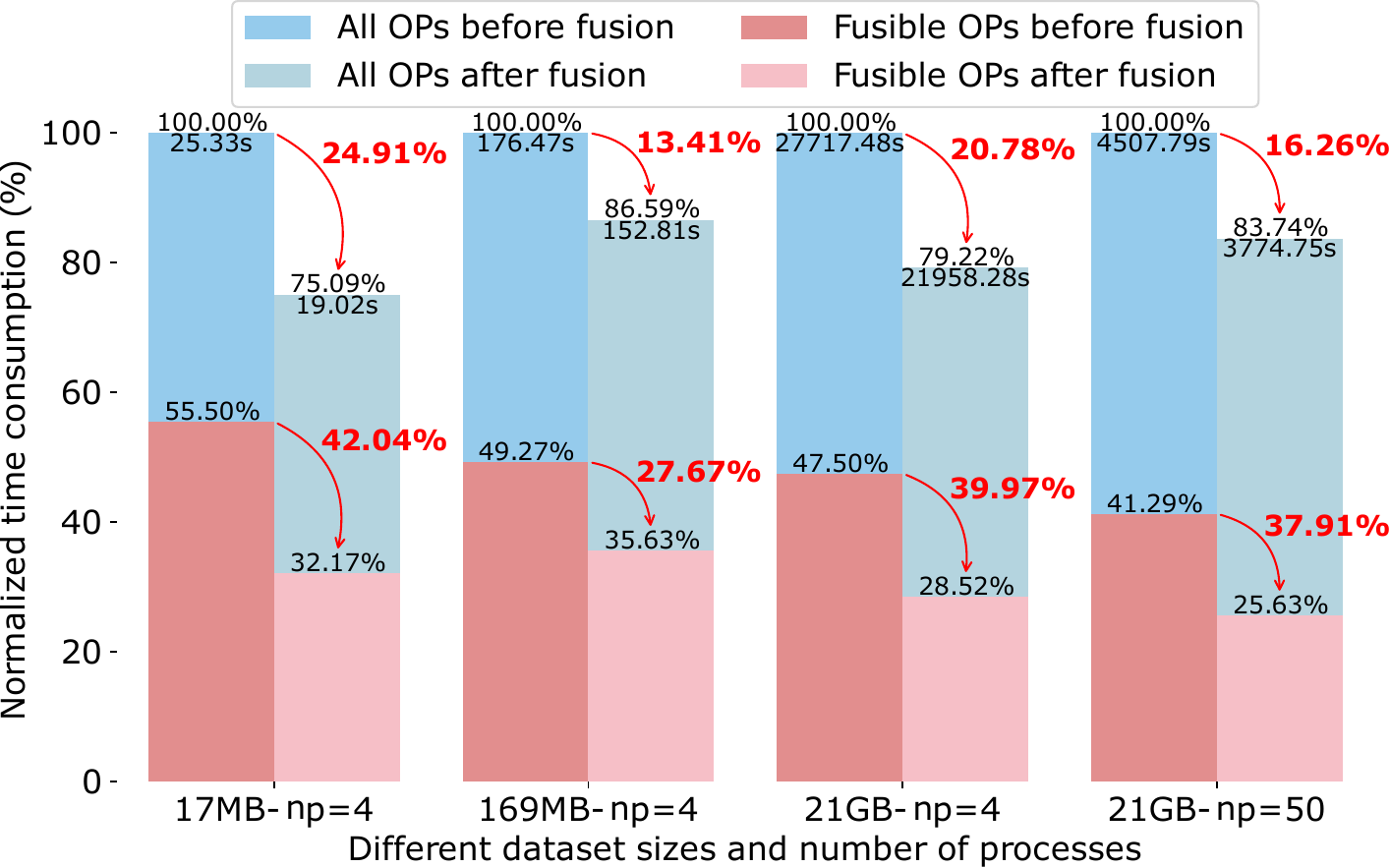}
		\end{minipage}%
	}
	\caption{Time comparison before and after OP fusion.}
	\label{fig:effect_of_op_fusion}
\end{figure}

The results are shown in Figure \ref{fig:effect_of_op_fusion}, where both the normalized and actual time consumption for each experimental setup are indicated. The results signify that our optimization strategy effectively saves up to 24.91\% of the total time for the entire process and saves at most 42.04\% of time for those fusible OPs. In addition, the findings showcase that the optimization performs efficiently regardless of variations in dataset sizes or the number of processes utilized.

\twosubsection{Effect of Quality Classifiers}
\label{exp:quality-classifier}
As described in Section \ref{sec:dedicated_tools}, \oursys provides built-in quality classifiers for LLM data processing, and here we present several empirical results regarding their performance. Specifically, we follow the training procedure of  the proprietary quality classifier used in GPT-3 \cite{brown2020gpt3} and extend its training pipeline to include Chinese text. 
In the evaluation of the collected data, we found that our reimplementation of the GPT-3 classifier and its Chinese adaptation achieved F1 scores of 97.47\% and 98.64\%, respectively. Further training and evaluation details are provided in the Appendix \ref{appendx:quality_score}.

\begin{table}[!ht]
	\centering
	\caption{Comparison of keeping ratio on CommonCrawl.}
	\label{tab:qc_keep_ratio}
	\begin{tblr}{
			width = \linewidth,
			colspec = {Q[20]Q[20]Q[20]},
			cells = {c},
			hline{1,5} = {-}{0.08em},
		}
		\textbf{ Quality Classifier} & \textbf{ Keeping Ratio @ label } & \textbf{ Keeping Ratio @ Pareto } \\
		\hline
		Original GPT-3 & - & 1.30\% \\
		\hline
		\textsc{Our GPT-3} & 3.22\% & 1.41\% \\
		\hline
		\textsc{Chinese} & 1.81\% & -
	\end{tblr}
\end{table}

Furthermore, we assess the filtering effectiveness of these classifiers by comparing their keeping ratios on CommonCrawl. The results are summarized in Table \ref{tab:qc_keep_ratio}, where we employ two data keeping methods used in GPT-3: (1) \textbf{label}: $doc_{score} > 0.5$; and (2) \textbf{Pareto} \cite{brown2020gpt3}: $doc_{score} > 1 - \text{np.random.pareto}(\alpha),\alpha=9$. 
The keeping ratios of our re-implemented GPT-3 quality classifiers are generally in line with the original one, and our Chinese extended version maintains a keeping ratio comparable to that of the English version.

\twosubsection{System Scalability}
\label{exp:scalability-main}
To verify the enhanced scalability of our system (as detailed in Sec. \ref{sec:sys_optimization}), we carry out a series of experiments to measure data processing times across multiple servers. 
Specifically, we adopt the StackExchange and arXiv datasets from RedPajama.
The total size of the StackExchange and arXiv datasets are 65GB and 140GB in jsonl format, respectively. 
We compare the performance of \oursys on Ray, \oursys on Beam (using the Flink backend), and original \oursys in these tests.
More details about the implementation and experimental platforms are in Appendix \ref{appendx:imp-details:scalability}.

\begin{figure}[!ht]
	\centering
	\subfigure{
		\begin{minipage}[]{0.4\textwidth}
			\centering
			\includegraphics[width=\textwidth]{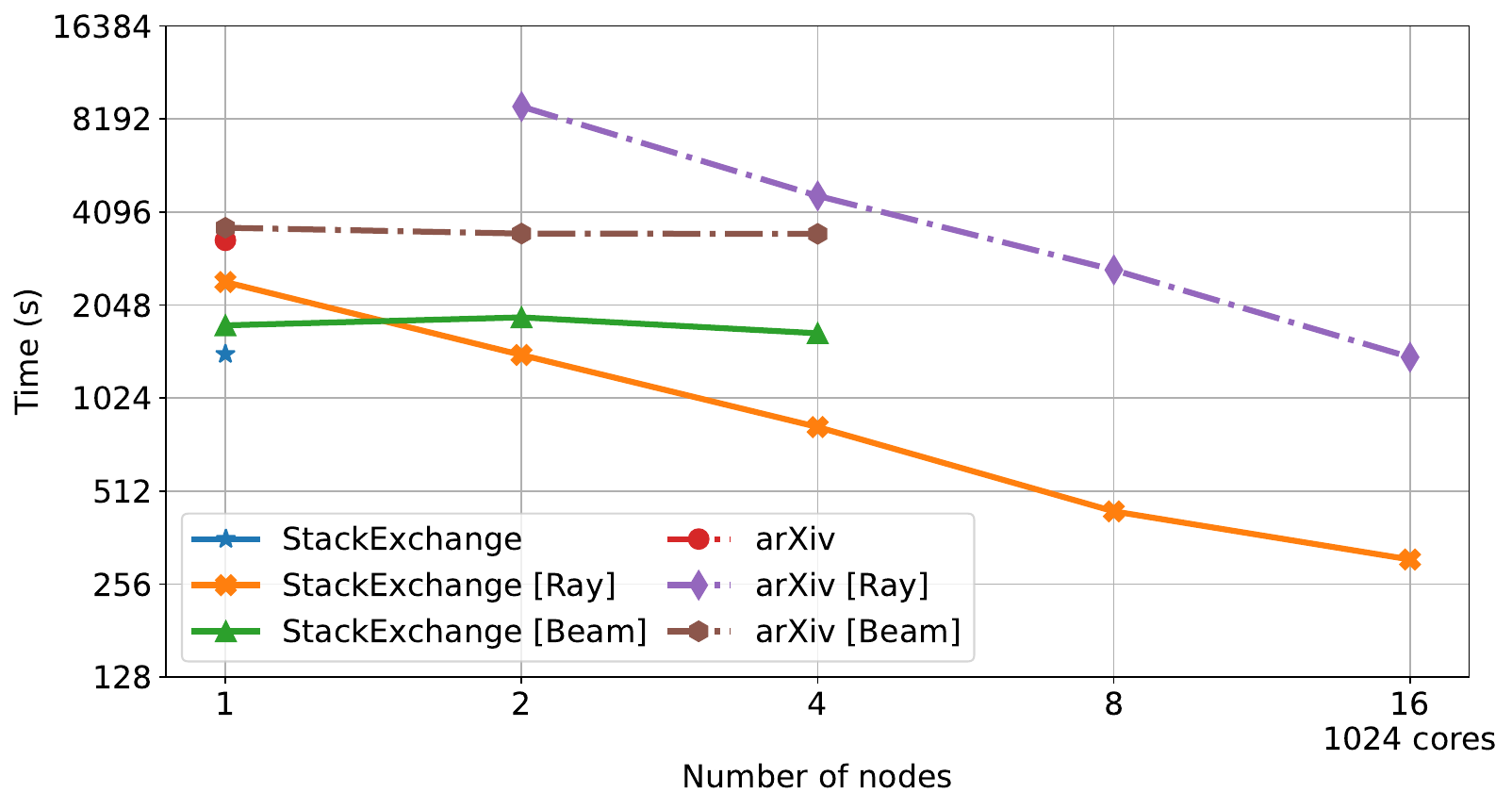}
		\end{minipage}%
	}
	\caption{Processing time with varying number of nodes. \oursys accelerates processing in distributed mode.}
	\label{fig:scalability}
\end{figure}

The experiment results are illustrated in Figure \ref{fig:scalability}. 
Notably, thanks to various optimizations, our original system outperforms both Ray and Beam in the single server scenario. Moreover, as the number of nodes increases, the processing time of our system on Ray decreases proportionally (up to 87.4\% and 84.6\% time reduction on StackExchange and arXiv respectively), demonstrating its effective scalability across multiple servers. 

Nonetheless, the processing time of \oursys on Beam remains almost unchanged as the number of nodes increases. 
Upon further investigation of the processing workflow, we found that the limited scalability of \oursys on Beam is primarily constrained by the data loading component of Beam, which leads to a dominant file loading time ratio and requires substantial development changes for adaptation and further performance optimization.

\subsection{Empowering Real-world Products}
\oursys has been adopted by several real-world LLM-based products, playing a crucial role in data understanding and processing. It evolves continually through the integration of feedback from real-world demands. A notable testament to its utility is its contribution to the development of several industrial LLMs from Alibaba Cloud's Tongyi suite \cite{tongyi}, such as Dianjin, which is used for financial analysis; Zhiwen, a reading assistance tool; and Xingchen, which specializes in AI character customization.
Moreover, the data processing capabilities of \oursys have been incorporated into Alibaba Cloud's Platform for AI (PAI) \cite{pai} to support more real-world applications. 

Our system's fine-grained OP abstraction, coupled with the extensive tools for LLM data-processing, empowers users to easily explore and refine data recipes tailored to the distinct textual attributes of diverse use cases.
For example, within the financial sector, it is crucial to accommodate data that includes numerous digits and standardized terminology. In the realm of reading assistance, the focus shifts to data characterized by extended text lengths and coherent structures. Conversely, character customization demands data rich in dialogue and varied enough to support personalized services. \oursys adeptly meets these varied demands by facilitating the combination of distinct OPs, hyper-parameters, and tools that adapt to the unique need of each real-world application.
 
\section{Conclusions}
\label{sec:conclusion}
To conclude, the introduction of \oursys reflects a new step forward in the field of data-centric LLM development. 
By offering a user-friendly, versatile, and efficient solution, \oursys effectively addresses the existing limitations of open-source tools for LLM data processing, which lean towards data reproducibility at the expense of adaptability and usability. 
The decoupling of traditionally linked components fosters greater abstraction and modularity, and the organic arrangement of over 50 built-in operators, dedicated tools, and abundant data recipes serves diverse needs for LLM pre-training and fine-tuning. 
Beyond supporting auto-evaluation, \oursys is carefully optimized and seamlessly integrated with both ecosystems for LLM  training and evaluation, as well as distributed computing. 
Empirical validation bears witness to substantial improvements in LLMs' performance using \oursys's data recipes, and shows advances in system efficiency and scalability. 
As such, \oursys stands as a compelling addition to the toolkit for LLM data processing, which we hope can shed light on broader research for the field of data-centric LLM development.

\clearpage
\newpage

\bibliographystyle{ACM-Reference-Format}
\bibliography{custom}


\begin{thebibliography}{110}


\ifx \showCODEN    \undefined \def \showCODEN     #1{\unskip}     \fi
\ifx \showDOI      \undefined \def \showDOI       #1{#1}\fi
\ifx \showISBNx    \undefined \def \showISBNx     #1{\unskip}     \fi
\ifx \showISBNxiii \undefined \def \showISBNxiii  #1{\unskip}     \fi
\ifx \showISSN     \undefined \def \showISSN      #1{\unskip}     \fi
\ifx \showLCCN     \undefined \def \showLCCN      #1{\unskip}     \fi
\ifx \shownote     \undefined \def \shownote      #1{#1}          \fi
\ifx \showarticletitle \undefined \def \showarticletitle #1{#1}   \fi
\ifx \showURL      \undefined \def \showURL       {\relax}        \fi
\providecommand\bibfield[2]{#2}
\providecommand\bibinfo[2]{#2}
\providecommand\natexlab[1]{#1}
\providecommand\showeprint[2][]{arXiv:#2}

\bibitem[Almazrouei et~al\mbox{.}(2023)]%
        {Ebtesam-arxiv-2023-Falcon40b}
\bibfield{author}{\bibinfo{person}{Ebtesam Almazrouei}, \bibinfo{person}{Hamza
  Alobeidli}, \bibinfo{person}{Abdulaziz Alshamsi}, \bibinfo{person}{Alessandro
  Cappelli}, \bibinfo{person}{Ruxandra Cojocaru}, \bibinfo{person}{Merouane
  Debbah}, \bibinfo{person}{Etienne Goffinet}, \bibinfo{person}{Daniel Heslow},
  \bibinfo{person}{Julien Launay}, \bibinfo{person}{Quentin Malartic},
  \bibinfo{person}{Badreddine Noune}, \bibinfo{person}{Baptiste Pannier}, {and}
  \bibinfo{person}{Guilherme Penedo}.} \bibinfo{year}{2023}\natexlab{}.
\newblock \showarticletitle{{Falcon-40B}: an open large language model with
  state-of-the-art performance}.
\newblock  (\bibinfo{year}{2023}).
\newblock


\bibitem[Arrow(2023)]%
        {apache-arrow}
\bibfield{author}{\bibinfo{person}{Apache Arrow}.}
  \bibinfo{year}{2023}\natexlab{}.
\newblock
\newblock
\urldef\tempurl%
\url{https://arrow.apache.org/}
\showURL{%
\tempurl}


\bibitem[Askell et~al\mbox{.}(2021)]%
        {Askell-arxiv-2021-Anthropic}
\bibfield{author}{\bibinfo{person}{Amanda Askell}, \bibinfo{person}{Yuntao
  Bai}, \bibinfo{person}{Anna Chen}, \bibinfo{person}{Dawn Drain},
  \bibinfo{person}{Deep Ganguli}, \bibinfo{person}{Tom Henighan},
  \bibinfo{person}{Andy Jones}, \bibinfo{person}{Nicholas Joseph},
  \bibinfo{person}{Benjamin Mann}, \bibinfo{person}{Nova DasSarma},
  \bibinfo{person}{Nelson Elhage}, \bibinfo{person}{Zac Hatfield{-}Dodds},
  \bibinfo{person}{Danny Hernandez}, \bibinfo{person}{Jackson Kernion},
  \bibinfo{person}{Kamal Ndousse}, \bibinfo{person}{Catherine Olsson},
  \bibinfo{person}{Dario Amodei}, \bibinfo{person}{Tom~B. Brown},
  \bibinfo{person}{Jack Clark}, \bibinfo{person}{Sam McCandlish},
  \bibinfo{person}{Chris Olah}, {and} \bibinfo{person}{Jared Kaplan}.}
  \bibinfo{year}{2021}\natexlab{}.
\newblock \showarticletitle{A General Language Assistant as a Laboratory for
  Alignment}.
\newblock \bibinfo{journal}{\emph{CoRR}}  \bibinfo{volume}{abs/2112.00861}
  (\bibinfo{year}{2021}).
\newblock


\bibitem[Bach et~al\mbox{.}(2022)]%
        {bach-etal-2022-promptsource}
\bibfield{author}{\bibinfo{person}{Stephen~H. Bach}, \bibinfo{person}{Victor
  Sanh}, \bibinfo{person}{Zheng~Xin Yong}, \bibinfo{person}{Albert Webson},
  \bibinfo{person}{Colin Raffel}, \bibinfo{person}{Nihal~V. Nayak},
  \bibinfo{person}{Abheesht Sharma}, \bibinfo{person}{Taewoon Kim},
  \bibinfo{person}{M.~Saiful Bari}, \bibinfo{person}{Thibault F{\'{e}}vry},
  \bibinfo{person}{Zaid Alyafeai}, \bibinfo{person}{Manan Dey},
  \bibinfo{person}{Andrea Santilli}, \bibinfo{person}{Zhiqing Sun},
  \bibinfo{person}{Srulik Ben{-}David}, \bibinfo{person}{Canwen Xu},
  \bibinfo{person}{Gunjan Chhablani}, \bibinfo{person}{Han Wang},
  \bibinfo{person}{Jason~Alan Fries}, \bibinfo{person}{Maged~Saeed AlShaibani},
  \bibinfo{person}{Shanya Sharma}, \bibinfo{person}{Urmish Thakker},
  \bibinfo{person}{Khalid Almubarak}, \bibinfo{person}{Xiangru Tang},
  \bibinfo{person}{Dragomir~R. Radev}, \bibinfo{person}{Mike~Tian{-}Jian
  Jiang}, {and} \bibinfo{person}{Alexander~M. Rush}.}
  \bibinfo{year}{2022}\natexlab{}.
\newblock \showarticletitle{PromptSource: An Integrated Development Environment
  and Repository for Natural Language Prompts}. In
  \bibinfo{booktitle}{\emph{{ACL} {(demo)}}}. \bibinfo{pages}{93--104}.
\newblock


\bibitem[Beam(2023)]%
        {apache-beam}
\bibfield{author}{\bibinfo{person}{Apache Beam}.}
  \bibinfo{year}{2023}\natexlab{}.
\newblock
\newblock
\urldef\tempurl%
\url{https://beam.apache.org/}
\showURL{%
\tempurl}


\bibitem[Biderman et~al\mbox{.}(2023)]%
        {Biderman-arxiv-2023-Pythia}
\bibfield{author}{\bibinfo{person}{Stella Biderman}, \bibinfo{person}{Hailey
  Schoelkopf}, \bibinfo{person}{Quentin~Gregory Anthony},
  \bibinfo{person}{Herbie Bradley}, \bibinfo{person}{Kyle O'Brien},
  \bibinfo{person}{Eric Hallahan}, \bibinfo{person}{Mohammad~Aflah Khan},
  \bibinfo{person}{Shivanshu Purohit}, \bibinfo{person}{USVSN~Sai Prashanth},
  \bibinfo{person}{Edward Raff}, \bibinfo{person}{Aviya Skowron},
  \bibinfo{person}{Lintang Sutawika}, {and} \bibinfo{person}{Oskar van~der
  Wal}.} \bibinfo{year}{2023}\natexlab{}.
\newblock \showarticletitle{Pythia: {A} Suite for Analyzing Large Language
  Models Across Training and Scaling}. In \bibinfo{booktitle}{\emph{{ICML}}},
  Vol.~\bibinfo{volume}{202}. \bibinfo{pages}{2397--2430}.
\newblock


\bibitem[Black et~al\mbox{.}(2022)]%
        {Black-CoRR-2022-GPT}
\bibfield{author}{\bibinfo{person}{Sid Black}, \bibinfo{person}{Stella
  Biderman}, \bibinfo{person}{Eric Hallahan}, \bibinfo{person}{Quentin
  Anthony}, \bibinfo{person}{Leo Gao}, \bibinfo{person}{Laurence Golding},
  \bibinfo{person}{Horace He}, \bibinfo{person}{Connor Leahy},
  \bibinfo{person}{Kyle McDonell}, \bibinfo{person}{Jason Phang},
  \bibinfo{person}{Michael Pieler}, \bibinfo{person}{USVSN~Sai Prashanth},
  \bibinfo{person}{Shivanshu Purohit}, \bibinfo{person}{Laria Reynolds},
  \bibinfo{person}{Jonathan Tow}, \bibinfo{person}{Ben Wang}, {and}
  \bibinfo{person}{Samuel Weinbach}.} \bibinfo{year}{2022}\natexlab{}.
\newblock \showarticletitle{GPT-NeoX-20B: An Open-Source Autoregressive
  Language Model}.
\newblock \bibinfo{journal}{\emph{CoRR}}  \bibinfo{volume}{abs/2204.06745}
  (\bibinfo{year}{2022}).
\newblock


\bibitem[Broder et~al\mbox{.}(2000)]%
        {BRODER2000630-minhash}
\bibfield{author}{\bibinfo{person}{Andrei~Z Broder}, \bibinfo{person}{Moses
  Charikar}, \bibinfo{person}{Alan~M Frieze}, {and} \bibinfo{person}{Michael
  Mitzenmacher}.} \bibinfo{year}{2000}\natexlab{}.
\newblock \showarticletitle{Min-Wise Independent Permutations}.
\newblock \bibinfo{journal}{\emph{J. Comput. System Sci.}}
  \bibinfo{volume}{60}, \bibinfo{number}{3} (\bibinfo{year}{2000}),
  \bibinfo{pages}{630--659}.
\newblock


\bibitem[Brown et~al\mbox{.}(2020)]%
        {brown2020gpt3}
\bibfield{author}{\bibinfo{person}{Tom~B. Brown}, \bibinfo{person}{Benjamin
  Mann}, \bibinfo{person}{Nick Ryder}, \bibinfo{person}{Melanie Subbiah},
  \bibinfo{person}{Jared Kaplan}, \bibinfo{person}{Prafulla Dhariwal},
  \bibinfo{person}{Arvind Neelakantan}, \bibinfo{person}{Pranav Shyam},
  \bibinfo{person}{Girish Sastry}, \bibinfo{person}{Amanda Askell},
  \bibinfo{person}{Sandhini Agarwal}, \bibinfo{person}{Ariel Herbert{-}Voss},
  \bibinfo{person}{Gretchen Krueger}, \bibinfo{person}{Tom Henighan},
  \bibinfo{person}{Rewon Child}, \bibinfo{person}{Aditya Ramesh},
  \bibinfo{person}{Daniel~M. Ziegler}, \bibinfo{person}{Jeffrey Wu},
  \bibinfo{person}{Clemens Winter}, \bibinfo{person}{Christopher Hesse},
  \bibinfo{person}{Mark Chen}, \bibinfo{person}{Eric Sigler},
  \bibinfo{person}{Mateusz Litwin}, \bibinfo{person}{Scott Gray},
  \bibinfo{person}{Benjamin Chess}, \bibinfo{person}{Jack Clark},
  \bibinfo{person}{Christopher Berner}, \bibinfo{person}{Sam McCandlish},
  \bibinfo{person}{Alec Radford}, \bibinfo{person}{Ilya Sutskever}, {and}
  \bibinfo{person}{Dario Amodei}.} \bibinfo{year}{2020}\natexlab{}.
\newblock \showarticletitle{Language Models are Few-Shot Learners}. In
  \bibinfo{booktitle}{\emph{{NeurIPS}}}.
\newblock


\bibitem[Bubeck et~al\mbox{.}(2023)]%
        {Bubeck-arxiv-2023-Sparks-AGI}
\bibfield{author}{\bibinfo{person}{S{\'{e}}bastien Bubeck},
  \bibinfo{person}{Varun Chandrasekaran}, \bibinfo{person}{Ronen Eldan},
  \bibinfo{person}{Johannes Gehrke}, \bibinfo{person}{Eric Horvitz},
  \bibinfo{person}{Ece Kamar}, \bibinfo{person}{Peter Lee},
  \bibinfo{person}{Yin~Tat Lee}, \bibinfo{person}{Yuanzhi Li},
  \bibinfo{person}{Scott~M. Lundberg}, \bibinfo{person}{Harsha Nori},
  \bibinfo{person}{Hamid Palangi}, \bibinfo{person}{Marco~T{\'{u}}lio Ribeiro},
  {and} \bibinfo{person}{Yi Zhang}.} \bibinfo{year}{2023}\natexlab{}.
\newblock \showarticletitle{Sparks of Artificial General Intelligence: Early
  experiments with {GPT-4}}.
\newblock \bibinfo{journal}{\emph{CoRR}}  \bibinfo{volume}{abs/2303.12712}
  (\bibinfo{year}{2023}).
\newblock


\bibitem[Cai et~al\mbox{.}(2023)]%
        {Cai-arxiv-2023-Tool}
\bibfield{author}{\bibinfo{person}{Tianle Cai}, \bibinfo{person}{Xuezhi Wang},
  \bibinfo{person}{Tengyu Ma}, \bibinfo{person}{Xinyun Chen}, {and}
  \bibinfo{person}{Denny Zhou}.} \bibinfo{year}{2023}\natexlab{}.
\newblock \showarticletitle{Large Language Models as Tool Makers}.
\newblock \bibinfo{journal}{\emph{CoRR}}  \bibinfo{volume}{abs/2305.17126}
  (\bibinfo{year}{2023}).
\newblock


\bibitem[Carbone et~al\mbox{.}(2015)]%
        {carbone2015apache-flink}
\bibfield{author}{\bibinfo{person}{Paris Carbone}, \bibinfo{person}{Asterios
  Katsifodimos}, \bibinfo{person}{Stephan Ewen}, \bibinfo{person}{Volker
  Markl}, \bibinfo{person}{Seif Haridi}, {and} \bibinfo{person}{Kostas
  Tzoumas}.} \bibinfo{year}{2015}\natexlab{}.
\newblock \showarticletitle{Apache Flink: Stream and batch processing in a
  single engine}.
\newblock \bibinfo{journal}{\emph{{IEEE} Data Eng. Bull.}}
  \bibinfo{volume}{38}, \bibinfo{number}{4} (\bibinfo{year}{2015}).
\newblock


\bibitem[Carlini et~al\mbox{.}(2023)]%
        {Carlini-arxiv-2022-Quantifying}
\bibfield{author}{\bibinfo{person}{Nicholas Carlini}, \bibinfo{person}{Daphne
  Ippolito}, \bibinfo{person}{Matthew Jagielski}, \bibinfo{person}{Katherine
  Lee}, \bibinfo{person}{Florian Tram{\`{e}}r}, {and} \bibinfo{person}{Chiyuan
  Zhang}.} \bibinfo{year}{2023}\natexlab{}.
\newblock \showarticletitle{Quantifying Memorization Across Neural Language
  Models}. In \bibinfo{booktitle}{\emph{{ICLR}}}.
\newblock


\bibitem[Charikar(2002)]%
        {simhash}
\bibfield{author}{\bibinfo{person}{Moses~S. Charikar}.}
  \bibinfo{year}{2002}\natexlab{}.
\newblock \showarticletitle{Similarity Estimation Techniques from Rounding
  Algorithms}. In \bibinfo{booktitle}{\emph{STOC}}. \bibinfo{pages}{380–388}.
\newblock


\bibitem[{ChatGLM2-6B }(2023)]%
        {Zhipu-GitHub-2023-ChatGLM2-6B}
\bibfield{author}{\bibinfo{person}{{ChatGLM2-6B }}.}
  \bibinfo{year}{2023}\natexlab{}.
\newblock
\newblock
\urldef\tempurl%
\url{https://github.com/THUDM/ChatGLM2-6B}
\showURL{%
\tempurl}


\bibitem[ChatLLaMA(2023)]%
        {ChatLLaMA}
\bibfield{author}{\bibinfo{person}{ChatLLaMA}.}
  \bibinfo{year}{2023}\natexlab{}.
\newblock
\newblock
\urldef\tempurl%
\url{https://github.com/nebuly-ai/nebuly/tree/main/optimization/chatllama}
\showURL{%
\tempurl}


\bibitem[Chen et~al\mbox{.}(2023)]%
        {chen2023alpagasus}
\bibfield{author}{\bibinfo{person}{Lichang Chen}, \bibinfo{person}{Shiyang Li},
  \bibinfo{person}{Jun Yan}, \bibinfo{person}{Hai Wang}, \bibinfo{person}{Kalpa
  Gunaratna}, \bibinfo{person}{Vikas Yadav}, \bibinfo{person}{Zheng Tang},
  \bibinfo{person}{Vijay Srinivasan}, \bibinfo{person}{Tianyi Zhou},
  \bibinfo{person}{Heng Huang}, {and} \bibinfo{person}{Hongxia Jin}.}
  \bibinfo{year}{2023}\natexlab{}.
\newblock \showarticletitle{AlpaGasus: Training A Better Alpaca with Fewer
  Data}.
\newblock \bibinfo{journal}{\emph{CoRR}}  \bibinfo{volume}{abs/2307.08701}
  (\bibinfo{year}{2023}).
\newblock


\bibitem[Chen et~al\mbox{.}(2021)]%
        {Chen-arxiv-2021-codex}
\bibfield{author}{\bibinfo{person}{Mark Chen}, \bibinfo{person}{Jerry Tworek},
  \bibinfo{person}{Heewoo Jun}, \bibinfo{person}{Qiming Yuan},
  \bibinfo{person}{Henrique~Pond{\'{e}} de Oliveira~Pinto},
  \bibinfo{person}{Jared Kaplan}, \bibinfo{person}{Harrison Edwards},
  \bibinfo{person}{Yuri Burda}, \bibinfo{person}{Nicholas Joseph},
  \bibinfo{person}{Greg Brockman}, \bibinfo{person}{Alex Ray},
  \bibinfo{person}{Raul Puri}, \bibinfo{person}{Gretchen Krueger},
  \bibinfo{person}{Michael Petrov}, \bibinfo{person}{Heidy Khlaaf},
  \bibinfo{person}{Girish Sastry}, \bibinfo{person}{Pamela Mishkin},
  \bibinfo{person}{Brooke Chan}, \bibinfo{person}{Scott Gray},
  \bibinfo{person}{Nick Ryder}, \bibinfo{person}{Mikhail Pavlov},
  \bibinfo{person}{Alethea Power}, \bibinfo{person}{Lukasz Kaiser},
  \bibinfo{person}{Mohammad Bavarian}, \bibinfo{person}{Clemens Winter},
  \bibinfo{person}{Philippe Tillet}, \bibinfo{person}{Felipe~Petroski Such},
  \bibinfo{person}{Dave Cummings}, \bibinfo{person}{Matthias Plappert},
  \bibinfo{person}{Fotios Chantzis}, \bibinfo{person}{Elizabeth Barnes},
  \bibinfo{person}{Ariel Herbert{-}Voss}, \bibinfo{person}{William~Hebgen
  Guss}, \bibinfo{person}{Alex Nichol}, \bibinfo{person}{Alex Paino},
  \bibinfo{person}{Nikolas Tezak}, \bibinfo{person}{Jie Tang},
  \bibinfo{person}{Igor Babuschkin}, \bibinfo{person}{Suchir Balaji},
  \bibinfo{person}{Shantanu Jain}, \bibinfo{person}{William Saunders},
  \bibinfo{person}{Christopher Hesse}, \bibinfo{person}{Andrew~N. Carr},
  \bibinfo{person}{Jan Leike}, \bibinfo{person}{Joshua Achiam},
  \bibinfo{person}{Vedant Misra}, \bibinfo{person}{Evan Morikawa},
  \bibinfo{person}{Alec Radford}, \bibinfo{person}{Matthew Knight},
  \bibinfo{person}{Miles Brundage}, \bibinfo{person}{Mira Murati},
  \bibinfo{person}{Katie Mayer}, \bibinfo{person}{Peter Welinder},
  \bibinfo{person}{Bob McGrew}, \bibinfo{person}{Dario Amodei},
  \bibinfo{person}{Sam McCandlish}, \bibinfo{person}{Ilya Sutskever}, {and}
  \bibinfo{person}{Wojciech Zaremba}.} \bibinfo{year}{2021}\natexlab{}.
\newblock \showarticletitle{Evaluating Large Language Models Trained on Code}.
\newblock \bibinfo{journal}{\emph{CoRR}}  \bibinfo{volume}{abs/2107.03374}
  (\bibinfo{year}{2021}).
\newblock


\bibitem[Chiang et~al\mbox{.}(2023)]%
        {vicuna2023}
\bibfield{author}{\bibinfo{person}{Wei-Lin Chiang}, \bibinfo{person}{Zhuohan
  Li}, \bibinfo{person}{Zi Lin}, \bibinfo{person}{Ying Sheng},
  \bibinfo{person}{Zhanghao Wu}, \bibinfo{person}{Hao Zhang},
  \bibinfo{person}{Lianmin Zheng}, \bibinfo{person}{Siyuan Zhuang},
  \bibinfo{person}{Yonghao Zhuang}, \bibinfo{person}{Joseph~E. Gonzalez},
  \bibinfo{person}{Ion Stoica}, {and} \bibinfo{person}{Eric~P. Xing}.}
  \bibinfo{year}{2023}\natexlab{}.
\newblock \bibinfo{title}{Vicuna: An Open-Source Chatbot Impressing GPT-4 with
  90\%* ChatGPT Quality}.
\newblock
\newblock
\urldef\tempurl%
\url{https://vicuna.lmsys.org}
\showURL{%
\tempurl}


\bibitem[Chung et~al\mbox{.}(2022)]%
        {Chung-arxiv-2022-Scaling}
\bibfield{author}{\bibinfo{person}{Hyung~Won Chung}, \bibinfo{person}{Le Hou},
  \bibinfo{person}{Shayne Longpre}, \bibinfo{person}{Barret Zoph},
  \bibinfo{person}{Yi Tay}, \bibinfo{person}{William Fedus},
  \bibinfo{person}{Eric Li}, \bibinfo{person}{Xuezhi Wang},
  \bibinfo{person}{Mostafa Dehghani}, \bibinfo{person}{Siddhartha Brahma},
  \bibinfo{person}{Albert Webson}, \bibinfo{person}{Shixiang~Shane Gu},
  \bibinfo{person}{Zhuyun Dai}, \bibinfo{person}{Mirac Suzgun},
  \bibinfo{person}{Xinyun Chen}, \bibinfo{person}{Aakanksha Chowdhery},
  \bibinfo{person}{Sharan Narang}, \bibinfo{person}{Gaurav Mishra},
  \bibinfo{person}{Adams Yu}, \bibinfo{person}{Vincent~Y. Zhao},
  \bibinfo{person}{Yanping Huang}, \bibinfo{person}{Andrew~M. Dai},
  \bibinfo{person}{Hongkun Yu}, \bibinfo{person}{Slav Petrov},
  \bibinfo{person}{Ed~H. Chi}, \bibinfo{person}{Jeff Dean},
  \bibinfo{person}{Jacob Devlin}, \bibinfo{person}{Adam Roberts},
  \bibinfo{person}{Denny Zhou}, \bibinfo{person}{Quoc~V. Le}, {and}
  \bibinfo{person}{Jason Wei}.} \bibinfo{year}{2022}\natexlab{}.
\newblock \showarticletitle{Scaling Instruction-Finetuned Language Models}.
\newblock \bibinfo{journal}{\emph{CoRR}}  \bibinfo{volume}{abs/2210.11416}
  (\bibinfo{year}{2022}).
\newblock


\bibitem[Cloud(2023a)]%
        {tongyi}
\bibfield{author}{\bibinfo{person}{Alibaba Cloud}.}
  \bibinfo{year}{2023}\natexlab{a}.
\newblock
\newblock
\urldef\tempurl%
\url{https://tongyi.aliyun.com}
\showURL{%
\tempurl}


\bibitem[Cloud(2023b)]%
        {pai}
\bibfield{author}{\bibinfo{person}{Alibaba Cloud}.}
  \bibinfo{year}{2023}\natexlab{b}.
\newblock
\newblock
\urldef\tempurl%
\url{https://www.alibabacloud.com/en/product/machine-learning}
\showURL{%
\tempurl}


\bibitem[Collet and Kucherawy(2021)]%
        {zstd-rfc8878}
\bibfield{author}{\bibinfo{person}{Yann Collet} {and} \bibinfo{person}{Murray
  Kucherawy}.} \bibinfo{year}{2021}\natexlab{}.
\newblock \bibinfo{title}{{Zstandard Compression and the 'application/zstd'
  Media Type}}.
\newblock \bibinfo{howpublished}{RFC 8878}.
\newblock


\bibitem[Computer(2023)]%
        {together2023redpajama}
\bibfield{author}{\bibinfo{person}{Together Computer}.}
  \bibinfo{year}{2023}\natexlab{}.
\newblock \bibinfo{booktitle}{\emph{RedPajama: An Open Source Recipe to
  Reproduce LLaMA training dataset}}.
\newblock
\urldef\tempurl%
\url{https://github.com/togethercomputer/RedPajama-Data}
\showURL{%
\tempurl}


\bibitem[Cormier et~al\mbox{.}(2021)]%
        {nature-data-recipe-term}
\bibfield{author}{\bibinfo{person}{Michael~J Cormier},
  \bibinfo{person}{Jonathan~R Belyeu}, \bibinfo{person}{Brent~S Pedersen},
  \bibinfo{person}{Joseph Brown}, \bibinfo{person}{Johannes K{\"o}ster}, {and}
  \bibinfo{person}{Aaron~R Quinlan}.} \bibinfo{year}{2021}\natexlab{}.
\newblock \showarticletitle{Go Get Data (GGD) is a framework that facilitates
  reproducible access to genomic data}.
\newblock \bibinfo{journal}{\emph{Nature Communications}} \bibinfo{volume}{12},
  \bibinfo{number}{1} (\bibinfo{year}{2021}), \bibinfo{pages}{2151}.
\newblock


\bibitem[Crawl(2023)]%
        {commoncrawl}
\bibfield{author}{\bibinfo{person}{Common Crawl}.}
  \bibinfo{year}{2023}\natexlab{}.
\newblock
\newblock
\urldef\tempurl%
\url{https://commoncrawl.org/}
\showURL{%
\tempurl}


\bibitem[Devlin et~al\mbox{.}(2019)]%
        {devlin-etal-2019-bert}
\bibfield{author}{\bibinfo{person}{Jacob Devlin}, \bibinfo{person}{Ming-Wei
  Chang}, \bibinfo{person}{Kenton Lee}, {and} \bibinfo{person}{Kristina
  Toutanova}.} \bibinfo{year}{2019}\natexlab{}.
\newblock \showarticletitle{{BERT}: Pre-training of Deep Bidirectional
  Transformers for Language Understanding}. In
  \bibinfo{booktitle}{\emph{{NAACL-HLT} {(1)}}}. \bibinfo{pages}{4171--4186}.
\newblock


\bibitem[Du et~al\mbox{.}(2022)]%
        {Du-ICML-2022-GLaM}
\bibfield{author}{\bibinfo{person}{Nan Du}, \bibinfo{person}{Yanping Huang},
  \bibinfo{person}{Andrew~M. Dai}, \bibinfo{person}{Simon Tong},
  \bibinfo{person}{Dmitry Lepikhin}, \bibinfo{person}{Yuanzhong Xu},
  \bibinfo{person}{Maxim Krikun}, \bibinfo{person}{Yanqi Zhou},
  \bibinfo{person}{Adams~Wei Yu}, \bibinfo{person}{Orhan Firat},
  \bibinfo{person}{Barret Zoph}, \bibinfo{person}{Liam Fedus},
  \bibinfo{person}{Maarten~P. Bosma}, \bibinfo{person}{Zongwei Zhou},
  \bibinfo{person}{Tao Wang}, \bibinfo{person}{Yu~Emma Wang},
  \bibinfo{person}{Kellie Webster}, \bibinfo{person}{Marie Pellat},
  \bibinfo{person}{Kevin Robinson}, \bibinfo{person}{Kathleen~S.
  Meier{-}Hellstern}, \bibinfo{person}{Toju Duke}, \bibinfo{person}{Lucas
  Dixon}, \bibinfo{person}{Kun Zhang}, \bibinfo{person}{Quoc~V. Le},
  \bibinfo{person}{Yonghui Wu}, \bibinfo{person}{Zhifeng Chen}, {and}
  \bibinfo{person}{Claire Cui}.} \bibinfo{year}{2022}\natexlab{}.
\newblock \showarticletitle{GLaM: Efficient Scaling of Language Models with
  Mixture-of-Experts}. In \bibinfo{booktitle}{\emph{{ICML}}}.
  \bibinfo{pages}{5547--5569}.
\newblock


\bibitem[EleutherAI(2023)]%
        {pythia-hf}
\bibfield{author}{\bibinfo{person}{EleutherAI}.}
  \bibinfo{year}{2023}\natexlab{}.
\newblock \bibinfo{title}{Pythia-1.4B}.
\newblock
\newblock
\urldef\tempurl%
\url{https://huggingface.co/EleutherAI/pythia-1.4b}
\showURL{%
\tempurl}


\bibitem[Feng et~al\mbox{.}(2023)]%
        {Feng-arXiv-2023-Towards}
\bibfield{author}{\bibinfo{person}{Guhao Feng}, \bibinfo{person}{Bohang Zhang},
  \bibinfo{person}{Yuntian Gu}, \bibinfo{person}{Haotian Ye},
  \bibinfo{person}{Di He}, {and} \bibinfo{person}{Liwei Wang}.}
  \bibinfo{year}{2023}\natexlab{}.
\newblock \showarticletitle{Towards Revealing the Mystery behind Chain of
  Thought: a Theoretical Perspective}.
\newblock \bibinfo{journal}{\emph{CoRR}}  \bibinfo{volume}{abs/2305.15408}
  (\bibinfo{year}{2023}).
\newblock


\bibitem[Gao et~al\mbox{.}(2021a)]%
        {Gao-arxiv-2021-Pile}
\bibfield{author}{\bibinfo{person}{Leo Gao}, \bibinfo{person}{Stella Biderman},
  \bibinfo{person}{Sid Black}, \bibinfo{person}{Laurence Golding},
  \bibinfo{person}{Travis Hoppe}, \bibinfo{person}{Charles Foster},
  \bibinfo{person}{Jason Phang}, \bibinfo{person}{Horace He},
  \bibinfo{person}{Anish Thite}, \bibinfo{person}{Noa Nabeshima},
  \bibinfo{person}{Shawn Presser}, {and} \bibinfo{person}{Connor Leahy}.}
  \bibinfo{year}{2021}\natexlab{a}.
\newblock \showarticletitle{The Pile: An 800GB Dataset of Diverse Text for
  Language Modeling}.
\newblock \bibinfo{journal}{\emph{CoRR}}  \bibinfo{volume}{abs/2101.00027}
  (\bibinfo{year}{2021}).
\newblock


\bibitem[Gao et~al\mbox{.}(2021b)]%
        {eval-harness}
\bibfield{author}{\bibinfo{person}{Leo Gao}, \bibinfo{person}{Jonathan Tow},
  \bibinfo{person}{Stella Biderman}, \bibinfo{person}{Sid Black},
  \bibinfo{person}{Anthony DiPofi}, \bibinfo{person}{Charles Foster},
  \bibinfo{person}{Laurence Golding}, \bibinfo{person}{Jeffrey Hsu},
  \bibinfo{person}{Kyle McDonell}, \bibinfo{person}{Niklas Muennighoff},
  \bibinfo{person}{Jason Phang}, \bibinfo{person}{Laria Reynolds},
  \bibinfo{person}{Eric Tang}, \bibinfo{person}{Anish Thite},
  \bibinfo{person}{Ben Wang}, \bibinfo{person}{Kevin Wang}, {and}
  \bibinfo{person}{Andy Zou}.} \bibinfo{year}{2021}\natexlab{b}.
\newblock \bibinfo{booktitle}{\emph{A framework for few-shot language model
  evaluation}}.
\newblock


\bibitem[Gehman et~al\mbox{.}(2020)]%
        {Gehman-2023-arxiv-RealToxicityPrompts}
\bibfield{author}{\bibinfo{person}{Samuel Gehman}, \bibinfo{person}{Suchin
  Gururangan}, \bibinfo{person}{Maarten Sap}, \bibinfo{person}{Yejin Choi},
  {and} \bibinfo{person}{Noah~A. Smith}.} \bibinfo{year}{2020}\natexlab{}.
\newblock \showarticletitle{RealToxicityPrompts: Evaluating Neural Toxic
  Degeneration in Language Models}. In \bibinfo{booktitle}{\emph{{EMNLP}
  (Findings)}}. \bibinfo{pages}{3356--3369}.
\newblock


\bibitem[Geng and Liu(2023)]%
        {openlm2023openllama}
\bibfield{author}{\bibinfo{person}{Xinyang Geng} {and} \bibinfo{person}{Hao
  Liu}.} \bibinfo{year}{2023}\natexlab{}.
\newblock \bibinfo{booktitle}{\emph{OpenLLaMA: An Open Reproduction of LLaMA}}.
\newblock
\urldef\tempurl%
\url{https://github.com/openlm-research/open_llama}
\showURL{%
\tempurl}


\bibitem[Gunasekar et~al\mbox{.}(2023)]%
        {gunasekar2023textbooks}
\bibfield{author}{\bibinfo{person}{Suriya Gunasekar}, \bibinfo{person}{Yi
  Zhang}, \bibinfo{person}{Jyoti Aneja}, \bibinfo{person}{Caio César~Teodoro
  Mendes}, \bibinfo{person}{Allie~Del Giorno}, \bibinfo{person}{Sivakanth
  Gopi}, \bibinfo{person}{Mojan Javaheripi}, \bibinfo{person}{Piero Kauffmann},
  \bibinfo{person}{Gustavo de Rosa}, \bibinfo{person}{Olli Saarikivi},
  \bibinfo{person}{Adil Salim}, \bibinfo{person}{Shital Shah},
  \bibinfo{person}{Harkirat~Singh Behl}, \bibinfo{person}{Xin Wang},
  \bibinfo{person}{Sébastien Bubeck}, \bibinfo{person}{Ronen Eldan},
  \bibinfo{person}{Adam~Tauman Kalai}, \bibinfo{person}{Yin~Tat Lee}, {and}
  \bibinfo{person}{Yuanzhi Li}.} \bibinfo{year}{2023}\natexlab{}.
\newblock \bibinfo{title}{Textbooks Are All You Need}.
\newblock
\newblock
\showeprint[arxiv]{2306.11644}~[cs.CL]


\bibitem[Gutenberg(2023)]%
        {Gutenberg}
\bibfield{author}{\bibinfo{person}{Project Gutenberg}.}
  \bibinfo{year}{2023}\natexlab{}.
\newblock
\newblock
\urldef\tempurl%
\url{https://www.gutenberg.org/}
\showURL{%
\tempurl}


\bibitem[Han et~al\mbox{.}(2021)]%
        {Han-AIopen-2021-PTM}
\bibfield{author}{\bibinfo{person}{Xu Han}, \bibinfo{person}{Zhengyan Zhang},
  \bibinfo{person}{Ning Ding}, \bibinfo{person}{Yuxian Gu},
  \bibinfo{person}{Xiao Liu}, \bibinfo{person}{Yuqi Huo},
  \bibinfo{person}{Jiezhong Qiu}, \bibinfo{person}{Yuan Yao},
  \bibinfo{person}{Ao Zhang}, \bibinfo{person}{Liang Zhang},
  \bibinfo{person}{Wentao Han}, \bibinfo{person}{Minlie Huang},
  \bibinfo{person}{Qin Jin}, \bibinfo{person}{Yanyan Lan},
  \bibinfo{person}{Yang Liu}, \bibinfo{person}{Zhiyuan Liu},
  \bibinfo{person}{Zhiwu Lu}, \bibinfo{person}{Xipeng Qiu},
  \bibinfo{person}{Ruihua Song}, \bibinfo{person}{Jie Tang},
  \bibinfo{person}{Ji{-}Rong Wen}, \bibinfo{person}{Jinhui Yuan},
  \bibinfo{person}{Wayne~Xin Zhao}, {and} \bibinfo{person}{Jun Zhu}.}
  \bibinfo{year}{2021}\natexlab{}.
\newblock \showarticletitle{Pre-trained models: Past, present and future}.
\newblock \bibinfo{journal}{\emph{{AI} Open}}  \bibinfo{volume}{2}
  (\bibinfo{year}{2021}), \bibinfo{pages}{225--250}.
\newblock


\bibitem[Hao et~al\mbox{.}(2023)]%
        {Hao-2023-arxiv-ToolkenGPT}
\bibfield{author}{\bibinfo{person}{Shibo Hao}, \bibinfo{person}{Tianyang Liu},
  \bibinfo{person}{Zhen Wang}, {and} \bibinfo{person}{Zhiting Hu}.}
  \bibinfo{year}{2023}\natexlab{}.
\newblock \showarticletitle{ToolkenGPT: Augmenting Frozen Language Models with
  Massive Tools via Tool Embeddings}.
\newblock \bibinfo{journal}{\emph{CoRR}}  \bibinfo{volume}{abs/2305.11554}
  (\bibinfo{year}{2023}).
\newblock


\bibitem[Hendrycks et~al\mbox{.}(2021)]%
        {Hendrycks-ICLR-2021-Measuring}
\bibfield{author}{\bibinfo{person}{Dan Hendrycks}, \bibinfo{person}{Collin
  Burns}, \bibinfo{person}{Steven Basart}, \bibinfo{person}{Andy Zou},
  \bibinfo{person}{Mantas Mazeika}, \bibinfo{person}{Dawn Song}, {and}
  \bibinfo{person}{Jacob Steinhardt}.} \bibinfo{year}{2021}\natexlab{}.
\newblock \showarticletitle{Measuring Massive Multitask Language
  Understanding}. In \bibinfo{booktitle}{\emph{{ICLR}}}.
\newblock


\bibitem[Honovich et~al\mbox{.}(2022)]%
        {Honovich-arxiv-2022-Unnatural}
\bibfield{author}{\bibinfo{person}{Or Honovich}, \bibinfo{person}{Thomas
  Scialom}, \bibinfo{person}{Omer Levy}, {and} \bibinfo{person}{Timo Schick}.}
  \bibinfo{year}{2022}\natexlab{}.
\newblock \showarticletitle{Unnatural Instructions: Tuning Language Models with
  (Almost) No Human Labor}.
\newblock \bibinfo{journal}{\emph{CoRR}}  \bibinfo{volume}{abs/2212.09689}
  (\bibinfo{year}{2022}).
\newblock


\bibitem[Institute(2023a)]%
        {falcon-hf}
\bibfield{author}{\bibinfo{person}{Technology~Innovation Institute}.}
  \bibinfo{year}{2023}\natexlab{a}.
\newblock \bibinfo{title}{Falcon-RW-1B}.
\newblock
\newblock
\urldef\tempurl%
\url{https://huggingface.co/tiiuae/falcon-rw-1b}
\showURL{%
\tempurl}


\bibitem[Institute(2023b)]%
        {flagalpha-llama}
\bibfield{author}{\bibinfo{person}{Technology~Innovation Institute}.}
  \bibinfo{year}{2023}\natexlab{b}.
\newblock \bibinfo{title}{Falcon-RW-1B}.
\newblock
\newblock
\urldef\tempurl%
\url{https://huggingface.co/FlagAlpha/Atom-7B}
\showURL{%
\tempurl}


\bibitem[Izacard et~al\mbox{.}(2022)]%
        {Izacard-arxiv-2022-Few}
\bibfield{author}{\bibinfo{person}{Gautier Izacard}, \bibinfo{person}{Patrick
  S.~H. Lewis}, \bibinfo{person}{Maria Lomeli}, \bibinfo{person}{Lucas
  Hosseini}, \bibinfo{person}{Fabio Petroni}, \bibinfo{person}{Timo Schick},
  \bibinfo{person}{Jane Dwivedi{-}Yu}, \bibinfo{person}{Armand Joulin},
  \bibinfo{person}{Sebastian Riedel}, {and} \bibinfo{person}{Edouard Grave}.}
  \bibinfo{year}{2022}\natexlab{}.
\newblock \showarticletitle{Few-shot Learning with Retrieval Augmented Language
  Models}.
\newblock \bibinfo{journal}{\emph{CoRR}}  \bibinfo{volume}{abs/2208.03299}
  (\bibinfo{year}{2022}).
\newblock


\bibitem[Jain et~al\mbox{.}(2020)]%
        {jain2020-data-quality-overview}
\bibfield{author}{\bibinfo{person}{Abhinav Jain}, \bibinfo{person}{Hima Patel},
  \bibinfo{person}{Lokesh Nagalapatti}, \bibinfo{person}{Nitin Gupta},
  \bibinfo{person}{Sameep Mehta}, \bibinfo{person}{Shanmukha Guttula},
  \bibinfo{person}{Shashank Mujumdar}, \bibinfo{person}{Shazia Afzal},
  \bibinfo{person}{Ruhi Sharma~Mittal}, {and} \bibinfo{person}{Vitobha
  Munigala}.} \bibinfo{year}{2020}\natexlab{}.
\newblock \showarticletitle{Overview and importance of data quality for machine
  learning tasks}. In \bibinfo{booktitle}{\emph{{KDD}}}.
  \bibinfo{pages}{3561--3562}.
\newblock


\bibitem[Ji et~al\mbox{.}(2023)]%
        {BELLE}
\bibfield{author}{\bibinfo{person}{Yunjie Ji}, \bibinfo{person}{Yong Deng},
  \bibinfo{person}{Yan Gong}, \bibinfo{person}{Yiping Peng},
  \bibinfo{person}{Qiang Niu}, \bibinfo{person}{Baochang Ma}, {and}
  \bibinfo{person}{Xiangang Li}.} \bibinfo{year}{2023}\natexlab{}.
\newblock \bibinfo{title}{BELLE: Be Everyone's Large Language model Engine}.
\newblock \bibinfo{howpublished}{\url{https://github.com/LianjiaTech/BELLE}}.
\newblock


\bibitem[jsonargparse(2023)]%
        {jsonargparse}
\bibfield{author}{\bibinfo{person}{jsonargparse}.}
  \bibinfo{year}{2023}\natexlab{}.
\newblock
\newblock
\urldef\tempurl%
\url{https://github.com/omni-us/jsonargparse}
\showURL{%
\tempurl}


\bibitem[Kandpal et~al\mbox{.}(2022)]%
        {Kandpal-ICML-2022-Deduplicating}
\bibfield{author}{\bibinfo{person}{Nikhil Kandpal}, \bibinfo{person}{Eric
  Wallace}, {and} \bibinfo{person}{Colin Raffel}.}
  \bibinfo{year}{2022}\natexlab{}.
\newblock \showarticletitle{Deduplicating Training Data Mitigates Privacy Risks
  in Language Models}. In \bibinfo{booktitle}{\emph{{ICML}}}.
  \bibinfo{pages}{10697--10707}.
\newblock


\bibitem[K{\"{o}}pf et~al\mbox{.}(2023)]%
        {kopf-arxiv-2023-openassistant}
\bibfield{author}{\bibinfo{person}{Andreas K{\"{o}}pf}, \bibinfo{person}{Yannic
  Kilcher}, \bibinfo{person}{Dimitri von R{\"{u}}tte}, \bibinfo{person}{Sotiris
  Anagnostidis}, \bibinfo{person}{Zhi{-}Rui Tam}, \bibinfo{person}{Keith
  Stevens}, \bibinfo{person}{Abdullah Barhoum}, \bibinfo{person}{Nguyen~Minh
  Duc}, \bibinfo{person}{Oliver Stanley}, \bibinfo{person}{Rich{\'{a}}rd
  Nagyfi}, \bibinfo{person}{Shahul ES}, \bibinfo{person}{Sameer Suri},
  \bibinfo{person}{David Glushkov}, \bibinfo{person}{Arnav Dantuluri},
  \bibinfo{person}{Andrew Maguire}, \bibinfo{person}{Christoph Schuhmann},
  \bibinfo{person}{Huu Nguyen}, {and} \bibinfo{person}{Alexander Mattick}.}
  \bibinfo{year}{2023}\natexlab{}.
\newblock \showarticletitle{OpenAssistant Conversations - Democratizing Large
  Language Model Alignment}.
\newblock \bibinfo{journal}{\emph{CoRR}}  \bibinfo{volume}{abs/2304.07327}
  (\bibinfo{year}{2023}).
\newblock


\bibitem[Kudo and Richardson(2018a)]%
        {Kudo2018SentencePieceAS}
\bibfield{author}{\bibinfo{person}{Taku Kudo} {and} \bibinfo{person}{John
  Richardson}.} \bibinfo{year}{2018}\natexlab{a}.
\newblock \showarticletitle{SentencePiece: A simple and language independent
  subword tokenizer and detokenizer for Neural Text Processing}. In
  \bibinfo{booktitle}{\emph{EMNLP}}.
\newblock


\bibitem[Kudo and Richardson(2018b)]%
        {Kudo-EMNLP-2018-SentencePiece}
\bibfield{author}{\bibinfo{person}{Taku Kudo} {and} \bibinfo{person}{John
  Richardson}.} \bibinfo{year}{2018}\natexlab{b}.
\newblock \showarticletitle{SentencePiece: {A} simple and language independent
  subword tokenizer and detokenizer for Neural Text Processing}. In
  \bibinfo{booktitle}{\emph{{EMNLP} (Demonstration)}}.
\newblock


\bibitem[Lauren{\c{c}}on et~al\mbox{.}(2022)]%
        {laurençon2023bigscience-roots}
\bibfield{author}{\bibinfo{person}{Hugo Lauren{\c{c}}on},
  \bibinfo{person}{Lucile Saulnier}, \bibinfo{person}{Thomas Wang},
  \bibinfo{person}{Christopher Akiki}, \bibinfo{person}{Albert~Villanova del
  Moral}, \bibinfo{person}{Teven~Le Scao}, \bibinfo{person}{Leandro von Werra},
  \bibinfo{person}{Chenghao Mou}, \bibinfo{person}{Eduardo~Gonz{\'{a}}lez
  Ponferrada}, \bibinfo{person}{Huu Nguyen}, \bibinfo{person}{J{\"{o}}rg
  Frohberg}, \bibinfo{person}{Mario Sasko}, \bibinfo{person}{Quentin Lhoest},
  \bibinfo{person}{Angelina McMillan{-}Major}, \bibinfo{person}{G{\'{e}}rard
  Dupont}, \bibinfo{person}{Stella Biderman}, \bibinfo{person}{Anna Rogers},
  \bibinfo{person}{Loubna~Ben Allal}, \bibinfo{person}{Francesco~De Toni},
  \bibinfo{person}{Giada Pistilli}, \bibinfo{person}{Olivier Nguyen},
  \bibinfo{person}{Somaieh Nikpoor}, \bibinfo{person}{Maraim Masoud},
  \bibinfo{person}{Pierre Colombo}, \bibinfo{person}{Javier de~la Rosa},
  \bibinfo{person}{Paulo Villegas}, \bibinfo{person}{Tristan Thrush},
  \bibinfo{person}{Shayne Longpre}, \bibinfo{person}{Sebastian Nagel},
  \bibinfo{person}{Leon Weber}, \bibinfo{person}{Manuel Mu{\~{n}}oz},
  \bibinfo{person}{Jian Zhu}, \bibinfo{person}{Daniel van Strien},
  \bibinfo{person}{Zaid Alyafeai}, \bibinfo{person}{Khalid Almubarak},
  \bibinfo{person}{Minh~Chien Vu}, \bibinfo{person}{Itziar Gonzalez{-}Dios},
  \bibinfo{person}{Aitor Soroa}, \bibinfo{person}{Kyle Lo},
  \bibinfo{person}{Manan Dey}, \bibinfo{person}{Pedro~Ortiz Suarez},
  \bibinfo{person}{Aaron Gokaslan}, \bibinfo{person}{Shamik Bose},
  \bibinfo{person}{David~Ifeoluwa Adelani}, \bibinfo{person}{Long Phan},
  \bibinfo{person}{Hieu Tran}, \bibinfo{person}{Ian Yu}, \bibinfo{person}{Suhas
  Pai}, \bibinfo{person}{Jenny Chim}, \bibinfo{person}{Violette Lepercq},
  \bibinfo{person}{Suzana Ilic}, \bibinfo{person}{Margaret Mitchell},
  \bibinfo{person}{Alexandra~Sasha Luccioni}, {and} \bibinfo{person}{Yacine
  Jernite}.} \bibinfo{year}{2022}\natexlab{}.
\newblock \showarticletitle{The BigScience {ROOTS} Corpus: {A} 1.6TB Composite
  Multilingual Dataset}. In \bibinfo{booktitle}{\emph{{NeurIPS}}}.
\newblock


\bibitem[Lee et~al\mbox{.}(2022)]%
        {Lee-ACL-2022-Deduplicating}
\bibfield{author}{\bibinfo{person}{Katherine Lee}, \bibinfo{person}{Daphne
  Ippolito}, \bibinfo{person}{Andrew Nystrom}, \bibinfo{person}{Chiyuan Zhang},
  \bibinfo{person}{Douglas Eck}, \bibinfo{person}{Chris Callison{-}Burch},
  {and} \bibinfo{person}{Nicholas Carlini}.} \bibinfo{year}{2022}\natexlab{}.
\newblock \showarticletitle{Deduplicating Training Data Makes Language Models
  Better}. In \bibinfo{booktitle}{\emph{{ACL} {(1)}}}.
  \bibinfo{pages}{8424--8445}.
\newblock


\bibitem[Lester et~al\mbox{.}(2021)]%
        {Lester-ACL-2021-The}
\bibfield{author}{\bibinfo{person}{Brian Lester}, \bibinfo{person}{Rami
  Al{-}Rfou}, {and} \bibinfo{person}{Noah Constant}.}
  \bibinfo{year}{2021}\natexlab{}.
\newblock \showarticletitle{The Power of Scale for Parameter-Efficient Prompt
  Tuning}. In \bibinfo{booktitle}{\emph{{EMNLP} {(1)}}}.
  \bibinfo{pages}{3045--3059}.
\newblock


\bibitem[Lewis et~al\mbox{.}(2020)]%
        {Lewis-ACL-2020-BART}
\bibfield{author}{\bibinfo{person}{Mike Lewis}, \bibinfo{person}{Yinhan Liu},
  \bibinfo{person}{Naman Goyal}, \bibinfo{person}{Marjan Ghazvininejad},
  \bibinfo{person}{Abdelrahman Mohamed}, \bibinfo{person}{Omer Levy},
  \bibinfo{person}{Veselin Stoyanov}, {and} \bibinfo{person}{Luke
  Zettlemoyer}.} \bibinfo{year}{2020}\natexlab{}.
\newblock \showarticletitle{{BART:} Denoising Sequence-to-Sequence Pre-training
  for Natural Language Generation, Translation, and Comprehension}. In
  \bibinfo{booktitle}{\emph{{ACL}}}. \bibinfo{pages}{7871--7880}.
\newblock


\bibitem[Lhoest et~al\mbox{.}(2021)]%
        {lhoest-etal-2021-datasets}
\bibfield{author}{\bibinfo{person}{Quentin Lhoest},
  \bibinfo{person}{Albert~Villanova del Moral}, \bibinfo{person}{Yacine
  Jernite}, \bibinfo{person}{Abhishek Thakur}, \bibinfo{person}{Patrick von
  Platen}, \bibinfo{person}{Suraj Patil}, \bibinfo{person}{Julien Chaumond},
  \bibinfo{person}{Mariama Drame}, \bibinfo{person}{Julien Plu},
  \bibinfo{person}{Lewis Tunstall}, \bibinfo{person}{Joe Davison},
  \bibinfo{person}{Mario Sasko}, \bibinfo{person}{Gunjan Chhablani},
  \bibinfo{person}{Bhavitvya Malik}, \bibinfo{person}{Simon Brandeis},
  \bibinfo{person}{Teven~Le Scao}, \bibinfo{person}{Victor Sanh},
  \bibinfo{person}{Canwen Xu}, \bibinfo{person}{Nicolas Patry},
  \bibinfo{person}{Angelina McMillan{-}Major}, \bibinfo{person}{Philipp
  Schmid}, \bibinfo{person}{Sylvain Gugger}, \bibinfo{person}{Cl{\'{e}}ment
  Delangue}, \bibinfo{person}{Th{\'{e}}o Matussi{\`{e}}re},
  \bibinfo{person}{Lysandre Debut}, \bibinfo{person}{Stas Bekman},
  \bibinfo{person}{Pierric Cistac}, \bibinfo{person}{Thibault Goehringer},
  \bibinfo{person}{Victor Mustar}, \bibinfo{person}{Fran{\c{c}}ois Lagunas},
  \bibinfo{person}{Alexander~M. Rush}, {and} \bibinfo{person}{Thomas Wolf}.}
  \bibinfo{year}{2021}\natexlab{}.
\newblock \showarticletitle{Datasets: {A} Community Library for Natural
  Language Processing}. In \bibinfo{booktitle}{\emph{{EMNLP} (Demos)}}.
  \bibinfo{pages}{175--184}.
\newblock


\bibitem[Li et~al\mbox{.}(2017)]%
        {li2017hyperband}
\bibfield{author}{\bibinfo{person}{Lisha Li}, \bibinfo{person}{Kevin Jamieson},
  \bibinfo{person}{Giulia DeSalvo}, \bibinfo{person}{Afshin Rostamizadeh},
  {and} \bibinfo{person}{Ameet Talwalkar}.} \bibinfo{year}{2017}\natexlab{}.
\newblock \showarticletitle{Hyperband: A novel bandit-based approach to
  hyperparameter optimization}.
\newblock \bibinfo{journal}{\emph{J. Mach. Learn. Res.}}  \bibinfo{volume}{18}
  (\bibinfo{year}{2017}), \bibinfo{pages}{185:1--185:52}.
\newblock


\bibitem[Li et~al\mbox{.}(2023a)]%
        {Li-2023-arxiv-Starcoder}
\bibfield{author}{\bibinfo{person}{Raymond Li}, \bibinfo{person}{Loubna~Ben
  Allal}, \bibinfo{person}{Yangtian Zi}, \bibinfo{person}{Niklas Muennighoff},
  \bibinfo{person}{Denis Kocetkov}, \bibinfo{person}{Chenghao Mou},
  \bibinfo{person}{Marc Marone}, \bibinfo{person}{Christopher Akiki},
  \bibinfo{person}{Jia Li}, \bibinfo{person}{Jenny Chim}, \bibinfo{person}{Qian
  Liu}, \bibinfo{person}{Evgenii Zheltonozhskii}, \bibinfo{person}{Terry~Yue
  Zhuo}, \bibinfo{person}{Thomas Wang}, \bibinfo{person}{Olivier Dehaene},
  \bibinfo{person}{Mishig Davaadorj}, \bibinfo{person}{Joel Lamy{-}Poirier},
  \bibinfo{person}{Jo{\~{a}}o Monteiro}, \bibinfo{person}{Oleh Shliazhko},
  \bibinfo{person}{Nicolas Gontier}, \bibinfo{person}{Nicholas Meade},
  \bibinfo{person}{Armel Zebaze}, \bibinfo{person}{Ming{-}Ho Yee},
  \bibinfo{person}{Logesh~Kumar Umapathi}, \bibinfo{person}{Jian Zhu},
  \bibinfo{person}{Benjamin Lipkin}, \bibinfo{person}{Muhtasham Oblokulov},
  \bibinfo{person}{Zhiruo Wang}, \bibinfo{person}{Rudra~Murthy V},
  \bibinfo{person}{Jason Stillerman}, \bibinfo{person}{Siva~Sankalp Patel},
  \bibinfo{person}{Dmitry Abulkhanov}, \bibinfo{person}{Marco Zocca},
  \bibinfo{person}{Manan Dey}, \bibinfo{person}{Zhihan Zhang},
  \bibinfo{person}{Nour Fahmy}, \bibinfo{person}{Urvashi Bhattacharyya},
  \bibinfo{person}{Wenhao Yu}, \bibinfo{person}{Swayam Singh},
  \bibinfo{person}{Sasha Luccioni}, \bibinfo{person}{Paulo Villegas},
  \bibinfo{person}{Maxim Kunakov}, \bibinfo{person}{Fedor Zhdanov},
  \bibinfo{person}{Manuel Romero}, \bibinfo{person}{Tony Lee},
  \bibinfo{person}{Nadav Timor}, \bibinfo{person}{Jennifer Ding},
  \bibinfo{person}{Claire Schlesinger}, \bibinfo{person}{Hailey Schoelkopf},
  \bibinfo{person}{Jan Ebert}, \bibinfo{person}{Tri Dao},
  \bibinfo{person}{Mayank Mishra}, \bibinfo{person}{Alex Gu},
  \bibinfo{person}{Jennifer Robinson}, \bibinfo{person}{Carolyn~Jane Anderson},
  \bibinfo{person}{Brendan Dolan{-}Gavitt}, \bibinfo{person}{Danish
  Contractor}, \bibinfo{person}{Siva Reddy}, \bibinfo{person}{Daniel Fried},
  \bibinfo{person}{Dzmitry Bahdanau}, \bibinfo{person}{Yacine Jernite},
  \bibinfo{person}{Carlos~Mu{\~{n}}oz Ferrandis}, \bibinfo{person}{Sean
  Hughes}, \bibinfo{person}{Thomas Wolf}, \bibinfo{person}{Arjun Guha},
  \bibinfo{person}{Leandro von Werra}, {and} \bibinfo{person}{Harm de Vries}.}
  \bibinfo{year}{2023}\natexlab{a}.
\newblock \showarticletitle{StarCoder: may the source be with you!}
\newblock \bibinfo{journal}{\emph{CoRR}}  \bibinfo{volume}{abs/2305.06161}
  (\bibinfo{year}{2023}).
\newblock


\bibitem[Li et~al\mbox{.}(2023b)]%
        {Li-arxiv-2023-ChatDoctor}
\bibfield{author}{\bibinfo{person}{Yunxiang Li}, \bibinfo{person}{Zihan Li},
  \bibinfo{person}{Kai Zhang}, \bibinfo{person}{Ruilong Dan}, {and}
  \bibinfo{person}{You Zhang}.} \bibinfo{year}{2023}\natexlab{b}.
\newblock \showarticletitle{ChatDoctor: {A} Medical Chat Model Fine-tuned on
  LLaMA Model using Medical Domain Knowledge}.
\newblock \bibinfo{journal}{\emph{CoRR}}  \bibinfo{volume}{abs/2303.14070}
  (\bibinfo{year}{2023}).
\newblock


\bibitem[Liang et~al\mbox{.}(2022)]%
        {liang2022holistic-helm}
\bibfield{author}{\bibinfo{person}{Percy Liang}, \bibinfo{person}{Rishi
  Bommasani}, \bibinfo{person}{Tony Lee}, \bibinfo{person}{Dimitris Tsipras},
  \bibinfo{person}{Dilara Soylu}, \bibinfo{person}{Michihiro Yasunaga},
  \bibinfo{person}{Yian Zhang}, \bibinfo{person}{Deepak Narayanan},
  \bibinfo{person}{Yuhuai Wu}, \bibinfo{person}{Ananya Kumar},
  \bibinfo{person}{Benjamin Newman}, \bibinfo{person}{Binhang Yuan},
  \bibinfo{person}{Bobby Yan}, \bibinfo{person}{Ce Zhang},
  \bibinfo{person}{Christian Cosgrove}, \bibinfo{person}{Christopher~D.
  Manning}, \bibinfo{person}{Christopher Ré}, \bibinfo{person}{Diana
  Acosta-Navas}, \bibinfo{person}{Drew~A. Hudson}, \bibinfo{person}{Eric
  Zelikman}, \bibinfo{person}{Esin Durmus}, \bibinfo{person}{Faisal Ladhak},
  \bibinfo{person}{Frieda Rong}, \bibinfo{person}{Hongyu Ren},
  \bibinfo{person}{Huaxiu Yao}, \bibinfo{person}{Jue Wang},
  \bibinfo{person}{Keshav Santhanam}, \bibinfo{person}{Laurel Orr},
  \bibinfo{person}{Lucia Zheng}, \bibinfo{person}{Mert Yuksekgonul},
  \bibinfo{person}{Mirac Suzgun}, \bibinfo{person}{Nathan Kim},
  \bibinfo{person}{Neel Guha}, \bibinfo{person}{Niladri Chatterji},
  \bibinfo{person}{Omar Khattab}, \bibinfo{person}{Peter Henderson},
  \bibinfo{person}{Qian Huang}, \bibinfo{person}{Ryan Chi},
  \bibinfo{person}{Sang~Michael Xie}, \bibinfo{person}{Shibani Santurkar},
  \bibinfo{person}{Surya Ganguli}, \bibinfo{person}{Tatsunori Hashimoto},
  \bibinfo{person}{Thomas Icard}, \bibinfo{person}{Tianyi Zhang},
  \bibinfo{person}{Vishrav Chaudhary}, \bibinfo{person}{William Wang},
  \bibinfo{person}{Xuechen Li}, \bibinfo{person}{Yifan Mai},
  \bibinfo{person}{Yuhui Zhang}, {and} \bibinfo{person}{Yuta Koreeda}.}
  \bibinfo{year}{2022}\natexlab{}.
\newblock \showarticletitle{Holistic Evaluation of Language Models}.
\newblock \bibinfo{journal}{\emph{CoRR}}  \bibinfo{volume}{abs/2211.09110}
  (\bibinfo{year}{2022}).
\newblock


\bibitem[Liu et~al\mbox{.}(2023)]%
        {Liu-2023-arxiv-G-Eval}
\bibfield{author}{\bibinfo{person}{Yang Liu}, \bibinfo{person}{Dan Iter},
  \bibinfo{person}{Yichong Xu}, \bibinfo{person}{Shuohang Wang},
  \bibinfo{person}{Ruochen Xu}, {and} \bibinfo{person}{Chenguang Zhu}.}
  \bibinfo{year}{2023}\natexlab{}.
\newblock \showarticletitle{G-Eval: {NLG} Evaluation using {GPT-4} with Better
  Human Alignment}.
\newblock \bibinfo{journal}{\emph{CoRR}}  \bibinfo{volume}{abs/2303.16634}
  (\bibinfo{year}{2023}).
\newblock


\bibitem[Longpre et~al\mbox{.}(2023a)]%
        {Longpre-arxiv-2023-The}
\bibfield{author}{\bibinfo{person}{Shayne Longpre}, \bibinfo{person}{Le Hou},
  \bibinfo{person}{Tu Vu}, \bibinfo{person}{Albert Webson},
  \bibinfo{person}{Hyung~Won Chung}, \bibinfo{person}{Yi Tay},
  \bibinfo{person}{Denny Zhou}, \bibinfo{person}{Quoc~V. Le},
  \bibinfo{person}{Barret Zoph}, \bibinfo{person}{Jason Wei}, {and}
  \bibinfo{person}{Adam Roberts}.} \bibinfo{year}{2023}\natexlab{a}.
\newblock \showarticletitle{The Flan Collection: Designing Data and Methods for
  Effective Instruction Tuning}.
\newblock \bibinfo{journal}{\emph{CoRR}}  \bibinfo{volume}{abs/2301.13688}
  (\bibinfo{year}{2023}).
\newblock


\bibitem[Longpre et~al\mbox{.}(2023b)]%
        {Longpre-arxiv-2023-pretrainer}
\bibfield{author}{\bibinfo{person}{Shayne Longpre}, \bibinfo{person}{Gregory
  Yauney}, \bibinfo{person}{Emily Reif}, \bibinfo{person}{Katherine Lee},
  \bibinfo{person}{Adam Roberts}, \bibinfo{person}{Barret Zoph},
  \bibinfo{person}{Denny Zhou}, \bibinfo{person}{Jason Wei},
  \bibinfo{person}{Kevin Robinson}, \bibinfo{person}{David Mimno}, {and}
  \bibinfo{person}{Daphne Ippolito}.} \bibinfo{year}{2023}\natexlab{b}.
\newblock \showarticletitle{A Pretrainer's Guide to Training Data: Measuring
  the Effects of Data Age, Domain Coverage, Quality, {\&} Toxicity}.
\newblock \bibinfo{journal}{\emph{CoRR}}  \bibinfo{volume}{abs/2305.13169}
  (\bibinfo{year}{2023}).
\newblock


\bibitem[Loshchilov and Hutter(2017)]%
        {Loshchilov-arxiv-2017-Fixing}
\bibfield{author}{\bibinfo{person}{Ilya Loshchilov} {and}
  \bibinfo{person}{Frank Hutter}.} \bibinfo{year}{2017}\natexlab{}.
\newblock \showarticletitle{Fixing Weight Decay Regularization in Adam}.
\newblock \bibinfo{journal}{\emph{CoRR}}  \bibinfo{volume}{abs/1711.05101}
  (\bibinfo{year}{2017}).
\newblock


\bibitem[LZ4(2023)]%
        {lz4}
\bibfield{author}{\bibinfo{person}{LZ4}.} \bibinfo{year}{2023}\natexlab{}.
\newblock
\newblock
\urldef\tempurl%
\url{https://www.lz4.org/}
\showURL{%
\tempurl}


\bibitem[Malinka et~al\mbox{.}(2023)]%
        {Malinka-arxiv-2023-Education}
\bibfield{author}{\bibinfo{person}{Kamil Malinka}, \bibinfo{person}{Martin
  Peres{\'{\i}}ni}, \bibinfo{person}{Anton Firc}, \bibinfo{person}{Ondrej
  Hujnak}, {and} \bibinfo{person}{Filip Janus}.}
  \bibinfo{year}{2023}\natexlab{}.
\newblock \showarticletitle{On the Educational Impact of ChatGPT: Is Artificial
  Intelligence Ready to Obtain a University Degree?}
\newblock \bibinfo{journal}{\emph{CoRR}}  \bibinfo{volume}{abs/2303.11146}
  (\bibinfo{year}{2023}).
\newblock


\bibitem[Moritz et~al\mbox{.}(2018)]%
        {ray}
\bibfield{author}{\bibinfo{person}{Philipp Moritz}, \bibinfo{person}{Robert
  Nishihara}, \bibinfo{person}{Stephanie Wang}, \bibinfo{person}{Alexey
  Tumanov}, \bibinfo{person}{Richard Liaw}, \bibinfo{person}{Eric Liang},
  \bibinfo{person}{Melih Elibol}, \bibinfo{person}{Zongheng Yang},
  \bibinfo{person}{William Paul}, \bibinfo{person}{Michael~I. Jordan}, {and}
  \bibinfo{person}{Ion Stoica}.} \bibinfo{year}{2018}\natexlab{}.
\newblock \showarticletitle{Ray: {A} Distributed Framework for Emerging {AI}
  Applications}. In \bibinfo{booktitle}{\emph{{OSDI}}}.
  \bibinfo{pages}{561--577}.
\newblock


\bibitem[Nijkamp et~al\mbox{.}(2023)]%
        {nijkamp-arxiv-2022-Codegen}
\bibfield{author}{\bibinfo{person}{Erik Nijkamp}, \bibinfo{person}{Bo Pang},
  \bibinfo{person}{Hiroaki Hayashi}, \bibinfo{person}{Lifu Tu},
  \bibinfo{person}{Huan Wang}, \bibinfo{person}{Yingbo Zhou},
  \bibinfo{person}{Silvio Savarese}, {and} \bibinfo{person}{Caiming Xiong}.}
  \bibinfo{year}{2023}\natexlab{}.
\newblock \showarticletitle{CodeGen: An Open Large Language Model for Code with
  Multi-Turn Program Synthesis}. In \bibinfo{booktitle}{\emph{{ICLR}}}.
\newblock


\bibitem[OpenAI(2022)]%
        {OpenAI-blog-2022-alignment}
\bibfield{author}{\bibinfo{person}{OpenAI}.} \bibinfo{year}{2022}\natexlab{}.
\newblock \showarticletitle{Our approach to alignment research}.
\newblock \bibinfo{journal}{\emph{OpenAI Blog}} (\bibinfo{date}{August}
  \bibinfo{year}{2022}).
\newblock


\bibitem[OpenAI(2023)]%
        {openai2023gpt4}
\bibfield{author}{\bibinfo{person}{OpenAI}.} \bibinfo{year}{2023}\natexlab{}.
\newblock \showarticletitle{GPT-4 Technical Report}.
\newblock \bibinfo{journal}{\emph{CoRR}}  \bibinfo{volume}{abs/2303.08774}
  (\bibinfo{year}{2023}).
\newblock


\bibitem[Ouyang et~al\mbox{.}(2022)]%
        {ouyang2022training-instructgpt}
\bibfield{author}{\bibinfo{person}{Long Ouyang}, \bibinfo{person}{Jeffrey Wu},
  \bibinfo{person}{Xu Jiang}, \bibinfo{person}{Diogo Almeida},
  \bibinfo{person}{Carroll~L. Wainwright}, \bibinfo{person}{Pamela Mishkin},
  \bibinfo{person}{Chong Zhang}, \bibinfo{person}{Sandhini Agarwal},
  \bibinfo{person}{Katarina Slama}, \bibinfo{person}{Alex Ray},
  \bibinfo{person}{John Schulman}, \bibinfo{person}{Jacob Hilton},
  \bibinfo{person}{Fraser Kelton}, \bibinfo{person}{Luke Miller},
  \bibinfo{person}{Maddie Simens}, \bibinfo{person}{Amanda Askell},
  \bibinfo{person}{Peter Welinder}, \bibinfo{person}{Paul~F. Christiano},
  \bibinfo{person}{Jan Leike}, {and} \bibinfo{person}{Ryan Lowe}.}
  \bibinfo{year}{2022}\natexlab{}.
\newblock \showarticletitle{Training language models to follow instructions
  with human feedback}. In \bibinfo{booktitle}{\emph{NeurIPS}}.
\newblock


\bibitem[Penedo et~al\mbox{.}(2023)]%
        {Penedo-2023-arxiv-Refinedweb}
\bibfield{author}{\bibinfo{person}{Guilherme Penedo}, \bibinfo{person}{Quentin
  Malartic}, \bibinfo{person}{Daniel Hesslow}, \bibinfo{person}{Ruxandra
  Cojocaru}, \bibinfo{person}{Alessandro Cappelli}, \bibinfo{person}{Hamza
  Alobeidli}, \bibinfo{person}{Baptiste Pannier}, \bibinfo{person}{Ebtesam
  Almazrouei}, {and} \bibinfo{person}{Julien Launay}.}
  \bibinfo{year}{2023}\natexlab{}.
\newblock \showarticletitle{The RefinedWeb Dataset for Falcon {LLM:}
  Outperforming Curated Corpora with Web Data, and Web Data Only}.
\newblock \bibinfo{journal}{\emph{CoRR}}  \bibinfo{volume}{abs/2306.01116}
  (\bibinfo{year}{2023}).
\newblock


\bibitem[Peters et~al\mbox{.}(2018)]%
        {Peters-NAACL-2018}
\bibfield{author}{\bibinfo{person}{Matthew~E. Peters}, \bibinfo{person}{Mark
  Neumann}, \bibinfo{person}{Mohit Iyyer}, \bibinfo{person}{Matt Gardner},
  \bibinfo{person}{Christopher Clark}, \bibinfo{person}{Kenton Lee}, {and}
  \bibinfo{person}{Luke Zettlemoyer}.} \bibinfo{year}{2018}\natexlab{}.
\newblock \showarticletitle{Deep Contextualized Word Representations}. In
  \bibinfo{booktitle}{\emph{{NAACL-HLT}}}. \bibinfo{pages}{2227--2237}.
\newblock


\bibitem[Qiao et~al\mbox{.}(2023)]%
        {qiao2023reasoning}
\bibfield{author}{\bibinfo{person}{Shuofei Qiao}, \bibinfo{person}{Yixin Ou},
  \bibinfo{person}{Ningyu Zhang}, \bibinfo{person}{Xiang Chen},
  \bibinfo{person}{Yunzhi Yao}, \bibinfo{person}{Shumin Deng},
  \bibinfo{person}{Chuanqi Tan}, \bibinfo{person}{Fei Huang}, {and}
  \bibinfo{person}{Huajun Chen}.} \bibinfo{year}{2023}\natexlab{}.
\newblock \bibinfo{title}{Reasoning with Language Model Prompting: A Survey}.
\newblock
\newblock
\showeprint[arxiv]{2212.09597}~[cs.CL]


\bibitem[Qingyi~Si(2023)]%
        {alpaca-cot}
\bibfield{author}{\bibinfo{person}{Zheng~Lin Qingyi~Si}.}
  \bibinfo{year}{2023}\natexlab{}.
\newblock \bibinfo{title}{Alpaca-CoT: An Instruction Fine-Tuning Platform with
  Instruction Data Collection and Unified Large Language Models Interface}.
\newblock
\newblock
\urldef\tempurl%
\url{https://github.com/PhoebusSi/alpaca-CoT}
\showURL{%
\tempurl}


\bibitem[Radford et~al\mbox{.}(2018)]%
        {radford2018gpt1}
\bibfield{author}{\bibinfo{person}{Alec Radford}, \bibinfo{person}{Karthik
  Narasimhan}, \bibinfo{person}{Tim Salimans}, \bibinfo{person}{Ilya
  Sutskever}, {et~al\mbox{.}}} \bibinfo{year}{2018}\natexlab{}.
\newblock \showarticletitle{Improving language understanding by generative
  pre-training}.
\newblock  (\bibinfo{year}{2018}).
\newblock


\bibitem[Radford et~al\mbox{.}(2019)]%
        {radford2019gpt2}
\bibfield{author}{\bibinfo{person}{Alec Radford}, \bibinfo{person}{Jeffrey Wu},
  \bibinfo{person}{Rewon Child}, \bibinfo{person}{David Luan},
  \bibinfo{person}{Dario Amodei}, \bibinfo{person}{Ilya Sutskever},
  {et~al\mbox{.}}} \bibinfo{year}{2019}\natexlab{}.
\newblock \showarticletitle{Language models are unsupervised multitask
  learners}.
\newblock \bibinfo{journal}{\emph{OpenAI blog}} \bibinfo{volume}{1},
  \bibinfo{number}{8} (\bibinfo{year}{2019}), \bibinfo{pages}{9}.
\newblock


\bibitem[Raffel et~al\mbox{.}(2020)]%
        {Raffel-JMLR-2020-Exploring}
\bibfield{author}{\bibinfo{person}{Colin Raffel}, \bibinfo{person}{Noam
  Shazeer}, \bibinfo{person}{Adam Roberts}, \bibinfo{person}{Katherine Lee},
  \bibinfo{person}{Sharan Narang}, \bibinfo{person}{Michael Matena},
  \bibinfo{person}{Yanqi Zhou}, \bibinfo{person}{Wei Li}, {and}
  \bibinfo{person}{Peter~J. Liu}.} \bibinfo{year}{2020}\natexlab{}.
\newblock \showarticletitle{Exploring the Limits of Transfer Learning with a
  Unified Text-to-Text Transformer}.
\newblock \bibinfo{journal}{\emph{J. Mach. Learn. Res.}}
  (\bibinfo{year}{2020}), \bibinfo{pages}{140:1--140:67}.
\newblock


\bibitem[Rasley et~al\mbox{.}(2020)]%
        {Rasley-KDD-2020-DeepSpeed}
\bibfield{author}{\bibinfo{person}{Jeff Rasley}, \bibinfo{person}{Samyam
  Rajbhandari}, \bibinfo{person}{Olatunji Ruwase}, {and}
  \bibinfo{person}{Yuxiong He}.} \bibinfo{year}{2020}\natexlab{}.
\newblock \showarticletitle{Deepspeed: System optimizations enable training
  deep learning models with over 100 billion parameters}. In
  \bibinfo{booktitle}{\emph{{KDD}}}. \bibinfo{pages}{3505--3506}.
\newblock


\bibitem[Ren et~al\mbox{.}(2023)]%
        {Ren-arXiv-2023-PanGusigma}
\bibfield{author}{\bibinfo{person}{Xiaozhe Ren}, \bibinfo{person}{Pingyi Zhou},
  \bibinfo{person}{Xinfan Meng}, \bibinfo{person}{Xinjing Huang},
  \bibinfo{person}{Yadao Wang}, \bibinfo{person}{Weichao Wang},
  \bibinfo{person}{Pengfei Li}, \bibinfo{person}{Xiaoda Zhang},
  \bibinfo{person}{Alexander Podolskiy}, \bibinfo{person}{Grigory Arshinov},
  \bibinfo{person}{Andrey Bout}, \bibinfo{person}{Irina Piontkovskaya},
  \bibinfo{person}{Jiansheng Wei}, \bibinfo{person}{Xin Jiang},
  \bibinfo{person}{Teng Su}, \bibinfo{person}{Qun Liu}, {and}
  \bibinfo{person}{Jun Yao}.} \bibinfo{year}{2023}\natexlab{}.
\newblock \showarticletitle{PanGu-{\(\Sigma\)}: Towards Trillion Parameter
  Language Model with Sparse Heterogeneous Computing}.
\newblock \bibinfo{journal}{\emph{CoRR}}  \bibinfo{volume}{abs/2303.10845}
  (\bibinfo{year}{2023}).
\newblock


\bibitem[Scao et~al\mbox{.}(2022)]%
        {Scao-arxiv-2022-BLOOM}
\bibfield{author}{\bibinfo{person}{Teven~Le Scao}, \bibinfo{person}{Angela
  Fan}, \bibinfo{person}{Christopher Akiki}, \bibinfo{person}{Ellie Pavlick},
  \bibinfo{person}{Suzana Ilic}, \bibinfo{person}{Daniel Hesslow},
  \bibinfo{person}{Roman Castagn{\'{e}}}, \bibinfo{person}{Alexandra~Sasha
  Luccioni}, \bibinfo{person}{Fran{\c{c}}ois Yvon}, \bibinfo{person}{Matthias
  Gall{\'{e}}}, \bibinfo{person}{Jonathan Tow}, \bibinfo{person}{Alexander~M.
  Rush}, \bibinfo{person}{Stella Biderman}, \bibinfo{person}{Albert Webson},
  \bibinfo{person}{Pawan~Sasanka Ammanamanchi}, \bibinfo{person}{Thomas Wang},
  \bibinfo{person}{Beno{\^{\i}}t Sagot}, \bibinfo{person}{Niklas Muennighoff},
  \bibinfo{person}{Albert~Villanova del Moral}, \bibinfo{person}{Olatunji
  Ruwase}, \bibinfo{person}{Rachel Bawden}, \bibinfo{person}{Stas Bekman},
  \bibinfo{person}{Angelina McMillan{-}Major}, \bibinfo{person}{Iz Beltagy},
  \bibinfo{person}{Huu Nguyen}, \bibinfo{person}{Lucile Saulnier},
  \bibinfo{person}{Samson Tan}, \bibinfo{person}{Pedro~Ortiz Suarez},
  \bibinfo{person}{Victor Sanh}, \bibinfo{person}{Hugo Lauren{\c{c}}on},
  \bibinfo{person}{Yacine Jernite}, \bibinfo{person}{Julien Launay},
  \bibinfo{person}{Margaret Mitchell}, \bibinfo{person}{Colin Raffel},
  \bibinfo{person}{Aaron Gokaslan}, \bibinfo{person}{Adi Simhi},
  \bibinfo{person}{Aitor Soroa}, \bibinfo{person}{Alham~Fikri Aji},
  \bibinfo{person}{Amit Alfassy}, \bibinfo{person}{Anna Rogers},
  \bibinfo{person}{Ariel~Kreisberg Nitzav}, \bibinfo{person}{Canwen Xu},
  \bibinfo{person}{Chenghao Mou}, \bibinfo{person}{Chris Emezue},
  \bibinfo{person}{Christopher Klamm}, \bibinfo{person}{Colin Leong},
  \bibinfo{person}{Daniel van Strien}, \bibinfo{person}{David~Ifeoluwa
  Adelani}, {and} \bibinfo{person}{et al.}} \bibinfo{year}{2022}\natexlab{}.
\newblock \showarticletitle{{BLOOM:} {A} 176B-Parameter Open-Access
  Multilingual Language Model}.
\newblock \bibinfo{journal}{\emph{CoRR}}  \bibinfo{volume}{abs/2211.05100}
  (\bibinfo{year}{2022}).
\newblock


\bibitem[Shahmirzadi et~al\mbox{.}(2019)]%
        {shahmirzadi2019text-simtext}
\bibfield{author}{\bibinfo{person}{Omid Shahmirzadi}, \bibinfo{person}{Adam
  Lugowski}, {and} \bibinfo{person}{Kenneth Younge}.}
  \bibinfo{year}{2019}\natexlab{}.
\newblock \showarticletitle{Text similarity in vector space models: a
  comparative study}. In \bibinfo{booktitle}{\emph{{ICMLA}}}.
  \bibinfo{pages}{659--666}.
\newblock


\bibitem[Shahriari et~al\mbox{.}(2015)]%
        {shahriari2015taking-bo}
\bibfield{author}{\bibinfo{person}{Bobak Shahriari}, \bibinfo{person}{Kevin
  Swersky}, \bibinfo{person}{Ziyu Wang}, \bibinfo{person}{Ryan~P Adams}, {and}
  \bibinfo{person}{Nando De~Freitas}.} \bibinfo{year}{2015}\natexlab{}.
\newblock \showarticletitle{Taking the human out of the loop: A review of
  Bayesian optimization}.
\newblock \bibinfo{journal}{\emph{Proc. {IEEE}}} \bibinfo{volume}{104},
  \bibinfo{number}{1} (\bibinfo{year}{2015}), \bibinfo{pages}{148--175}.
\newblock


\bibitem[Shazeer(2020)]%
        {Shazeer-arxiv-2020-GLU}
\bibfield{author}{\bibinfo{person}{Noam Shazeer}.}
  \bibinfo{year}{2020}\natexlab{}.
\newblock \showarticletitle{{GLU} Variants Improve Transformer}.
\newblock   \bibinfo{volume}{abs/2002.05202} (\bibinfo{year}{2020}).
\newblock


\bibitem[Shen et~al\mbox{.}(2023)]%
        {Shen-2023-arXiv-Hugginggpt}
\bibfield{author}{\bibinfo{person}{Yongliang Shen}, \bibinfo{person}{Kaitao
  Song}, \bibinfo{person}{Xu Tan}, \bibinfo{person}{Dongsheng Li},
  \bibinfo{person}{Weiming Lu}, {and} \bibinfo{person}{Yueting Zhuang}.}
  \bibinfo{year}{2023}\natexlab{}.
\newblock \showarticletitle{Hugginggpt: Solving ai tasks with chatgpt and its
  friends in huggingface}.
\newblock \bibinfo{journal}{\emph{arXiv preprint arXiv:2303.17580}}
  (\bibinfo{year}{2023}).
\newblock


\bibitem[Shoeybi et~al\mbox{.}(2019)]%
        {shoeybi2020megatronlm}
\bibfield{author}{\bibinfo{person}{Mohammad Shoeybi}, \bibinfo{person}{Mostofa
  Patwary}, \bibinfo{person}{Raul Puri}, \bibinfo{person}{Patrick LeGresley},
  \bibinfo{person}{Jared Casper}, {and} \bibinfo{person}{Bryan Catanzaro}.}
  \bibinfo{year}{2019}\natexlab{}.
\newblock \showarticletitle{Megatron-LM: Training Multi-Billion Parameter
  Language Models Using Model Parallelism}.
\newblock \bibinfo{journal}{\emph{CoRR}}  \bibinfo{volume}{abs/1909.08053}
  (\bibinfo{year}{2019}).
\newblock


\bibitem[{Soldaini, Luca and Lo, Kyle and Kinney, Rodney and Naik, Aakanksha
  and Ravichander, Abhilasha and Bhagia, Akshita and Groeneveld, Dirk and
  Schwenk, Dustin and Magnusson, Ian and Chandu, Khyathi}(2023)]%
        {DolmaToolkit}
\bibfield{author}{\bibinfo{person}{{Soldaini, Luca and Lo, Kyle and Kinney,
  Rodney and Naik, Aakanksha and Ravichander, Abhilasha and Bhagia, Akshita and
  Groeneveld, Dirk and Schwenk, Dustin and Magnusson, Ian and Chandu,
  Khyathi}}.} \bibinfo{year}{2023}\natexlab{}.
\newblock \bibinfo{booktitle}{\emph{{The Dolma Toolkit}}}.
\newblock
\newblock
\shownote{{Apache 2.0 License, Version \texttt{0.9.0},
  \url{https://github.com/allenai/dolma}}}.


\bibitem[Streamlit(2023)]%
        {streamlit}
\bibfield{author}{\bibinfo{person}{Streamlit}.}
  \bibinfo{year}{2023}\natexlab{}.
\newblock
\newblock
\urldef\tempurl%
\url{https://streamlit.io/}
\showURL{%
\tempurl}


\bibitem[Su et~al\mbox{.}(2021)]%
        {su-2021-arxiv-reformer}
\bibfield{author}{\bibinfo{person}{Jianlin Su}, \bibinfo{person}{Yu Lu},
  \bibinfo{person}{Shengfeng Pan}, \bibinfo{person}{Bo Wen}, {and}
  \bibinfo{person}{Yunfeng Liu}.} \bibinfo{year}{2021}\natexlab{}.
\newblock \showarticletitle{RoFormer: Enhanced Transformer with Rotary Position
  Embedding}.
\newblock \bibinfo{journal}{\emph{CoRR}}  \bibinfo{volume}{abs/2104.09864}
  (\bibinfo{year}{2021}).
\newblock


\bibitem[Su et~al\mbox{.}(2023)]%
        {su-2023-arxiv-pandagpt}
\bibfield{author}{\bibinfo{person}{Yixuan Su}, \bibinfo{person}{Tian Lan},
  \bibinfo{person}{Huayang Li}, \bibinfo{person}{Jialu Xu},
  \bibinfo{person}{Yan Wang}, {and} \bibinfo{person}{Deng Cai}.}
  \bibinfo{year}{2023}\natexlab{}.
\newblock \showarticletitle{PandaGPT: One Model To Instruction-Follow Them
  All}.
\newblock \bibinfo{journal}{\emph{CoRR}}  \bibinfo{volume}{abs/2305.16355}
  (\bibinfo{year}{2023}).
\newblock


\bibitem[Sun et~al\mbox{.}(2021)]%
        {Sun-arXiv-2021-ERNIE3.0}
\bibfield{author}{\bibinfo{person}{Yu Sun}, \bibinfo{person}{Shuohuan Wang},
  \bibinfo{person}{Shikun Feng}, \bibinfo{person}{Siyu Ding},
  \bibinfo{person}{Chao Pang}, \bibinfo{person}{Junyuan Shang},
  \bibinfo{person}{Jiaxiang Liu}, \bibinfo{person}{Xuyi Chen},
  \bibinfo{person}{Yanbin Zhao}, \bibinfo{person}{Yuxiang Lu},
  \bibinfo{person}{Weixin Liu}, \bibinfo{person}{Zhihua Wu},
  \bibinfo{person}{Weibao Gong}, \bibinfo{person}{Jianzhong Liang},
  \bibinfo{person}{Zhizhou Shang}, \bibinfo{person}{Peng Sun},
  \bibinfo{person}{Wei Liu}, \bibinfo{person}{Xuan Ouyang},
  \bibinfo{person}{Dianhai Yu}, \bibinfo{person}{Hao Tian},
  \bibinfo{person}{Hua Wu}, {and} \bibinfo{person}{Haifeng Wang}.}
  \bibinfo{year}{2021}\natexlab{}.
\newblock \showarticletitle{{ERNIE} 3.0: Large-scale Knowledge Enhanced
  Pre-training for Language Understanding and Generation}.
\newblock \bibinfo{journal}{\emph{CoRR}}  \bibinfo{volume}{abs/2107.02137}
  (\bibinfo{year}{2021}).
\newblock


\bibitem[Sun(2023)]%
        {Sun-arxiv-2023-A}
\bibfield{author}{\bibinfo{person}{Zhongxiang Sun}.}
  \bibinfo{year}{2023}\natexlab{}.
\newblock \showarticletitle{A Short Survey of Viewing Large Language Models in
  Legal Aspect}.
\newblock \bibinfo{journal}{\emph{CoRR}}  \bibinfo{volume}{abs/2303.09136}
  (\bibinfo{year}{2023}).
\newblock


\bibitem[Taori et~al\mbox{.}(2023)]%
        {Taori-github-2023-Stanford-alpaca}
\bibfield{author}{\bibinfo{person}{Rohan Taori}, \bibinfo{person}{Ishaan
  Gulrajani}, \bibinfo{person}{Tianyi Zhang}, \bibinfo{person}{Yann Dubois},
  \bibinfo{person}{Xuechen Li}, \bibinfo{person}{Carlos Guestrin},
  \bibinfo{person}{Percy Liang}, {and} \bibinfo{person}{Tatsunori~B.
  Hashimoto}.} \bibinfo{year}{2023}\natexlab{}.
\newblock \bibinfo{title}{Stanford Alpaca: An Instruction-following LLaMA
  model}.
\newblock
  \bibinfo{howpublished}{\url{https://github.com/tatsu-lab/stanford_alpaca}}.
\newblock


\bibitem[Touvron et~al\mbox{.}(2023)]%
        {Touvron-arxiv-2023-LLaMA}
\bibfield{author}{\bibinfo{person}{Hugo Touvron}, \bibinfo{person}{Thibaut
  Lavril}, \bibinfo{person}{Gautier Izacard}, \bibinfo{person}{Xavier
  Martinet}, \bibinfo{person}{Marie{-}Anne Lachaux},
  \bibinfo{person}{Timoth{\'{e}}e Lacroix}, \bibinfo{person}{Baptiste
  Rozi{\`{e}}re}, \bibinfo{person}{Naman Goyal}, \bibinfo{person}{Eric Hambro},
  \bibinfo{person}{Faisal Azhar}, \bibinfo{person}{Aur{\'{e}}lien Rodriguez},
  \bibinfo{person}{Armand Joulin}, \bibinfo{person}{Edouard Grave}, {and}
  \bibinfo{person}{Guillaume Lample}.} \bibinfo{year}{2023}\natexlab{}.
\newblock \showarticletitle{LLaMA: Open and Efficient Foundation Language
  Models}.
\newblock \bibinfo{journal}{\emph{CoRR}}  \bibinfo{volume}{abs/2302.13971}
  (\bibinfo{year}{2023}).
\newblock


\bibitem[Wang et~al\mbox{.}(2022b)]%
        {Wang-ICML-2022-What}
\bibfield{author}{\bibinfo{person}{Thomas Wang}, \bibinfo{person}{Adam
  Roberts}, \bibinfo{person}{Daniel Hesslow}, \bibinfo{person}{Teven~Le Scao},
  \bibinfo{person}{Hyung~Won Chung}, \bibinfo{person}{Iz Beltagy},
  \bibinfo{person}{Julien Launay}, {and} \bibinfo{person}{Colin Raffel}.}
  \bibinfo{year}{2022}\natexlab{b}.
\newblock \showarticletitle{What Language Model Architecture and Pretraining
  Objective Works Best for Zero-Shot Generalization?}. In
  \bibinfo{booktitle}{\emph{International Conference on Machine Learning,
  {ICML} 2022, 17-23 July 2022, Baltimore, Maryland, {USA}}}
  \emph{(\bibinfo{series}{Proceedings of Machine Learning Research},
  Vol.~\bibinfo{volume}{162})}. \bibinfo{pages}{22964--22984}.
\newblock


\bibitem[Wang et~al\mbox{.}(2022a)]%
        {Wang-EMNLP-2022-Super}
\bibfield{author}{\bibinfo{person}{Yizhong Wang}, \bibinfo{person}{Swaroop
  Mishra}, \bibinfo{person}{Pegah Alipoormolabashi}, \bibinfo{person}{Yeganeh
  Kordi}, \bibinfo{person}{Amirreza Mirzaei}, \bibinfo{person}{Atharva Naik},
  \bibinfo{person}{Arjun Ashok}, \bibinfo{person}{Arut~Selvan Dhanasekaran},
  \bibinfo{person}{Anjana Arunkumar}, \bibinfo{person}{David Stap},
  \bibinfo{person}{Eshaan Pathak}, \bibinfo{person}{Giannis Karamanolakis},
  \bibinfo{person}{Haizhi~Gary Lai}, \bibinfo{person}{Ishan Purohit},
  \bibinfo{person}{Ishani Mondal}, \bibinfo{person}{Jacob Anderson},
  \bibinfo{person}{Kirby Kuznia}, \bibinfo{person}{Krima Doshi},
  \bibinfo{person}{Kuntal~Kumar Pal}, \bibinfo{person}{Maitreya Patel},
  \bibinfo{person}{Mehrad Moradshahi}, \bibinfo{person}{Mihir Parmar},
  \bibinfo{person}{Mirali Purohit}, \bibinfo{person}{Neeraj Varshney},
  \bibinfo{person}{Phani~Rohitha Kaza}, \bibinfo{person}{Pulkit Verma},
  \bibinfo{person}{Ravsehaj~Singh Puri}, \bibinfo{person}{Rushang Karia},
  \bibinfo{person}{Savan Doshi}, \bibinfo{person}{Shailaja~Keyur Sampat},
  \bibinfo{person}{Siddhartha Mishra}, \bibinfo{person}{Sujan~Reddy A},
  \bibinfo{person}{Sumanta Patro}, \bibinfo{person}{Tanay Dixit}, {and}
  \bibinfo{person}{Xudong Shen}.} \bibinfo{year}{2022}\natexlab{a}.
\newblock \showarticletitle{Super-NaturalInstructions: Generalization via
  Declarative Instructions on 1600+ {NLP} Tasks}. In
  \bibinfo{booktitle}{\emph{{EMNLP}}}. \bibinfo{pages}{5085--5109}.
\newblock


\bibitem[Wei et~al\mbox{.}(2022a)]%
        {Wei-ICLR-2022-Finetuned}
\bibfield{author}{\bibinfo{person}{Jason Wei}, \bibinfo{person}{Maarten Bosma},
  \bibinfo{person}{Vincent~Y. Zhao}, \bibinfo{person}{Kelvin Guu},
  \bibinfo{person}{Adams~Wei Yu}, \bibinfo{person}{Brian Lester},
  \bibinfo{person}{Nan Du}, \bibinfo{person}{Andrew~M. Dai}, {and}
  \bibinfo{person}{Quoc~V. Le}.} \bibinfo{year}{2022}\natexlab{a}.
\newblock \showarticletitle{Finetuned Language Models are Zero-Shot Learners}.
  In \bibinfo{booktitle}{\emph{{ICLR}}}.
\newblock


\bibitem[Wei et~al\mbox{.}(2022b)]%
        {Wei-arxiv-2022-Emergent}
\bibfield{author}{\bibinfo{person}{Jason Wei}, \bibinfo{person}{Yi Tay},
  \bibinfo{person}{Rishi Bommasani}, \bibinfo{person}{Colin Raffel},
  \bibinfo{person}{Barret Zoph}, \bibinfo{person}{Sebastian Borgeaud},
  \bibinfo{person}{Dani Yogatama}, \bibinfo{person}{Maarten Bosma},
  \bibinfo{person}{Denny Zhou}, \bibinfo{person}{Donald Metzler},
  \bibinfo{person}{Ed~H. Chi}, \bibinfo{person}{Tatsunori Hashimoto},
  \bibinfo{person}{Oriol Vinyals}, \bibinfo{person}{Percy Liang},
  \bibinfo{person}{Jeff Dean}, {and} \bibinfo{person}{William Fedus}.}
  \bibinfo{year}{2022}\natexlab{b}.
\newblock \showarticletitle{Emergent Abilities of Large Language Models}.
\newblock \bibinfo{journal}{\emph{CoRR}}  \bibinfo{volume}{abs/2206.07682}
  (\bibinfo{year}{2022}).
\newblock


\bibitem[Wei et~al\mbox{.}(2023b)]%
        {Wei-2023-arXiv-symbol}
\bibfield{author}{\bibinfo{person}{Jerry~W. Wei}, \bibinfo{person}{Le Hou},
  \bibinfo{person}{Andrew~K. Lampinen}, \bibinfo{person}{Xiangning Chen},
  \bibinfo{person}{Da Huang}, \bibinfo{person}{Yi Tay}, \bibinfo{person}{Xinyun
  Chen}, \bibinfo{person}{Yifeng Lu}, \bibinfo{person}{Denny Zhou},
  \bibinfo{person}{Tengyu Ma}, {and} \bibinfo{person}{Quoc~V. Le}.}
  \bibinfo{year}{2023}\natexlab{b}.
\newblock \showarticletitle{Symbol tuning improves in-context learning in
  language models}.
\newblock \bibinfo{journal}{\emph{CoRR}}  \bibinfo{volume}{abs/2305.08298}
  (\bibinfo{year}{2023}).
\newblock


\bibitem[Wei et~al\mbox{.}(2023a)]%
        {wei-arxiv-2023-zero}
\bibfield{author}{\bibinfo{person}{Xiang Wei}, \bibinfo{person}{Xingyu Cui},
  \bibinfo{person}{Ning Cheng}, \bibinfo{person}{Xiaobin Wang},
  \bibinfo{person}{Xin Zhang}, \bibinfo{person}{Shen Huang},
  \bibinfo{person}{Pengjun Xie}, \bibinfo{person}{Jinan Xu},
  \bibinfo{person}{Yufeng Chen}, \bibinfo{person}{Meishan Zhang},
  \bibinfo{person}{Yong Jiang}, {and} \bibinfo{person}{Wenjuan Han}.}
  \bibinfo{year}{2023}\natexlab{a}.
\newblock \showarticletitle{Zero-Shot Information Extraction via Chatting with
  ChatGPT}.
\newblock \bibinfo{journal}{\emph{CoRR}}  \bibinfo{volume}{abs/2302.10205}
  (\bibinfo{year}{2023}).
\newblock


\bibitem[Wikipedia(2023)]%
        {Wikipedia}
\bibfield{author}{\bibinfo{person}{Wikipedia}.}
  \bibinfo{year}{2023}\natexlab{}.
\newblock
\newblock
\urldef\tempurl%
\url{https://en.wikipedia.org/wiki/Main_Page}
\showURL{%
\tempurl}


\bibitem[Wolf et~al\mbox{.}(2020)]%
        {Wolf-EMNLP-2020-Transformers}
\bibfield{author}{\bibinfo{person}{Thomas Wolf}, \bibinfo{person}{Lysandre
  Debut}, \bibinfo{person}{Victor Sanh}, \bibinfo{person}{Julien Chaumond},
  \bibinfo{person}{Clement Delangue}, \bibinfo{person}{Anthony Moi},
  \bibinfo{person}{Pierric Cistac}, \bibinfo{person}{Tim Rault},
  \bibinfo{person}{R{\'{e}}mi Louf}, \bibinfo{person}{Morgan Funtowicz},
  \bibinfo{person}{Joe Davison}, \bibinfo{person}{Sam Shleifer},
  \bibinfo{person}{Patrick von Platen}, \bibinfo{person}{Clara Ma},
  \bibinfo{person}{Yacine Jernite}, \bibinfo{person}{Julien Plu},
  \bibinfo{person}{Canwen Xu}, \bibinfo{person}{Teven~Le Scao},
  \bibinfo{person}{Sylvain Gugger}, \bibinfo{person}{Mariama Drame},
  \bibinfo{person}{Quentin Lhoest}, {and} \bibinfo{person}{Alexander~M. Rush}.}
  \bibinfo{year}{2020}\natexlab{}.
\newblock \showarticletitle{Transformers: State-of-the-Art Natural Language
  Processing}. In \bibinfo{booktitle}{\emph{{EMNLP} (Demos)}}.
  \bibinfo{pages}{38--45}.
\newblock


\bibitem[Wu et~al\mbox{.}(2023)]%
        {wu-arxiv-2023-bloomberggpt}
\bibfield{author}{\bibinfo{person}{Shijie Wu}, \bibinfo{person}{Ozan Irsoy},
  \bibinfo{person}{Steven Lu}, \bibinfo{person}{Vadim Dabravolski},
  \bibinfo{person}{Mark Dredze}, \bibinfo{person}{Sebastian Gehrmann},
  \bibinfo{person}{Prabhanjan Kambadur}, \bibinfo{person}{David~S. Rosenberg},
  {and} \bibinfo{person}{Gideon Mann}.} \bibinfo{year}{2023}\natexlab{}.
\newblock \showarticletitle{BloombergGPT: {A} Large Language Model for
  Finance}.
\newblock \bibinfo{journal}{\emph{CoRR}}  \bibinfo{volume}{abs/2303.17564}
  (\bibinfo{year}{2023}).
\newblock


\bibitem[Xie et~al\mbox{.}(2023)]%
        {Xie-arxiv-2023-doremi}
\bibfield{author}{\bibinfo{person}{Sang~Michael Xie}, \bibinfo{person}{Hieu
  Pham}, \bibinfo{person}{Xuanyi Dong}, \bibinfo{person}{Nan Du},
  \bibinfo{person}{Hanxiao Liu}, \bibinfo{person}{Yifeng Lu},
  \bibinfo{person}{Percy Liang}, \bibinfo{person}{Quoc~V. Le},
  \bibinfo{person}{Tengyu Ma}, {and} \bibinfo{person}{Adams~Wei Yu}.}
  \bibinfo{year}{2023}\natexlab{}.
\newblock \showarticletitle{DoReMi: Optimizing Data Mixtures Speeds Up Language
  Model Pretraining}.
\newblock \bibinfo{journal}{\emph{CoRR}}  \bibinfo{volume}{abs/2305.10429}
  (\bibinfo{year}{2023}).
\newblock


\bibitem[Yang et~al\mbox{.}(2023)]%
        {Yang-2023-arxiv-FinGPT}
\bibfield{author}{\bibinfo{person}{Hongyang Yang}, \bibinfo{person}{Xiao-Yang
  Liu}, {and} \bibinfo{person}{Christina~Dan Wang}.}
  \bibinfo{year}{2023}\natexlab{}.
\newblock \showarticletitle{FinGPT: Open-Source Financial Large Language
  Models}.
\newblock \bibinfo{journal}{\emph{CoRR}}  \bibinfo{volume}{abs/2306.06031}
  (\bibinfo{year}{2023}).
\newblock


\bibitem[Zeng et~al\mbox{.}(2022)]%
        {Zeng-arxiv-2022-GLM}
\bibfield{author}{\bibinfo{person}{Aohan Zeng}, \bibinfo{person}{Xiao Liu},
  \bibinfo{person}{Zhengxiao Du}, \bibinfo{person}{Zihan Wang},
  \bibinfo{person}{Hanyu Lai}, \bibinfo{person}{Ming Ding},
  \bibinfo{person}{Zhuoyi Yang}, \bibinfo{person}{Yifan Xu},
  \bibinfo{person}{Wendi Zheng}, \bibinfo{person}{Xiao Xia},
  \bibinfo{person}{Weng~Lam Tam}, \bibinfo{person}{Zixuan Ma},
  \bibinfo{person}{Yufei Xue}, \bibinfo{person}{Jidong Zhai},
  \bibinfo{person}{Wenguang Chen}, \bibinfo{person}{Peng Zhang},
  \bibinfo{person}{Yuxiao Dong}, {and} \bibinfo{person}{Jie Tang}.}
  \bibinfo{year}{2022}\natexlab{}.
\newblock \showarticletitle{{GLM-130B:} An Open Bilingual Pre-trained Model}.
\newblock   \bibinfo{volume}{abs/2210.02414} (\bibinfo{year}{2022}).
\newblock


\bibitem[Zhang and Sennrich(2019)]%
        {Zhang-NIPS-2019-Root}
\bibfield{author}{\bibinfo{person}{Biao Zhang} {and} \bibinfo{person}{Rico
  Sennrich}.} \bibinfo{year}{2019}\natexlab{}.
\newblock \showarticletitle{Root Mean Square Layer Normalization}. In
  \bibinfo{booktitle}{\emph{{NeurIPS}}}. \bibinfo{pages}{12360--12371}.
\newblock


\bibitem[Zhang et~al\mbox{.}(2022)]%
        {Zhang-arxiv-2022-OPT}
\bibfield{author}{\bibinfo{person}{Susan Zhang}, \bibinfo{person}{Stephen
  Roller}, \bibinfo{person}{Naman Goyal}, \bibinfo{person}{Mikel Artetxe},
  \bibinfo{person}{Moya Chen}, \bibinfo{person}{Shuohui Chen},
  \bibinfo{person}{Christopher Dewan}, \bibinfo{person}{Mona~T. Diab},
  \bibinfo{person}{Xian Li}, \bibinfo{person}{Xi~Victoria Lin},
  \bibinfo{person}{Todor Mihaylov}, \bibinfo{person}{Myle Ott},
  \bibinfo{person}{Sam Shleifer}, \bibinfo{person}{Kurt Shuster},
  \bibinfo{person}{Daniel Simig}, \bibinfo{person}{Punit~Singh Koura},
  \bibinfo{person}{Anjali Sridhar}, \bibinfo{person}{Tianlu Wang}, {and}
  \bibinfo{person}{Luke Zettlemoyer}.} \bibinfo{year}{2022}\natexlab{}.
\newblock \showarticletitle{{OPT:} Open Pre-trained Transformer Language
  Models}.
\newblock \bibinfo{journal}{\emph{CoRR}}  \bibinfo{volume}{abs/2205.01068}
  (\bibinfo{year}{2022}).
\newblock


\bibitem[Zhao et~al\mbox{.}(2023)]%
        {rucLLMSurvey}
\bibfield{author}{\bibinfo{person}{Wayne~Xin Zhao}, \bibinfo{person}{Kun Zhou},
  \bibinfo{person}{Junyi Li}, \bibinfo{person}{Tianyi Tang},
  \bibinfo{person}{Xiaolei Wang}, \bibinfo{person}{Yupeng Hou},
  \bibinfo{person}{Yingqian Min}, \bibinfo{person}{Beichen Zhang},
  \bibinfo{person}{Junjie Zhang}, \bibinfo{person}{Zican Dong},
  \bibinfo{person}{Yifan Du}, \bibinfo{person}{Chen Yang},
  \bibinfo{person}{Yushuo Chen}, \bibinfo{person}{Zhipeng Chen},
  \bibinfo{person}{Jinhao Jiang}, \bibinfo{person}{Ruiyang Ren},
  \bibinfo{person}{Yifan Li}, \bibinfo{person}{Xinyu Tang},
  \bibinfo{person}{Zikang Liu}, \bibinfo{person}{Peiyu Liu},
  \bibinfo{person}{Jian{-}Yun Nie}, {and} \bibinfo{person}{Ji{-}Rong Wen}.}
  \bibinfo{year}{2023}\natexlab{}.
\newblock \showarticletitle{A Survey of Large Language Models}.
\newblock \bibinfo{journal}{\emph{CoRR}}  \bibinfo{volume}{abs/2303.18223}
  (\bibinfo{year}{2023}).
\newblock


\bibitem[Zhong et~al\mbox{.}(2023)]%
        {Zhong-2023-arxiv-AGIEval}
\bibfield{author}{\bibinfo{person}{Wanjun Zhong}, \bibinfo{person}{Ruixiang
  Cui}, \bibinfo{person}{Yiduo Guo}, \bibinfo{person}{Yaobo Liang},
  \bibinfo{person}{Shuai Lu}, \bibinfo{person}{Yanlin Wang},
  \bibinfo{person}{Amin Saied}, \bibinfo{person}{Weizhu Chen}, {and}
  \bibinfo{person}{Nan Duan}.} \bibinfo{year}{2023}\natexlab{}.
\newblock \showarticletitle{AGIEval: {A} Human-Centric Benchmark for Evaluating
  Foundation Models}.
\newblock \bibinfo{journal}{\emph{CoRR}}  \bibinfo{volume}{abs/2304.06364}
  (\bibinfo{year}{2023}).
\newblock


\bibitem[Zhu et~al\mbox{.}(2015)]%
        {Zhu-ICCV-2015-Aligning}
\bibfield{author}{\bibinfo{person}{Yukun Zhu}, \bibinfo{person}{Ryan Kiros},
  \bibinfo{person}{Richard~S. Zemel}, \bibinfo{person}{Ruslan Salakhutdinov},
  \bibinfo{person}{Raquel Urtasun}, \bibinfo{person}{Antonio Torralba}, {and}
  \bibinfo{person}{Sanja Fidler}.} \bibinfo{year}{2015}\natexlab{}.
\newblock \showarticletitle{Aligning Books and Movies: Towards Story-Like
  Visual Explanations by Watching Movies and Reading Books}. In
  \bibinfo{booktitle}{\emph{{ICCV}}}. \bibinfo{pages}{19--27}.
\newblock


\end{thebibliography}

\newpage
\appendix

\section*{Appendix of \oursys: A One-Stop Data Processing System for Large
Language Models}
\label{sec:appendix}

\section{Additional Details of \oursys}
\label{appendx:details}

\subsection{Base Classes of OPs in \oursys}
\label{sec:appendix_base_class_ops}
We illustrate the core base classes of operators (OPs) in \oursys at listing \ref{listing:base_op_classes}.

\subsection{Theoretical Analysis of Space Usage for Caches and Checkpoints}
\label{appendx:details:space_usage_cache_ckpt}

Caches are generated after some of the functions of \pacterm{Dataset}, such as \pacterm{map}, \pacterm{filter}. Generally, caches can be categorized into cache data and indices. The total size of a set of indices is very small so we can ignore these parts when conducting the space usage analysis. On the contrary, the size of the cache data is nearly the same as the input dataset. Here we assume that the sizes of cache data and checkpoints are all the same as the input dataset's size. And there must be one cache data file for the original dataset after it's loaded.

Assume that there are $M$ Mappers, $F$ Filters, and $D$ Deduplicators in the processing configuration, and the size of the original dataset is $S$, the detailed analysis for cache mode and checkpoint mode is shown below.

\para{Space Usage of Cache Mode.}
Caches are generated after each OP. Mappers, Filters, and Deduplicators only generate one set of cache data. Besides, the first Filter would generate an extra set of cache data because a new column for storing statistics will be added to the dataset. Therefore the total disk space usage of caches is:
$$Space_{[cache\_mode]} = (1 + M + F + \mathbb{I}(F > 0) + D) \times S,$$
where $\mathbb{I}(\cdot)$ is the indicator function, which returns $1$ when $\cdot$ is true, otherwise returns $0$.

\para{Space Usage of Checkpoint Mode.}
Checkpoints are only generated when any exception or error occurs. However, caches are still stored after disabling the cache mode due to the features of \pacterm{Dataset}. We clean up older caches after each OP. The detailed cleanup pipeline is: 1). OP$_i$ finished, 2). caches for OP$_i$ generated, 3). caches for OP$_{i-1}$ cleaned up. Thus there exists at most two sets of caches at the same time theoretically in step 2. Considering the caches of the original dataset, the peak disk space usage of caches in checkpoint mode is:
$$Space_{[checkpoint\_mode]} = 3 \times S. $$

\begin{listing}\begin{minted}
	[
	linenos,
	frame=lines,
	fontsize=\footnotesize,
	]
	{python}
class Formatter:
    ...
    def load_dataset(self, *args) -> Dataset:
        ...
    ...
	
class Mapper:
    ...
    def process(self, sample: Dict) -> Dict:
        ...
    ...
	
class Filter:
    ...
    def compute_stats(self, sample: Dict) -> Dict:
        ...
	
    def process(self, sample: Dict) -> bool:
        ...
    ...
	
class Deduplicator:
    ...
    def compute_hash(self, sample: Dict) -> Dict:
        ...
	
    def process(self, dataset: Dataset) -> Dataset:
        ...
    ...
	\end{minted}
	\caption{The illustration of OP base classes in \oursys.}
	\label{listing:base_op_classes}
\end{listing}

\section{Additional Numerical Results}
\label{appendx:exp}

\begin{table}[!ht]
\centering
\caption{Evaluation results of three types of quality classifiers.}
\label{tab:qc_eval_result}
\begin{tblr}{
  width = \linewidth,
  colspec = {Q[20]Q[20]Q[20]Q[20]},
  cells = {c},
  hline{1,5} = {-}{0.08em},
}
\textbf{ Quality Classifier } & \textbf{ Precision } & \textbf{ Recall } & \textbf{ F1 } \\
\hline
\hline
\textsc{GPT-3} & 96.82\% & 98.14\% & 97.47\% \\
\hline
\textsc{Chinese} & 98.00\% & 99.30\% & 98.64\% \\
\hline
\textsc{Code} & 71.23\% & 54.21\% & 61.56\%
\end{tblr}
\end{table}

\begin{table*}[!ht]
\centering
\caption{Training configuration of 3 types of quality classifiers.}
\label{tab:qc_configs}
\begin{tblr}{
  width = \linewidth,
  colspec = {Q[20]Q[20]Q[10]Q[50]Q[50]},
  cells = {c,m},
  hline{1,5} = {-}{0.08em},
}
\textbf{Quality Classifier} & \textbf{Tokenizer} & \textbf{Keep Method} & \textbf{Positive Datasets} & \textbf{Negative Datasets} \\
\hline
\hline
\textsc{GPT-3} & Standard Tokenizer & pareto & Wikipedia-en \& books1 \& OpenWebText2 & CommonCrawl \\
\hline
\textsc{Chinese} & Sentencepiece & label & Wikipeida-zh \& Wudao & Samples in Chinese from CommonCrawl \\
\hline
\textsc{Code} & Sentencepiece & label & Samples with max\_stars\_count>=1372 from TheStack & Random Samples from the rest of TheStack
\end{tblr}
\end{table*}

\subsection{Quality Classifier}
\label{appendx:quality_score}

Firstly, we will show how we can reproduce the GPT-3 and achieve comparable performance.

We follow the training procedure of quality classifier in GPT-3 \cite{brown2020gpt3} that used a logistic regression classifier with features from standard tokenizer and HashingTF of PySpark. Based on this, we expand this training pipeline to Chinese text and various code types. The training details are listed in Table \ref{tab:qc_configs}, where the keeping method includes:
\begin{itemize}
    \item label: $doc\_score > 0.5$
    \item pareto \cite{brown2020gpt3}: $doc\_score > 1 - np.random.pareto(\alpha),\;\alpha=9$
\end{itemize}

We split these datasets into training and evaluation splits with a split ratio of 4:1. Then these classifiers trained on the training split are evaluated on the evaluation split. Experimental results are shown in Table \ref{tab:qc_eval_result}. As we can see, reproduced GPT-3 and its Chinese version perform well except for the Code version. We speculate that the positive and negative splitting method for Code quality classifier now might not be a good choice, and we leave this issue to future research.

Besides, we compare keeping ratios when using these classifiers to re-sample CommonCrawl between the original GPT-3 quality classifier and our reproduced classifiers, which is shown in Table \ref{tab:qc_keep_ratio}. The keeping ratio of the original GPT-3 quality classifier is estimated by the data size before and after filtering described in GPT-3 paper \cite{brown2020gpt3}. We can see that the keeping ratios of our reproduced GPT-3 quality classifiers are basically aligned with the original one.

\subsection{Data Recipes}
\label{appendx:data-recipe}

For pre-training data, we acquired a vast amount of raw textual corpora primarily following the procedural guidelines of RedPajama \cite{together2023redpajama} and the Pile \cite{Gao-arxiv-2021-Pile}. The common subsets were merged and subjected to \oursys refinements. The resultant data recipe is presented in Table \ref{tab:recipe_pt}, which covers 15 prominent components. We use the SentencePiece \cite{Kudo-EMNLP-2018-SentencePiece} tokenizer as implemented in GPT-NeoX-20B \cite{Black-CoRR-2022-GPT} to prepare text and report the counted number of tokens. The sampling proportion is the normalization of token numbers, except for Books and Wikipedia, which undergo 2 and 2.5 epochs respectively, to enhance the weighting of high-quality corpora.

\begin{table}[!ht]
\centering
\caption{Statistics of \oursys's pre-training data.}
\label{tab:recipe_pt}
\begin{tblr}{
  width = \linewidth,
  colspec = {Q[l]S[table-format=12.0,group-separator={,}]S[table-format=2.2]},
  hline{1,Z} = {-}{0.08em},
}
\textbf{Component} & \textbf{\#Tokens} & \textbf{Sampling prop.} \\
\hline
\hline
CommonCrawl        & 360925581674   & 44.91\%                 \\
C4                 & 181951688729   & 22.64\%                 \\
GitHub             & 65076921292    & 8.10\%                  \\
Books              & 26389944579    & 6.57\%                  \\
Wikipedia          & 17615935449    & 5.48\%                  \\
arXiv              & 29093082586    & 3.62\%                  \\
PubMed Central     & 25589708647    & 3.18\%                  \\
StackExchange      & 19793629900    & 2.46\%                  \\
FreeLaw            & 13057506102    & 1.62\%                  \\
PubMed Abstracts   & 5208343613     & 0.65\%                  \\
USPTO              & 4021281155     & 0.50\%                  \\
EuroParl           & 780962770      & 0.10\%                  \\
HackerNews         & 485584871      & 0.06\%                  \\
PhilPapers         & 478040431      & 0.06\%                  \\
HIH ExPorter       & 436414852      & 0.05\%                  \\
\end{tblr}
\end{table}

For fine-tuning data, we merge and refine tens of Alpaca-CoT datasets. 
Each dataset can be categorized into English, Chinese and Multilingual by language; 
into instruct fine-tuning, and chat fine-tuning including sinlge-round dialog, multi-round dialog and preference by usage;
multi-task and task-specific by task type;
and human-generated, self-instruct, and mixed collection of datasets by the generation method. 
The detailed numbers of datasets for each category are presented in Table \ref{tab:recipe_post}.

\begin{table}[!ht]
\centering
\caption{Statistics of \oursys fine-tuning data used in our experiments. $^*$These tags are newly added by \oursys compared to the original tag sets of Alpaca-CoT \cite{alpaca-cot}.``CFT'' indicates Chat Fine-Tuning.}
\label{tab:recipe_post}
\begin{tblr}{
  width = \linewidth,
  cell{2}{1} = {r=3}{},
  cell{5}{1} = {r=4}{},
  cell{9}{1} = {r=2}{},
  cell{11}{1} = {r=4}{},
  column{Z} = {c},
  hline{1,Z} = {-}{0.08em},
}
\textbf{Category} & \textbf{Sub-Category} & \textbf{\#Datasets} \\
\hline
\hline
Language & English        & 28                  \\
& Chinese                 & 14                \\
& Multilingual             & 3                  \\
\hline
Usage$^*$ & Instruct Fine-Tuning (IFT)   & 17                    \\
& CFT: Single-Round Dialog             & 23                 \\
& CFT: Multi-Round Dialog              & 2                    \\
& CFT: Preference    & 5                 \\
\hline
Task Type & Multi-Task & 27 \\
& Task-Specific & 13 \\
\hline
Generation Method & Human-Generated & 3 \\
& Self-Instruct & 12 \\
& Mixted & 5 \\
& Collection of Datasets & 19 \\
\end{tblr}
\end{table}

More information about these datasets can be found on the \oursys recipes page\footnote{\href{https://github.com/alibaba/data-juicer/blob/main/configs/data_juicer_recipes}{https://github.com/alibaba/data-juicer/blob/main/configs/data\_juicer\_recipes}} of our repository.

\subsection{Experiments Details}
\label{appendx:imp-details}

\twosubsection{Models and Training For Pre-training Data.}
\label{appendx:imp-details:pretrain}
We adhere to the official paper \cite{Touvron-arxiv-2023-LLaMA} and leverage open-source implementation \cite{openlm2023openllama} to build standard LLaMA models. Basically, it is to apply RMSNorm \cite{Zhang-NIPS-2019-Root}, the SwiGLU activation \cite{Shazeer-arxiv-2020-GLU}, and rotary positional embedding \cite{su-2021-arxiv-reformer} on the decoder-only transformer architecture. The LLaMA-1.3B model is composed of 24 transformer layers, each with 16 self-attention heads and 2048 bottleneck units.

LLMs are pre-trained using the AdamW optimizer \cite{Loshchilov-arxiv-2017-Fixing} with hyper-parameters $\beta_1 = 0.9$ and $\beta_2 = 0.95$. For LLaMA-1.3B, the initial learning rate gradually increases to 2e-5 using 1\% warm-up steps and finally decays to 10\% through a cosine schedule. The weight decay is set to 0.1 and the gradient $\ell_2$-norm is clipped to 1.0.

\twosubsection{Models and Training of Fine-Tuning Data.}
\label{appendx:imp-details:posttune}
In fine-tuning, we choose LLaMA-7B as our basic model and fine-tuned it for 3 epochs. 
We follow the hyper-parameter settings in Alpaca \cite{Taori-github-2023-Stanford-alpaca}. Specifically, the optimizer is AdamW with a learning rate of 2e-5, global batch size of 256, and weight decay of 0. The learning rate schedules in a cosine style with 3\% initial warm-up steps.

Regarding the data recipes in Table \ref{tab:post-tune-res}, for \textbf{(CFT, EN)} case, we consider 5 competitive subsets (Alpaca, GPTeacher, FastChat, Guanaco, and CodeAlpaca) from Alpaca-CoT as candidate datasets; for \textbf{(CFT, ZH)} case, we use (AlpacaGPT4, Belle, Instinwild) as candidate datasets.
Generally speaking, we bucket from these candidate datasets according to more than a dozen built-in analytical dimensions, sampling a fixed amount of data from each dimension to increase the diversity of the processed data as appropriately as possible.
More detailed hyper-parameters of data processing can be found in our released data recipes.

Both the pre-trained and fine-tuned reference models are released in our homepage.

\twosubsection{System Performance Experiments.}
\label{appendx:imp-details:sys_perf}
The experiments of end-to-end processing mentioned in section \ref{exp:end2end_perf} are all conducted on the same machine with 128 cores of Intel(R) Xeon(R) Platinum 8369B models and about 990GB memory. Before starting these experiments, the original datasets, third-party models, and other assets will be prepared in advance for both baselines and \oursys, and the intermediate cache files will be cleaned after every complete process for \oursys. After processing, we use the same number of processes for processing the dataset to export the result dataset to the local SSD.

As for the resource monitoring tool, it's implemented based on the \textsc{psutil}\footnote{\href{https://github.com/giampaolo/psutil}{https://github.com/giampaolo/psutil}} library. It samples the memory for all related processes every second during the processing pipeline. Then we compute the average memory usage by summing the memory usage over all processes and dividing by the number of processes used in each experiment. Finally, we aggregate all data and compute the average memory usage over time.

\begin{table*}[!h]
	\centering
	\caption{Evaluation results on 16 core tasks of HELM benchmark.}
	\label{tab:helm_16core}
	\begin{tblr}{
			width = \linewidth,
			colspec = {Q[c]Q[si={table-format=2.1},c]Q[si={table-format=2.1},c]Q[si={table-format=2.1},c]Q[si={table-format=2.1},c]},
			column{1} = {l},
			row{1} = {guard},
			hline{1,18} = {-}{0.08em},
		}
		\textbf{Task} & \textbf{Falcon-1.3B} & \textbf{Pythia-1.4B} & {\textbf{LLaMA-1.3B }\\\textbf{(\oursys)}} & {\textbf{LLaMA-1.3B}\\\textbf{(\oursys IFT)}} \\
		\hline
		\hline
		MMLU                           & 24.7           & 26.0            & 25.9                         & 27.0                            \\
		BoolQ                          & 63.0            & 56.0            & 49.0                         & 56.0                             \\
		NarrativeQA                    &  32.1           &  31.5           & 38.2                         & 49.9                            \\
		NaturalQuestions (closed-book) & 10.7            & 10.5            & 10.1                         & 11.2                            \\
		NaturalQuestions (open-book)   & 50.0             & 49.8            & 45.9                         & 54.3                            \\
		QuAC                           & 24.3            &  26.5           & 26.0                         &  21.7                           \\
		HellaSwag                      & 67.0            & 57.0            & 56.0                         &  52.0                           \\
		OpenbookQA                     & 44.0            & 34.0            & 40.0                         & 43.0                            \\
		TruthfulQA                     & 19.0            & 21.0            & 33.0                         & 33.0                            \\
		MS MARCO (regular)             & 16.8            &  12.9           & 11.2                         & 12.1                            \\
		MS MARCO (TREC)                & 33.5            & 27.4            & 26.9                         & 28.1                            \\
		IMDB                           &  55.0           &  84.0           & 80.0                         & 84.0                            \\
		XSUM                           & 5.7            & { 6.5}            & 5.2                         & 5.3                            \\
		CNN/DailyMail                  & 4.0            &  8.4           & 7.8                         & 11.1                            \\
		CivilComments                  & 49.4          & 49.7            & 50.1                         & 50.0                            \\
		RAFT                           & 44.3            & 42.3            & 42.1                         & 49.3                           \\
	\end{tblr}
\end{table*}

\twosubsection{End-to-end System Baselines}
\label{appendx:sys_baselines}
We mainly compared the end-to-end system performance between our \oursys and two state-of-the-art baselines in the above experiments w.r.t system performance: RedPajama \cite{together2023redpajama} and Dolma  \cite{DolmaToolkit}. Besides the empirical comparsiton in Sec.\ref{exp:end2end_perf}, here we make more detailed introduction and comparison about them.

\textbf{RedPajama.} \footnote{We compared RedPajama in our experiments with its github commit ID as: 45b37c2a1d1e495b0f48549ef3ce03ff029f7881.}
The RedPajama project, developed by \textit{Together AI}, initially aims to reproduce the LLaMA training dataset \cite{Touvron-arxiv-2023-LLaMA} and open-source the entire code for data collection and processing, making it a significant and popular contribution to the LLM community. This is the primary reason for selecting it as our baseline. RedPajama provides a reproduced version of all seven subsets of the LLaMA training dataset, including arXiv, Books, C4, CommonCrawl, GitHub Code, Stack Exchange, and Wikipedia.

While RedPajama has made valuable contributions, our work explores different aspects and offers complementary features. 
For instance:
(1) RedPajama's design is closely tied to specific datasets, which present challenges for adapting its data processing pipelines to other datasets.
(2) Its focus on reproducing the LLaMA datasets lead to trade-offs in efficiency, which is not the primary concern of the RedPajama project.
(3) The current data processing component in RedPajama lacks systematization and customization. Adding new data processing methods to the existing pipelines would require understanding and modifying a significant portion of the code. 
As a result, most users typically opt to utilize the RedPajama Dataset directly rather than attempting to customize or improve its data processing pipelines.

\textbf{Dolma.}  \footnote{We compared Dolma in our experiments with its github commit ID as: 5a010a2685914b1db7744426abfb4b9ece52da95.}
The Dolma project, originating from \textit{Allen AI}, comprises two components: the Dolma Dataset and the Dolma Toolkit. It is also a newly established data processing initiative. 
We selected the Dolma Toolkit as a baseline because its objective of generating pre-training data for language modeling aligns with one of our target data types (we focus on both pre-training and fine-tuning data). 
The toolkit offers numerous ``Taggers'' that enable attribute tagging (analogous to 'stats' in \oursys) for each document sample. These tags are then used to filter out samples with undesirable attributes. Users have the flexibility to create custom taggers tailored to their specific needs.

However, we encountered several limitations when using Dolma for dataset processing. 
Firstly, Dolma's workflow involves multiple stages—tagging, deduplication, mixing, and various configurations—lacking support for an end-to-end data processing pipeline. 
Secondly, to leverage high-performance parallel processing, users are required to partition the input dataset into multiple shards in advance, incurring additional overhead. 
Thirdly, Dolma imposes certain requirements on input datasets, such as mandatory fields and a specific directory structure, necessitating further preprocessing before use. Moreover, it restricts input formats to JSONL or its gzipped variant. These constraints diminish the toolkit's flexibility, thereby increasing the cost of use and rendering the Dolma Toolkit relatively less user-friendly.

\twosubsection{Scalability.}
\label{appendx:imp-details:scalability}
Our experiments are performed on a platform comprising 16 servers, each equipped with a 64-core Intel(R) Xeon(R) Platinum CPU (mix of 8269CY and 8163 models) and 512 GB of memory. The network bandwidth shared among these servers is 20 Gbps. We utilize NAS storage to house both the raw data and the processed results.
For the scalability experiments, we consider the two baselines as follows:

 $\bullet$   \textit{\oursys on Ray}: We implement a Ray \cite{ray} executor for \oursys, which only adaptes the underlying interfaces of the \pacterm{HuggingFace-datasets} with \pacterm{Ray-datasets}, while all OPs of \oursys remain unchanged. 
This implies that users' code based on our native Python version can be seamlessly migrated from a single-machine version to distributed computing environments.

$\bullet$ \textit{\oursys on Beam}: This method is based on Apache Beam with the Apache Flink Runner. When compared to the Ray version, the Beam version requires additional code development to meet the demands of the Beam data processing pipeline. This includes the adaptations of several OPs and the replacement of the Formatter/Exporter with counterparts in Beam.

\subsection{Per-Task Evaluation}
\label{appendx:per-task-res}

For a thorough and consolidated assessment, we have summarized the individual scores of evaluated LLMs on the 16 core HELM assessment tasks in Table \ref{tab:helm_16core}.

\end{document}